\documentclass[acmlarge,bookmarksnumbered,unicode,pagebackref=true]{acmart}
\AtBeginDocument{%
  \providecommand\BibTeX{{%
    \normalfont B\kern-0.5em{\scshape i\kern-0.25em b}\kern-0.8em\TeX}}}

\setcopyright{acmlicensed}
\acmJournal{JOCCH}
\acmYear{2022} \acmVolume{1} \acmNumber{1} \acmArticle{1} \acmMonth{1} \acmPrice{15.00}\acmDOI{10.1145/3569089}

\acmJournal{JOCCH}

\usepackage{amsmath,amsfonts}%
\usepackage{soul}
\usepackage{nth}
\usepackage{microtype}
\usepackage{booktabs, multirow}%
\usepackage{url}
\usepackage{xspace}
\usepackage[export]{adjustbox}
\usepackage{subcaption}
\usepackage{latexsym}
\usepackage{cleveref}
\crefname{section}{Sec.}{Sections}
\crefname{figure}{Fig.}{Figs.}
\crefname{table}{Tab.}{Tabs.}
\crefname{equation}{Eq.}{Eqs.}
\crefname{appsec}{Appendix}{Appendices}

\usepackage{parskip}
\usepackage{url}
\usepackage{graphicx}
\usepackage{cjhebrew}

\newcommand{\eg}{\emph{e.\,g.,}\xspace}
\newcommand{\ie}{\emph{i.\,e.,}\xspace}
\newcommand{\etal}{\emph{et al.}\xspace}

\begin{document}

\title[Enhance Poses in Greek Vase Paintings]{Enhancing Human Pose Estimation in Ancient Vase Paintings via Perceptually-grounded Style Transfer Learning}

\author{Prathmesh Madhu}
\authornote{Both authors contributed equally to this research.}
\email{prathmesh.madhu@fau.de}
\orcid{}
\affiliation{%
  \institution{Pattern Recognition Lab, Friedrich Alexander University}
  \streetaddress{Martensstr. 3}
  \city{Erlangen}
  \state{Bavaria}
  \postcode{91054}
  \country{Germany}
}

\author{Angel Villar-Corrales}
\authornotemark[1]
\email{villar@ais.uni-bonn.de}
\orcid{}
\affiliation{%
  \institution{Autonomous Intelligent Systems, 
University of Bonn}
  \streetaddress{Friedrich-Hirzebruch-Allee 5}
  \city{Bonn}
  \state{North Rhine-Westphalia}
  \postcode{53115}
  \country{Germany}
}

\author{Ronak Kosti}
\email{ronak.kosti@fau.de}
\affiliation{%
  \institution{Pattern Recognition Lab, Friedrich Alexander University}
  \streetaddress{Martensstr. 3}
  \city{Erlangen}
  \state{Bavaria}
  \postcode{91054}
  \country{Germany}
}

\author{Torsten Bendschus}
\email{torsten.bendschus@fau.de}
\affiliation{%
  \institution{Institut f\"ur Klassische Arch\"aologie, FAU}
  \streetaddress{Kochstr. 4 / Nr. 19}
  \city{Erlangen}
  \state{Bavaria}
  \postcode{91054}
  \country{Germany}
}

\author{Corinna Reinhardt}
\email{corinna.reinhardt@fau.de}
\affiliation{%
  \institution{Institut f\"ur Klassische Arch\"aologie, FAU}
  \streetaddress{Kochstr. 4 / Nr. 19}
  \city{Erlangen}
  \state{Bavaria}
  \postcode{91054}
  \country{Germany}
}

\author{Peter Bell}
\email{peter.bell@uni-marburg.de}
\affiliation{%
  \institution{German Studies and Art Studies, Philipps University of Marburg}
  \streetaddress{Deutschhausstr. 3}
  \city{Marburg}
  \state{Hesse}
  \postcode{35037}
  \country{Germany}
}

\author{Andreas Maier}
\email{andreas.maier@fau.de}
\affiliation{%
  \institution{Pattern Recognition Lab, Friedrich Alexander University}
  \streetaddress{Martensstr. 3}
  \city{Erlangen}
  \state{Bavaria}
  \postcode{91054}
  \country{Germany}
}

\author{Vincent Christlein}
\email{vincent.christlein@fau.de}
\affiliation{%
  \institution{Pattern Recognition Lab, Friedrich Alexander University}
  \streetaddress{Martensstr. 3}
  \city{Erlangen}
  \state{Bavaria}
  \postcode{91054}
  \country{Germany}
}

\renewcommand{\shortauthors}{Madhu and Villar-Corrales, et al.}

\begin{abstract}
  Human pose estimation (HPE) is a central part of understanding the visual 
    narration and body movements of characters depicted in artwork collections, 
    such as Greek vase paintings. Unfortunately, existing HPE methods do not 
    generalise well across domains resulting in poorly recognised poses. 
    Therefore, we propose a two step approach: (1) adapting a dataset of natural 
    images of known person and pose annotations to the style of Greek vase 
    paintings by means of image style-transfer. We introduce a 
    perceptually-grounded style transfer training to enforce perceptual 
    consistency. Then, we fine-tune the base model with this newly created dataset. 
    We show that using style-transfer learning significantly improves the SOTA 
    performance on unlabelled data by more than 6\%~mean average precision (mAP) as 
    well as mean average recall (mAR). (2) To improve the already strong results 
    further, we created a small dataset (ClassArch) consisting of ancient Greek 
    vase paintings from the 6--\nth{5} century BCE with person and pose 
    annotations. We show that fine-tuning on this data with a style-transferred 
    model improves the performance further. In a thorough ablation study, we give a 
    targeted analysis of the influence of style intensities, revealing that the 
    model learns generic domain styles. Additionally, we provide a pose-based image 
    retrieval to demonstrate the effectiveness of our method. The code and pretrained models can be found at \url{https://github.com/angelvillar96/STLPose}.
\end{abstract}



\maketitle

\section{Introduction}\label{sec:intro}
\begin{figure}[t]
 	\begin{subfigure}{0.32\linewidth}
 		\centering
 		\includegraphics[width=\textwidth]{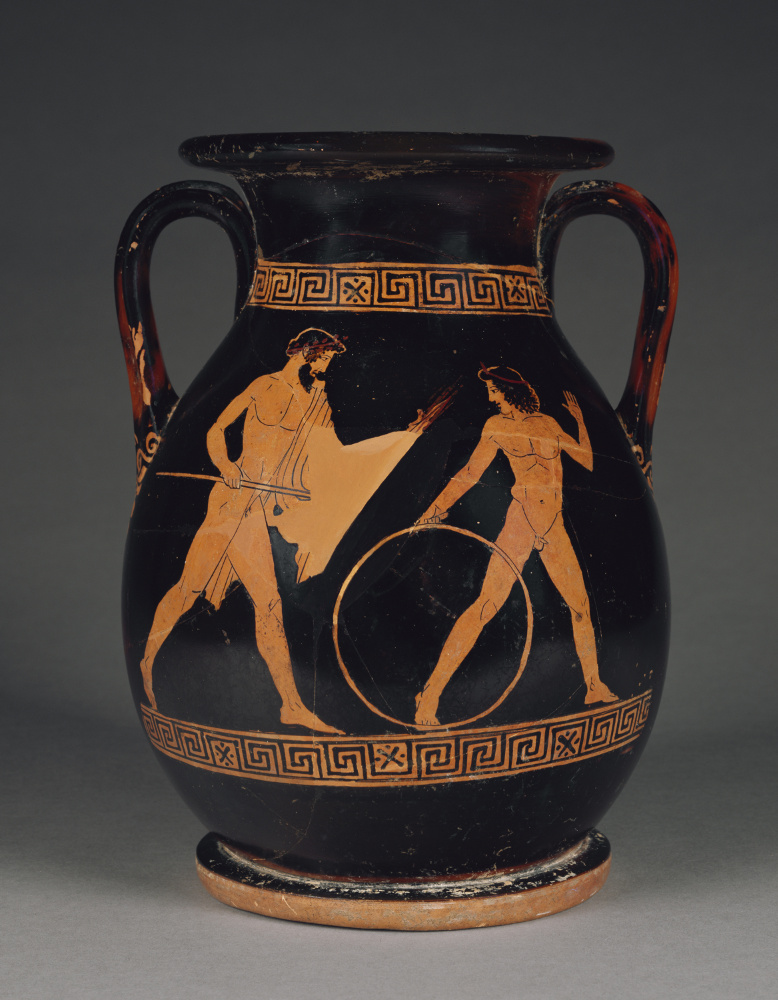}
 		\caption{Original} %
 		\label{fig:mainfig1}
 	\end{subfigure}
 	\begin{subfigure}{0.32\linewidth}
 		\centering
 		\includegraphics[width=\textwidth]{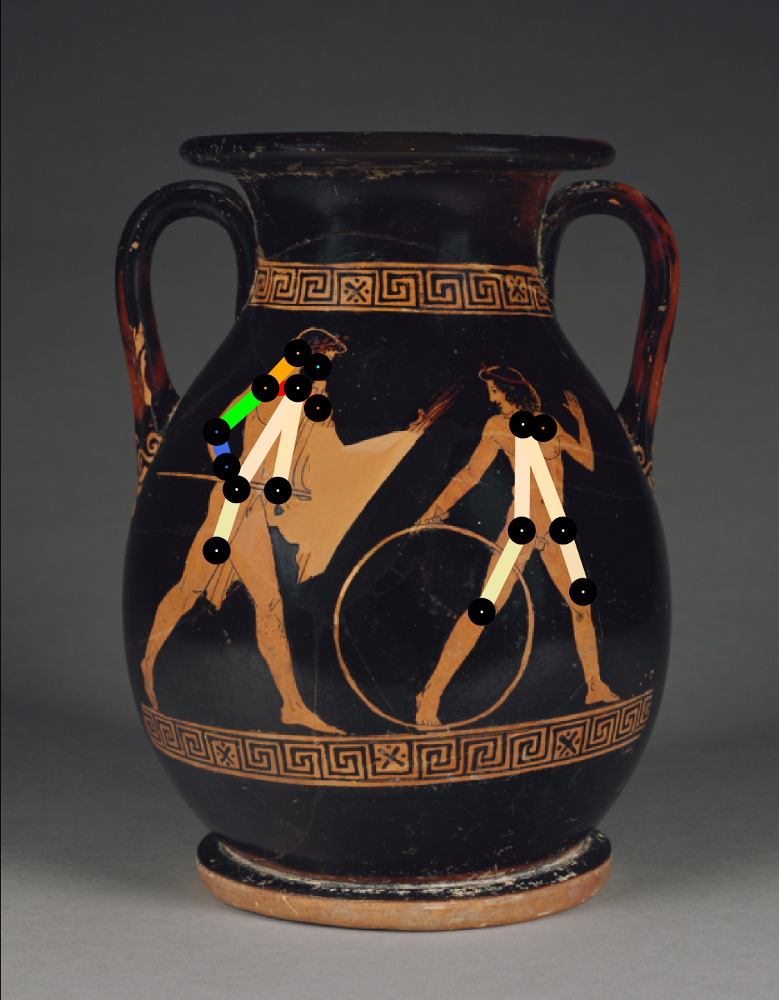}
 		\caption{OpenPose}
 		\label{fig:mainfig2}
 	\end{subfigure}
 	\begin{subfigure}{0.32\linewidth}
 		\centering
 		\includegraphics[width=\textwidth]{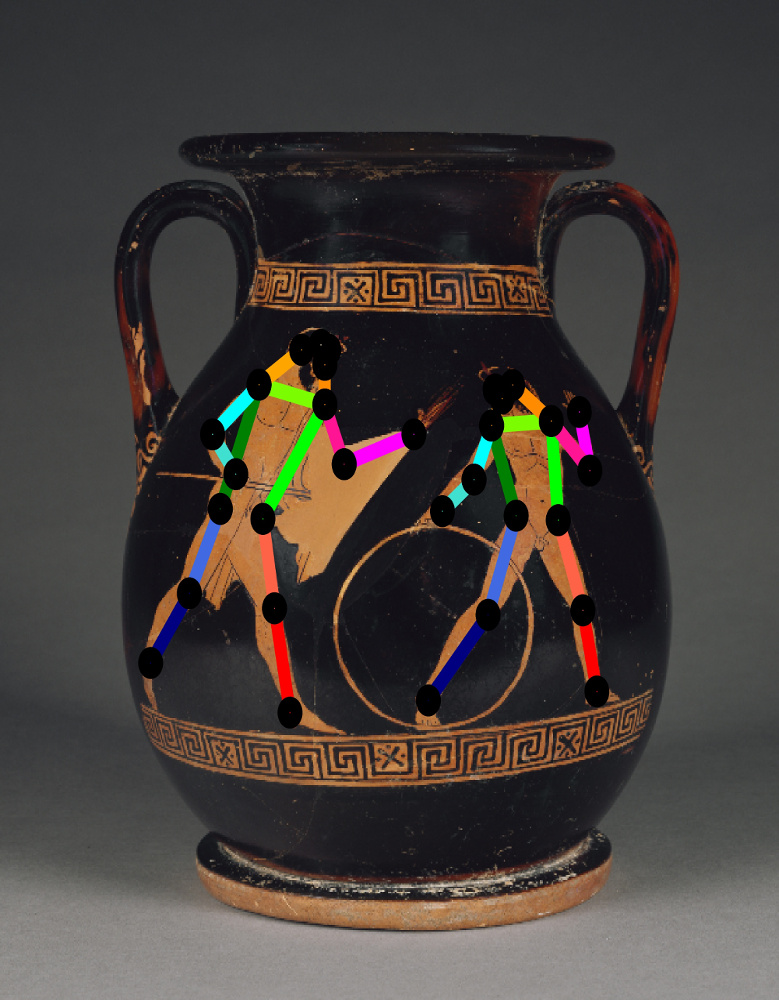}
 		\caption{Ours}
 		\label{fig:mainfig3}
 	\end{subfigure}
 	\caption{Attic red-figure, \textit{Zeus and Ganymede}, in ancient 
 	Greek vase paintings: \subref{fig:mainfig1} original image, 
 	\subref{fig:mainfig2} pose estimation by 
 	OpenPose \citep{Cao_RealTimePoseEstimationAffinityFields_2017}, 
 	\subref{fig:mainfig3} our method.}
 	\label{fig:mainfig}
 \end{figure}

Human pose estimation (HPE) is highly challenging as it is difficult to have 
one method that can generalise across all domains. Estimating human pose 
involves localising each visible body-keypoint (Fig.~\ref{fig:mainfig3}), 
however the state-of-the-art (SOTA) methods underperform when tested on 
different domains, for example ancient Greek vase paintings 
(Fig.~\ref{fig:mainfig}). HPE is central to understanding the visual narration 
and 
body movements of the characters depicted in these paintings and the recent 
rapid digitisation of art collections has created an opportunity to use HPE as 
a tool to digitally examine such artworks. These digital copies are usually 
either photographic reproductions \citep{Mensink_RijksmuseumChallenge_2014} or 
scans of existing archives \citep{Seguin_ArtDigitizationTechniques_2018}. In 
addition to the preservation of cultural heritage, these digital collections 
allow remote access of the invaluable artistic data to the general public. 
However, due to content complexity and large size, navigating within such 
collections is often daunting. 

To address the challenge of efficiently analysing large digital collections, 
several computer vision and image analysis techniques have been used for 
applications, such as artist 
identification \citep{Johnson_ImageProcessingArtistIdentification_2008}, object 
recognition \citep{Cai_RecognisingObjectsInArt_2015, 
Crowley_ObjectRetrievalInPainiting_2014,westlake2016detecting}, character 
recognition \citep{Madhu_RecognizingCharactersInArt_2019}, artistic image 
classification \citep{Carneiro_ArtisticImageClassification_2012, 
Saleh_Elgammal_2016,cetinic2018fine} and 
pose-matching \citep{Jenicek_LinkingArtThroughHumanPoses_2019}. However, when 
these methods are evaluated on a different domain, they show sub-optimal 
performance, c.f. Fig.~\ref{fig:mainfig}. Hence, an important challenge is to learn effective representations using little data. Human pose representation is one such example.

\begin{figure}[t]
 	\begin{subfigure}{0.19\linewidth}
 		\includegraphics[width=\linewidth,height=0.7\linewidth]{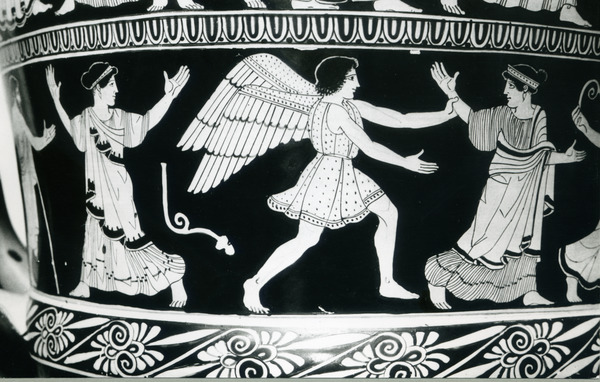}
 	\end{subfigure}
 	\begin{subfigure}{0.19\linewidth}
        \includegraphics[width=\linewidth,height=0.7\linewidth]{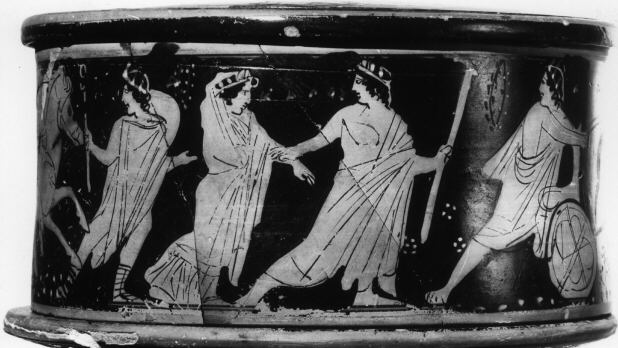}
 	\end{subfigure}
 	\begin{subfigure}{0.19\linewidth}
        \includegraphics[width=\linewidth,height=0.7\linewidth]{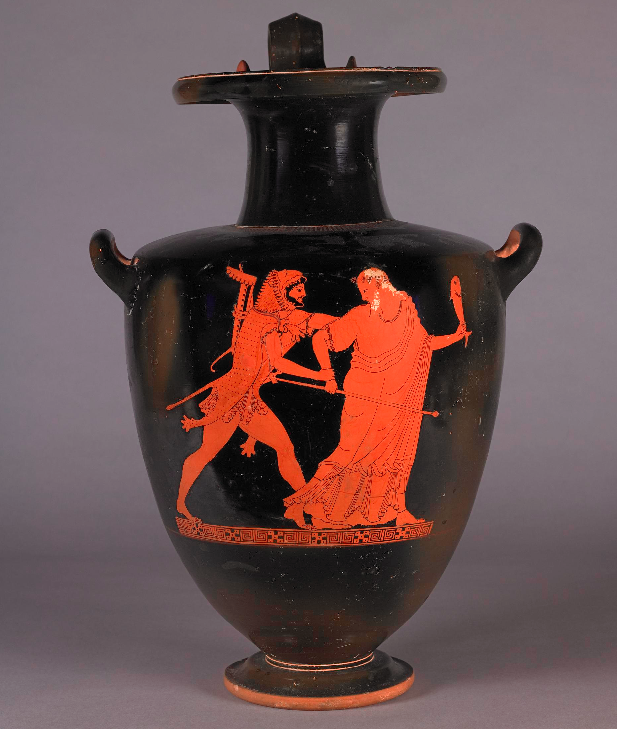}
 	\end{subfigure}
 	\begin{subfigure}{0.19\linewidth}
        \includegraphics[width=\linewidth,height=0.7\linewidth]{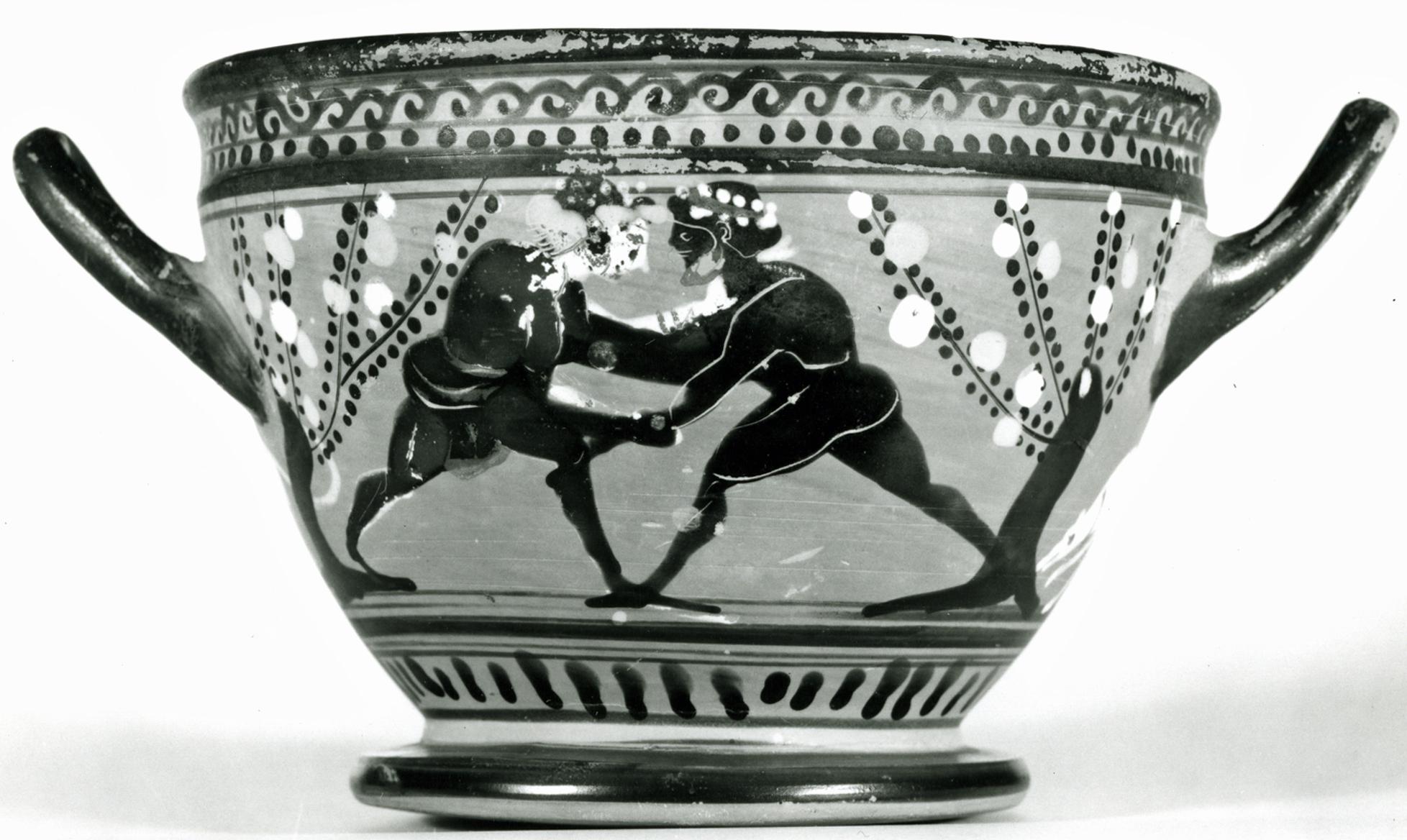}
 	\end{subfigure}
 	\begin{subfigure}{0.19\linewidth}
        \includegraphics[width=\linewidth,height=0.7\linewidth]{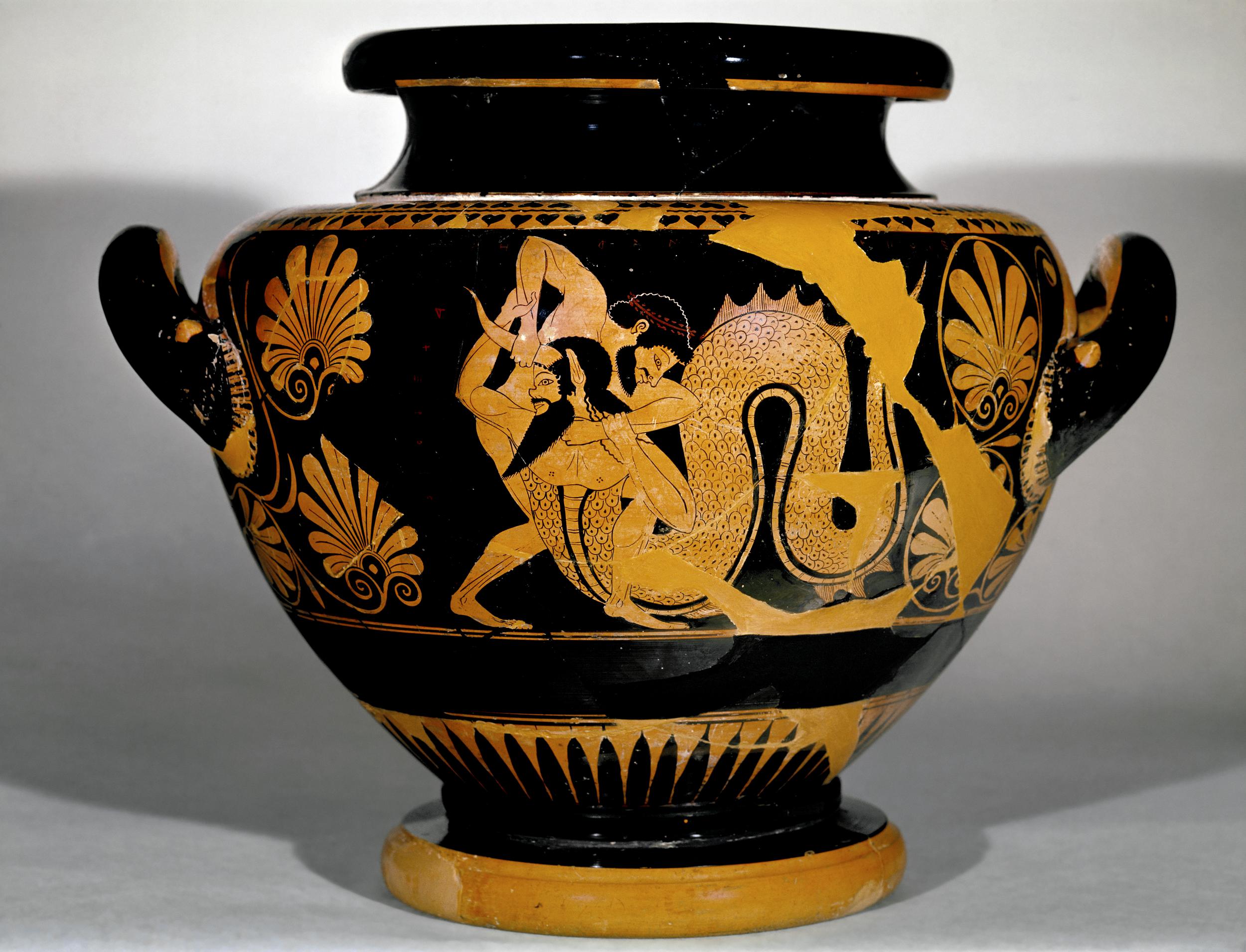}
 	\end{subfigure}

 	\begin{subfigure}{0.19\linewidth}
 		\includegraphics[width=\linewidth,height=0.7\linewidth]{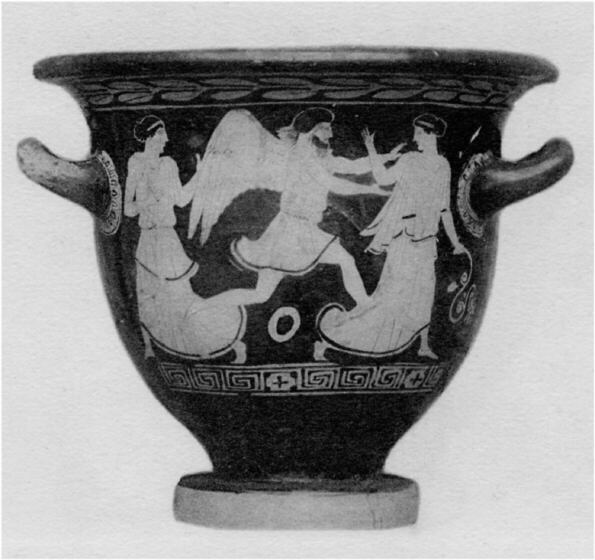}
 	\end{subfigure}
 	\begin{subfigure}{0.19\linewidth}
 		\includegraphics[width=\linewidth,height=0.7\linewidth]{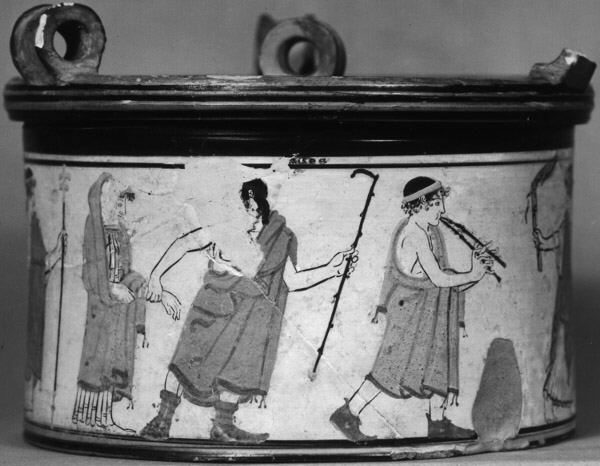}
 	\end{subfigure}
 	\begin{subfigure}{0.19\linewidth}
 		\includegraphics[width=\linewidth,height=0.7\linewidth]{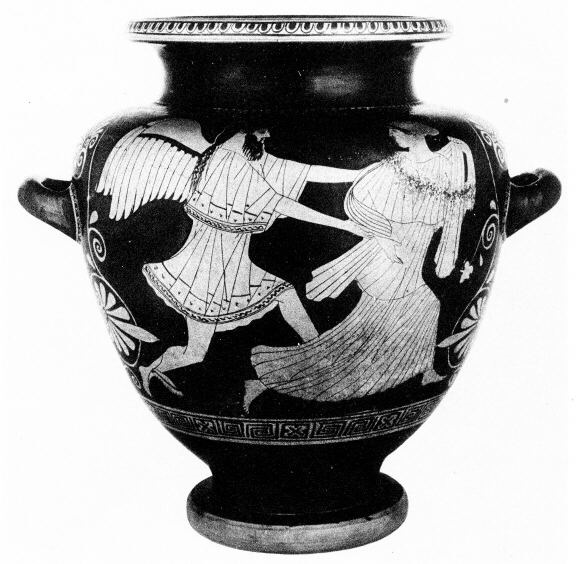}
 	\end{subfigure}
 	\begin{subfigure}{0.19\linewidth}
 		\includegraphics[width=\linewidth,height=0.7\linewidth]{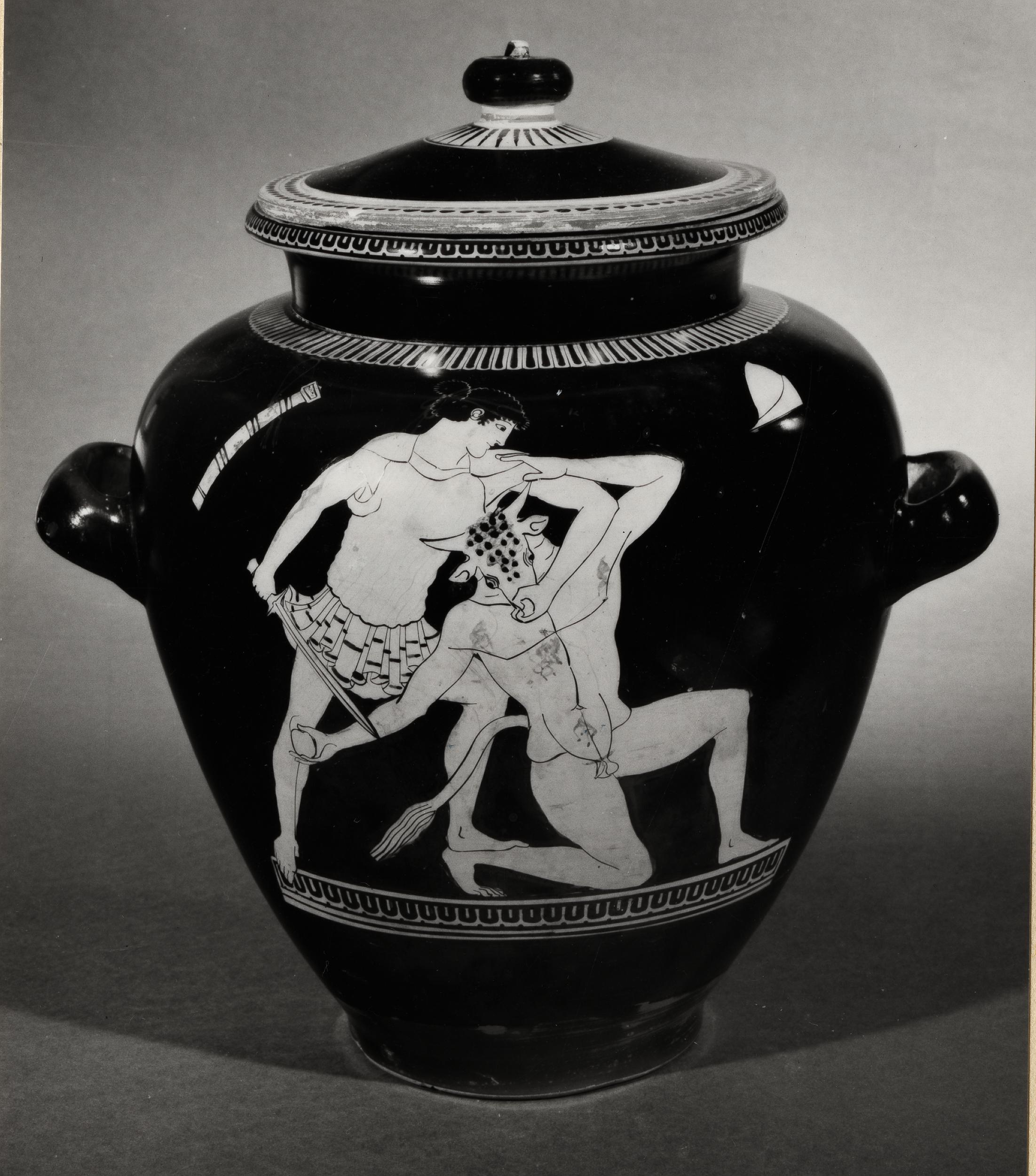}
 	\end{subfigure}
 	\begin{subfigure}{0.19\linewidth}
 		\includegraphics[width=\linewidth,height=0.7\linewidth]{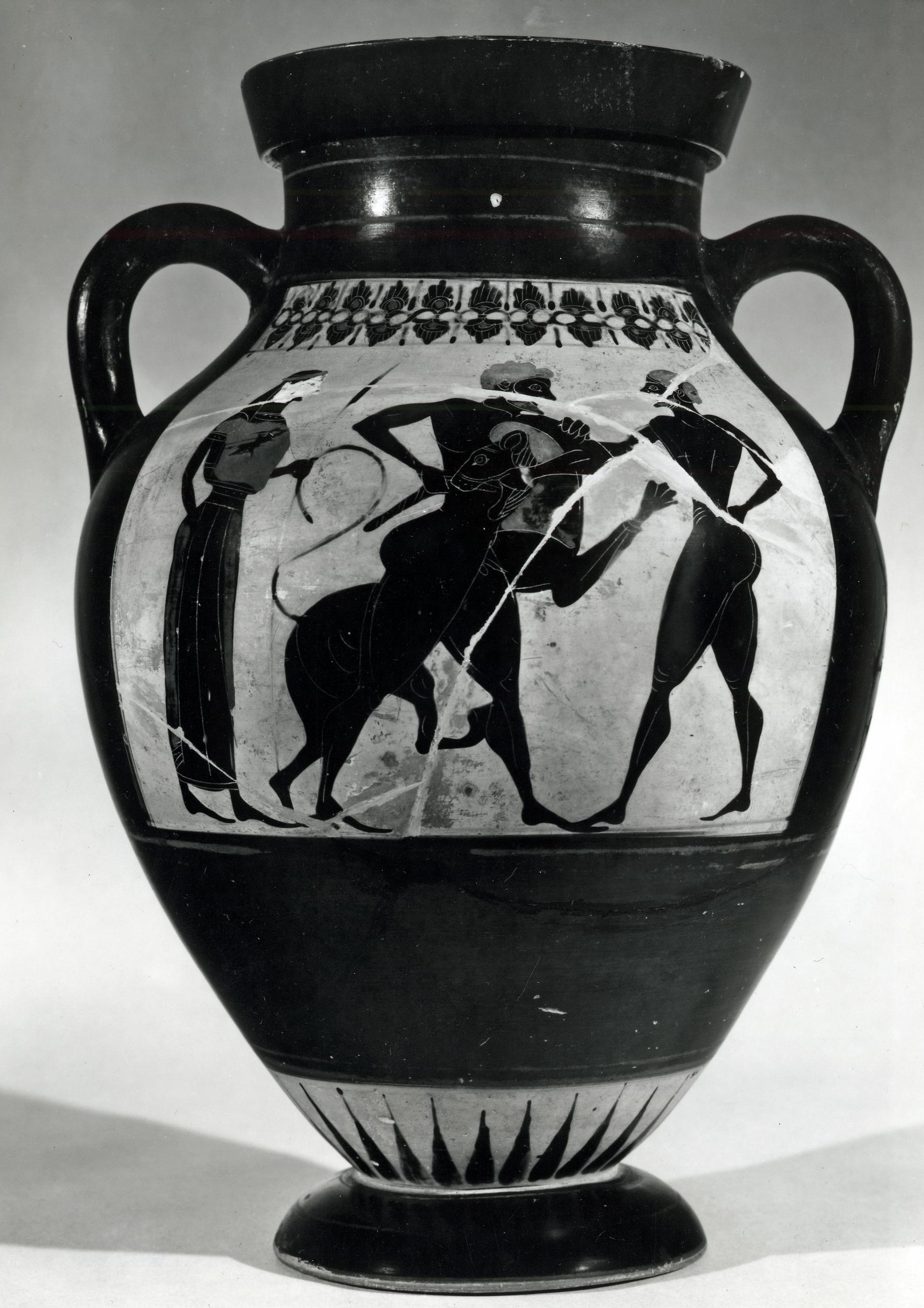}
 	\end{subfigure}
 	
 	\caption{(\textit{\nth{1} column}) Divine pursuit scene in ancient Greek vase paintings. The central character, \textit{a winged persecutor}, is depicted with a similar pose, \ie arms extended towards the right (observer viewpoint) and legs with large strides. (\textit{\nth{2} column}) Leading the bride scene, with the central character \textit{bride} depicted with similar poses with her left 
 	hand extended forward (observer viewpoint) held by the groom. (\textit{\nth{3} column}) Abduction scene, where character on the left is abducting the character on the right. (\textit{\nth{4}~\&~\nth{5} columns}) Wrestling in Agonal and Mythological contexts between two main characters.}
 	\label{fig:pursuits}
 \end{figure}

 The understanding of visual narration in Greek vase paintings is one of the 
 main objectives in the field of Classical Archaeology. In order to display the 
 actions and situations of a narrative, as well as to characterise the 
 protagonists, ancient Greek artists made use of a broad variety of often 
 similar image elements \citep{Giuliani_BildUndMithosGesichteGriechischenKunst_2003}. Some of the key aspects of the narrative are illustrated by meaningful interactions 
 and compositional relationships (\eg postures or gestures) between the 
 characters displayed in the painting \citep{Mcniven_GesturesInAtticVasePaintings_1983}. For example, the divine pursuit scene (Fig.~\ref{fig:pursuits},~\nth{1}~\&~\nth{2} column) is a recurrent narrative in Greek vase paintings, often characterised by a 
 character moving fast from left to right and reaching out with both his arms 
 to catch a woman on her forearm or shoulder \citep{Stansbury_StructuralDifferenciationPursuitScenes_2009}. 

 In this work, we propose to exploit these recurrent character interactions and 
 postures in order to navigate semantically through collections of Greek vase 
 paintings. We address image retrieval in such databases by measuring the 
 similarity between character postures. Since ancient Greek artists made use of 
 the postures to depict similar narratives, the retrieved images should display 
 the same scene.

 For human pose-based image retrieval in Greek vase paintings, we need a 
 reliable human pose estimation algorithm. We propose a two-step approach: (1) 
 First, we apply style-transfer \citep{Huang_ArbitraryStyleTransferAdaptiveInstanceNormalization_2017} to the COCO \citep{Lin_COCODataset_2014} dataset to generate a synthetic annotated dataset with the style of Greek vase paintings and fine-tune the baseline person detection and pose estimation models on this dataset (Fig.~\ref{fig:approach} middle). (2) Second, we fine-tune these models on a newly generated dataset for Classical Archaeology (Fig.~\ref{fig:approach} bottom). We show that both steps improve the person detection and pose estimation tasks and thus the retrieval performance considerably. 

 In particular, our main contributions are: 
 \begin{enumerate}
	
 	\item We introduce the Styled-COCO-Persons~(\textbf{SCP}) and the 
 	ClassArch~(\textbf{CA}) datasets. \textit{SCP} is a synthetic dataset, 
 	generated by applying style transfer, with different style-intensities, to 
 	the images from COCO (only `person' class) to mimic the style of the 
 	\textbf{CA} dataset (Greek vase paintings). The \textbf{CA} dataset 
 	consists of 1783 images (characters) from 1000+ Greek vase paintings along 
 	with pose keypoint annotations. 
	
 	\item We show that by just using styles of the \textbf{CA} dataset on 
 	real images, one can improve the task of human pose estimation in Greek 
 	vase paintings without requiring any annotations. We also show that 
 	fine-tuning this model with the small \textbf{CA} dataset modestly enhances
 	the performance compared to direct transfer learning. Moreover, styled-tuned models outperform state-of-the-art fine-tuned methods on the \textbf{SCP} and \textbf{CA} datasets.
	
 	\item We introduce a perceptual loss for style-transfer and show that this is beneficial for both person detection and pose estimation. 
	
 	\item 
 	Additionally, we show that our styled transfer learning based  pipeline  is  also  
 	beneficial for retrieving and discovering similar images based on poses of the character in narratives from ancient Greek vase paintings.

 \end{enumerate}

  \section{Related work}\label{sec:relwork}
 The task of representing human poses has been studied since the early days of 
 computer vision \citep{Nevatia_DescriptionAndRecognitionOfCurvedObjects_1977}. 
 However, its importance is much older. \textit{Pathosformeln} \citep{pathos}, 
 the iconic study of basic constructs (units) of body language is one of the 
 firsts to view the body gesture (or posture) also as a way of voicing inner 
 emotions. Body movement's depiction is essential and central to the 
 historian \citep{bell2013nonverbal,bell2019ikonographie}, since it gives a way to recognise the 
 inner emotions or expressions of the character. Impett \etal \citep{impett2017totentanz} studies it with a geometrical construct by operationalisation of body movements, which could be considered a way of Pathosformeln. McNiven \etal \citep{Mcniven_GesturesInAtticVasePaintings_1983} also used human poses as the basis to study the interactions between characters in ancient vase paintings. 

 Since several years, Convolutional Neural Networks (CNNs) have been dominating 
 computer vision tasks, and HPE is not an exception. After \textit{DeepPose} \citep{Toshev_DeepPose_2014}, methods followed that improved HPE by using CNN cascades \citep{wei2016convolutional} and graphical models \citep{Tompson_JointModelGraphicalPoseEstimation_2014}. Two major types of approaches are popular with HPE: \emph{bottom-up} and \emph{top-down}.

 \textbf{Bottom-up pose estimation} \citep{Pishchulin_DeepCut_2016, Cao_RealTimePoseEstimationAffinityFields_2017, Insafutdinov_DeeperCut_2016} 
 directly estimate the location of all keypoints and assemble them into pose 
 skeletons for all people in the image simultaneously. They use CNNs like 
 ResNet \citep{He_DeepResidualLearningResNet_2016} and DenseNet \citep{Huang_DenselyConnectedConvolutionalNetworksDenseNet_2017} as 
 backbones to predict keypoints and optimisation-based matching techniques like 
 \textit{DeepCut} \citep{Pishchulin_DeepCut_2016} and  \textit{DeeperCut} \citep{Insafutdinov_DeeperCut_2016} for combining the 
 keypoints into poses. Cao \etal \citep{Cao_RealTimePoseEstimationAffinityFields_2017} introduced \textit{Part Affinity Fields} (PAFs) which is able to estimate poses in 
 real-time by solving a bi-partite graph matching problem, as a way to solve 
 the optimisation problem of aggregating poses. Bottom-up techniques' lack of 
 structural information leads to many false positives and often being 
 outperformed by top-down pose estimation approaches.

 \textbf{Top-down pose estimation} \citep{Papandreou_MultiPersonPoseEstimationInTheWild_2017, 
 Fang_AlphaPose_2017, Chen_CascadedPyramidNetworksPoseEstimation_2018, 
 Xiao_SimpleBaselinePoseEstimation_2018} approach HPE in two steps 
 (Fig.~\ref{fig:approach},~\textit{first row}). The first step addresses the 
 problem of detecting all person instances in the image, whereas the second 
 step aims at predicting the body-keypoints for each of the detected person. 
 For person detection, a specific CNN, \eg from the R-CNN family 
 \citep{Girshick_RCNN_2014, Girshick_FastRCNN_2015, Girshick_FasterRCNN_2015} is 
 normally used. The second step involves using a single-person pose estimation 
 model to process each of the person instances independently. 
 Tompson \etal~\citep{Tompson_JointModelGraphicalPoseEstimation_2014} first 
 proposed the use of a CNN with multi-resolution receptive fields for the task 
 of body joint localisation. More recent methods, however, focus on refinement 
 techniques, such as Iterative Error 
 Feedback \citep{Carreira_HumanPoseEstimationWithIterativeErrorFeedback_2016}, 
 Stacked-hour-glass 
 networks \citep{Newell_StackedHourglassNetworksPoseEstimation_2016} and 
 PoseFix \citep{Moon_2019_CVPR}. Multi-scale approach by 
 Sun \etal \citep{Sun_HighResolutionPoseEstimationHRNet_2019} uses a novel 
 architecture to maintain high-resolution representations through the whole 
 estimation process by repeated multi-scale feature fusions. These SOTA methods 
 have improved the performance on their respective benchmark datasets, however, 
 they fail to generalise to domains like Greek vase paintings. %

 \textbf{Domain adaptation} based approaches mainly aim at bringing the distributions of source domain closer to that of the target, so that a single model can be used for both domains. Methods using domain adaptation via style-transfer mainly transfer the style of target to the source while training online as a way of domain adaption, using style loss \citep{dabmvc} or even adapting the domain progressively \citep{inoue2018cross}. 
 Some methods also use feature level alignment for aligning the two domains \citep{li2017universal}, and others enforce it via self-similarity and domain-dissimilarity loss \citep{deng2018image}.
 Recently, various methods have focused on reducing data bias in order to enhance the transfer-ability of features in task-specific layers \citep{maria2017autodial, long2015learning}. 
 \citep{roy2019unsupervised} and \citep{sun2016deep} proposed unsupervised domain adaptation techniques, using minimum entropy consensus and co-variance of source and target features respectively. 
 
 A few works using generative models have also been proposed. \citep{russo2018source} proposed a method for optimising bi-directional image transformations and using class consistency loss, while \citep{hoffman2018cycada} proposed a cycle GAN that adapts representations to combine feature-level and pixel level while enforcing structural (cycle loss) and semantic consistency (task-specific). 
 A work closely related to ours \citep{Jenicek_LinkingArtThroughHumanPoses_2019}, highlights the importance of using poses for artwork discovery and its subsequent analysis. 
 Their study, however, is based on artworks with some case study images that are relatively easy for SOTA methods, such as OpenPose \citep{Cao_RealTimePoseEstimationAffinityFields_2017} to estimate the poses. One of the first work done on Greek Vase paintings was by Crowley \etal~\citep{crowley2013gods} to generate a correspondence between descriptions and unknown regions in the images of vase paintings.
 In their work, they describe the challenges of working with Greek vases and annotate the images automatically.  
 On a similar note, estimating poses for characters in ancient Greek Vase paintings presents a completely different challenge, since OpenPose fails very often (Fig.~\ref{fig:mainfig}).

 Our work's focus, instead, is on using the style transfer to generate a 
 synthetic dataset from already existing labelled dataset like COCO to improve 
 the pose estimation on unlabelled data like ancient Greek vases, by enforcing 
 a pre-computed perceptual consistency loss. 

 \section{Datasets}\label{sec:datasets}
 In order to train deep networks in a supervised fashion, it is very important 
 to have a high-quality annotated dataset. Hence we work with 3 main datasets 
 \textit{viz}.\ \textbf{COCO-Persons (CP)} with images of only the `person' 
 category, its corresponding styled counterpart called 
 \textbf{Styled-COCO-Persons (SCP)}, and we also introduce our own annotated 
 dataset called \textbf{ClassArch (CA)}. Each dataset is labelled with person 
 bounding boxes and their corresponding body pose keypoints, details shown 
 in Tab.~\ref{tab:datasets}. We focus on these datasets for training and 
 evaluation 
 of our models.

 \begin{table}[t]\centering
 \addtolength{\tabcolsep}{-1.5pt}
 \caption{\textbf{Datasets} used in our experiments. For 
 COCO-Persons~(\emph{CP}), we use images from the person category of 
 COCO \citep{Lin_COCODataset_2014}. Styled-COCO-Persons~(\emph{SCP}) is 
 generated by using \emph{CP} images as \textit{content} and various splits of 
 ClassArch~(\emph{CA}) images as \textit{styles} (\textit{Images:} images, 
 \textit{Persons:} person bounding boxes, \textit{Poses:} pose annotations).}
 \label{tab:datasets}
 \small
 \begin{tabular}{lrrrrrl}\toprule
 \textbf{Datasets$\rightarrow$} &\multicolumn{3}{c}{\textbf{CP \& SCP}} 
 &\multicolumn{3}{c}{\textbf{CA}} \\\cmidrule{1-7}
 \textbf{Split$\rightarrow$} & \textbf{Train} & \textbf{Val} & \textbf{Total} 
 &\textbf{Train} & \textbf{Val} &\textbf{Total} \\\midrule
 \textbf{Images} & 64115 & 2693 &\textbf{66808} & 1210 & 303 &\textbf{1513} \\
 \textbf{Persons} & 257252 & 10777 &\textbf{268029} & 2098 & 531 &\textbf{2629} \\
 \textbf{Poses} & 149813 & 6352 &\textbf{156165} & 1425 & 303 &\textbf{1728} \\
 \bottomrule
 \end{tabular}
 \end{table}

\textbf{(a) COCO-Persons}\emph{~(CP)}\label{subsubsec:cp}
The Common Objects in COntext dataset (COCO) \citep{Lin_COCODataset_2014} was specifically designed for the detection and segmentation of objects in their natural context. COCO has 328K images with over 2.5M labelled instances divided into 91 semantic object categories (\eg car, person, dog, banana, \textit{etc.}). We only consider images that include ``person" instances, along with their corresponding bounding boxes and pose-keypoints. We call this split as \textbf{COCO - Persons} \emph{(CP)}. This split is taken across the training and validation sets only since the labels for the test set are not publicly available. Consequently, we use the validation set for testing our models. Tab.~\ref{tab:datasets} shows the exact splits for the dataset in terms of images, persons and pose-keypoints. Figs.~\ref{fig:dataset_cp}~\&~\ref{fig:dataset_cp_labels} illustrate some samples of \textit{CP} dataset.

 \textbf{(b) ClassArch}\emph{~(CA)}\label{subsubsec:ca}
 We introduce a challenging dataset from the domain of Classical Archaeology, 
 called \textbf{ClassArch}\emph{~(CA)} dataset. We chose five different 
 recurrent narratives, \textit{viz}.\ `Pursuits', `Leading of the Bride', 
 `Abductions', and `Wrestling' in \textit{Agonal} and \textit{Mythological} 
 contexts, taken from the period between the \nth{6} and \nth{5} century BCE. 
 Pose-based analysis of such paintings is of critical importance for Classical 
 Archaeology as discussed in Sec.~\ref{sec:intro}. Each of the narratives in 
 \textit{CA} has its own set of characters, which appear recurrently and are 
 depicted with similar features and in almost identical poses. 
 Figs.~\ref{fig:dataset_ca}~\&~\ref{fig:dataset_ca_labels}  illustrate some 
 images and 
 their 
 corresponding person bounding boxes and person keypoints of the \textit{CA} 
 dataset. Fig.~\ref{fig:pursuits} displays two examples from the `Pursuit' 
 (\nth{1}~\&~\nth{2} column) and `Leading the bride' (bottom row) narratives. 
 In both scenes, the main characters (`persecutor'/`fleeing' \& `bride') are 
 depicted with similar posture in every image. \textit{CA} has different sets 
 of labels associated with it. There are 1513 images, with 2629 person 
 annotations and 1728 pose annotations. More detailed splits are shown 
 in Tab.~\ref{tab:datasets}.

 \textbf{(c) Styled-COCO-Persons (SCP)}\label{subsubsec:styledcoco}
 The images in \textit{CP} significantly differ in semantic content and style 
 from the ancient Greek vase paintings, \textit{c.f.} Fig.~\ref{fig:dataset_cp} 
 vs.~Fig.~\ref{fig:dataset_ca}).
 To bridge this domain gap between \textit{CP} and \textit{CA}, we use style 
 transfer to adapt the style of the CP dataset to vase paintings.

\begin{figure}[tb]
 	\centering
 	\includegraphics[scale=0.4]{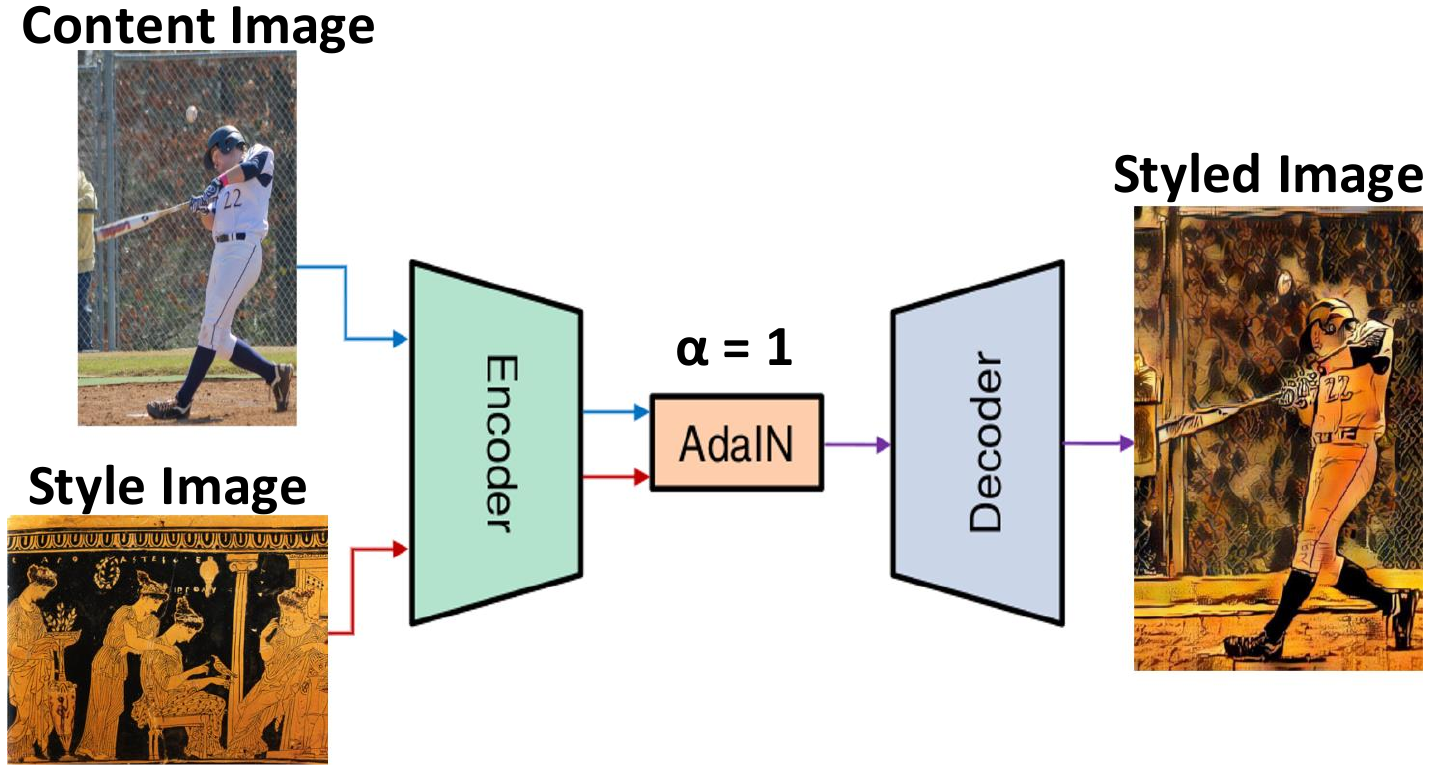}
 	\caption{Style transfer using 
 	\textit{AdaIN} \citep{Huang_ArbitraryStyleTransferAdaptiveInstanceNormalization_2017}
 	 with full style intensity ($\alpha=1$). \textit{AdaIN} adjusts the first 
 	and second order moments of the `Content Image' to match those of the 
 	`Style Image'. A `Styled Image' (style-transferred) is generated with the 
 	semantic content of the `content image' and style of the `Style Image'.}
 	\label{fig:styling_process}
 \end{figure}

\begin{figure*}[t]
 	\begin{subfigure}{0.16\linewidth}
 		\centering
 		\includegraphics[width=\linewidth,height=0.8\linewidth]{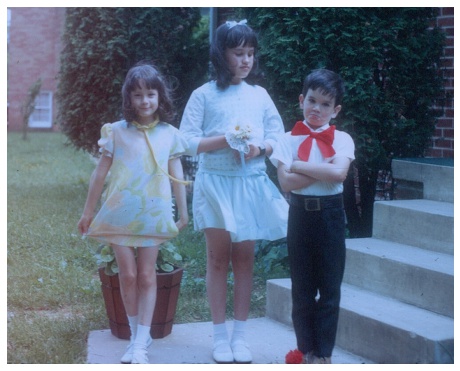}
 	\end{subfigure}
 	\begin{subfigure}{0.16\linewidth}
 		\centering
 		\includegraphics[width=\linewidth,height=0.8\linewidth]{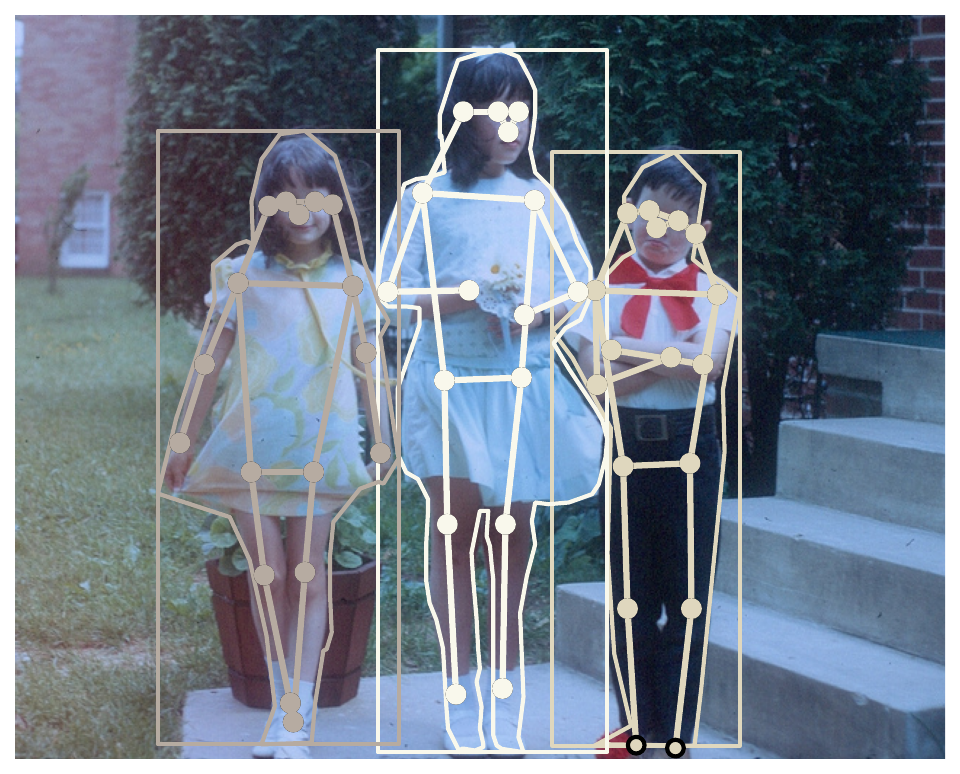}
 	\end{subfigure}
 	\begin{subfigure}{0.16\linewidth}
 		\centering
 		\includegraphics[width=\linewidth,height=0.8\linewidth]{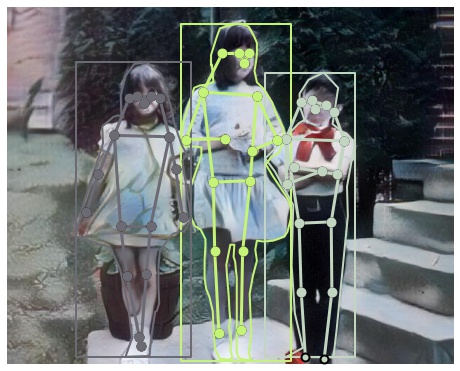}
 	\end{subfigure}
 	\begin{subfigure}{0.16\linewidth}
 		\centering
 		\includegraphics[width=\linewidth,height=0.8\linewidth]{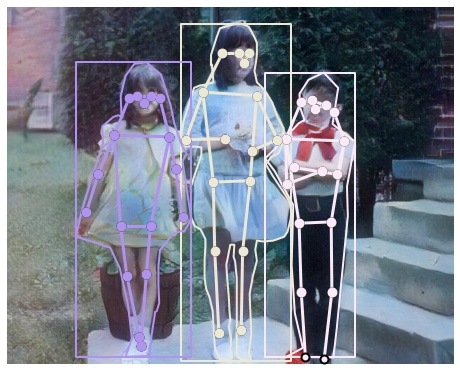}
 	\end{subfigure}
 	\begin{subfigure}{0.16\linewidth}
 		\centering
 		\includegraphics[width=\linewidth,height=0.8\linewidth]{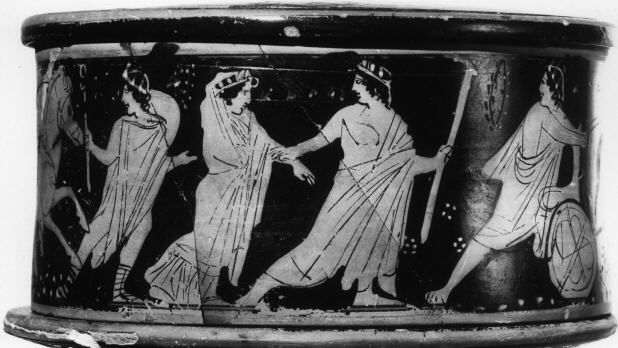}
 	\end{subfigure}
 	\begin{subfigure}{0.16\linewidth}
 		\centering
 		\includegraphics[width=\linewidth,height=0.8\linewidth]{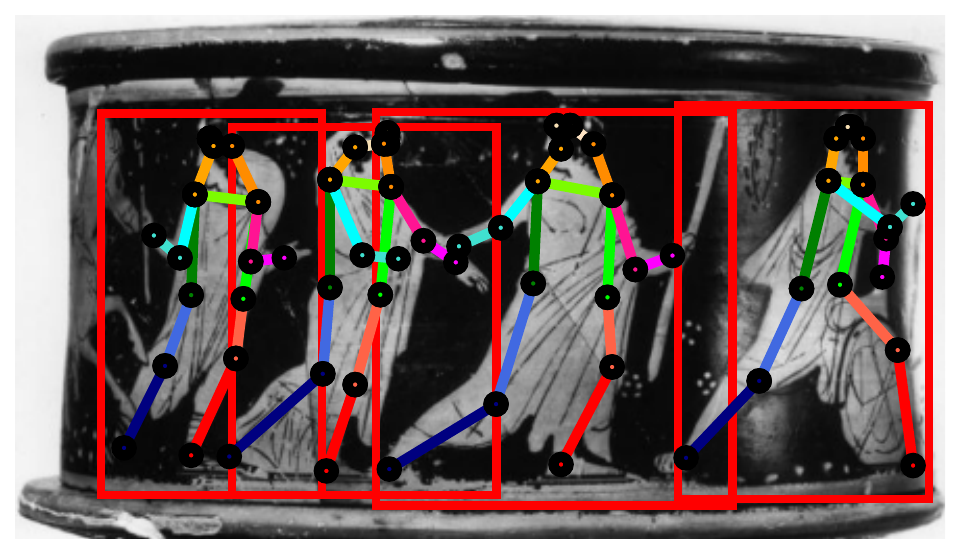}
 	\end{subfigure}%

 	\begin{subfigure}{0.16\linewidth}
 		\centering
 		\includegraphics[width=\linewidth,height=0.8\linewidth]{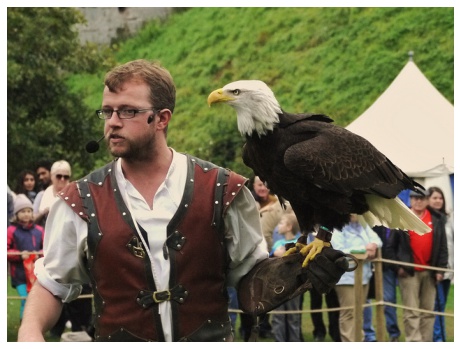}
 		\caption{\emph{CP} Images}
 		\label{fig:dataset_cp}
 	\end{subfigure}
 	\begin{subfigure}{0.16\linewidth}
 		\centering
 		\includegraphics[width=\linewidth,height=0.8\linewidth]{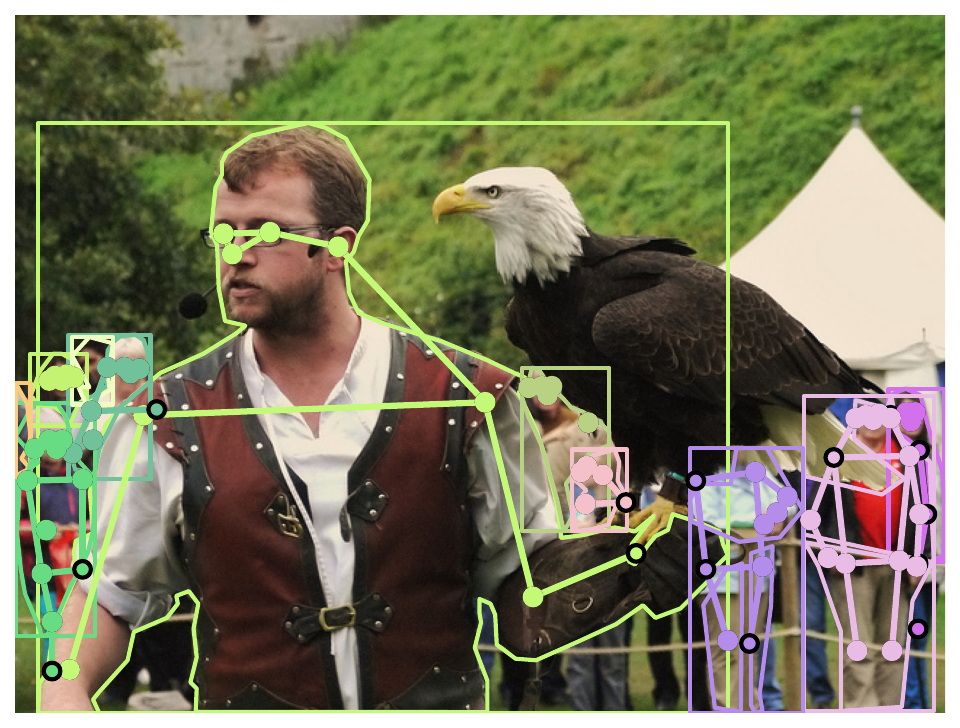}
 		\caption{\emph{CP} Labels}
 		\label{fig:dataset_cp_labels}
 	\end{subfigure}
 	\begin{subfigure}{0.16\linewidth}
 		\centering
 		\includegraphics[width=\linewidth,height=0.8\linewidth]{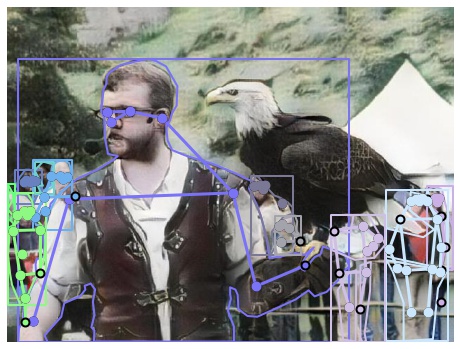}
 		\caption{\emph{SCP}, $\alpha=0.5$}
 		\label{fig:dataset_scp_alpha05}
 	\end{subfigure}
 	\begin{subfigure}{0.16\linewidth}
 		\centering
 		\includegraphics[width=\linewidth,height=0.8\linewidth]{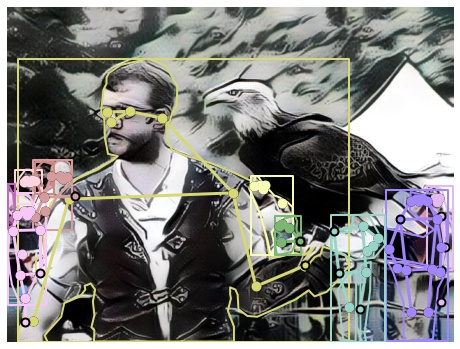}
 		\caption{\emph{SCP}, $\alpha=U$}
 		\label{fig:dataset_scp_alphau}
 	\end{subfigure}
 	\begin{subfigure}{0.16\linewidth}
 		\centering
 		\includegraphics[width=\linewidth,height=0.8\linewidth]{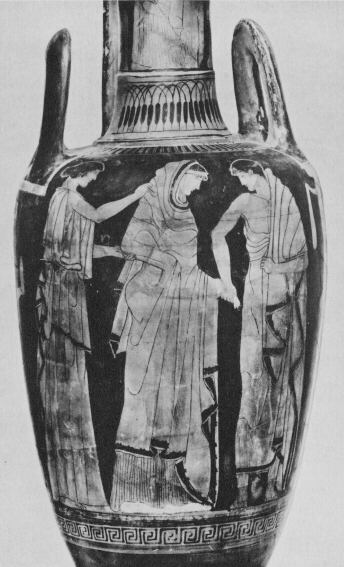}
 		\caption{\emph{CA} Images}
 		\label{fig:dataset_ca}
 	\end{subfigure}
 	\begin{subfigure}{0.16\linewidth}
 		\centering
 		\includegraphics[width=\linewidth,height=0.8\linewidth]{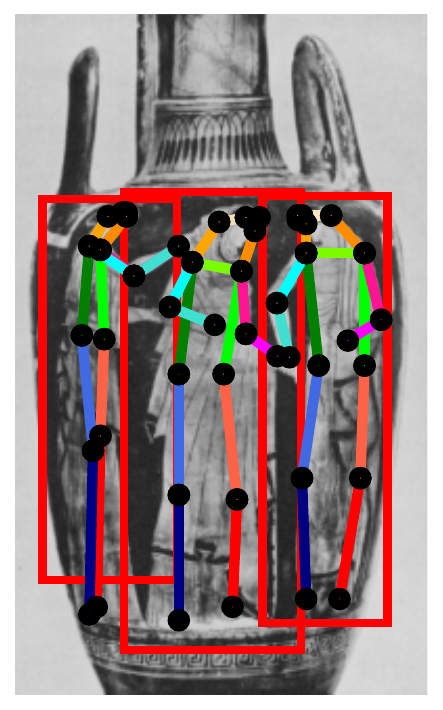}
 		\caption{\emph{CA} Labels}
 		\label{fig:dataset_ca_labels}
 	\end{subfigure}%
 	\caption{\textbf{Dataset Samples}:\subref{fig:dataset_cp} Images \&  
 	\subref{fig:dataset_cp_labels} Labels of \emph{CP} dataset; 
 	\subref{fig:dataset_scp_alpha05} \&  \subref{fig:dataset_scp_alphau} are 
 	samples from the \emph{SCP} dataset with $\alpha=0.5$ and  $\alpha=U$ 
 	respectively; \subref{fig:dataset_ca} shows images with 
 	\subref{fig:dataset_ca_labels} the corresponding labels of our \emph{CA} 
 	dataset. Each labelled example shows the corresponding person bounding 
 	boxes and their pose keypoints.}
 	\label{fig:dataset}
 \end{figure*}

 Style transfer algorithms render a synthetic image that combines the semantic 
 information from one input (denoted as \textit{content image}) with the 
 texture from the user-defined \textit{style 
 image} \citep{Gatys_ImageStyleTransferUsingCNNs_2016}. We apply an efficient 
 and fast style transfer technique using adaptive instance normalisation 
 (\textit{AdaIN}) \citep{Huang_ArbitraryStyleTransferAdaptiveInstanceNormalization_2017}
  to create \textit{SCP}, a synthetic dataset that combines the semantic 
 content of the \textit{CP} with the style of \textit{CA}. 
 Fig.~\ref{fig:styling_process} illustrates the style transfer procedure. We 
 can 
 visually observe that images of 
 Figs.~\ref{fig:dataset_scp_alpha05}~\&~\ref{fig:dataset_scp_alphau} are more 
 closer in 
 styles with Fig.~\ref{fig:dataset_ca} than Fig.~\ref{fig:dataset_cp}, \ie  
 \textit{SCP} 
 is closer in style with \textit{CA}, than \textit{CP} is with \textit{CA}.
 The \textit{SCP} dataset will be released along with the code.

 \textbf{Alpha $(\alpha)$ and Style-Sets}\label{subsubsec:ss}
 Huang~\etal \citep{Huang_ArbitraryStyleTransferAdaptiveInstanceNormalization_2017}
  suggest a content-style trade-off technique to control the intensity of style 
 transferred to the content image using $\alpha\in[0,1]$. Based on this, we 
 generate 2 groups of \textit{SCP}. First with $\alpha=0.5$, meaning that we 
 only transfer half of the style intensity to the content images; and a second 
 one in which $\alpha$ is chosen randomly from the uniform distribution 
 $U[0,1]$. The second group contains images across the whole spectrum, from no 
 style ($\alpha=0$) to full style ($\alpha=1$).

 Additionally, we generate two more \textit{SCP} dataset variations with the method described above using a different dataset of style images. We name this dataset as \textbf{Red-Black} figures 
 (100 in total) or just \textbf{RB}, they are similar in style to 
 the \textit{CA} dataset but do not have any labels. Our hypothesis is that the 
 model should be able to learn the styles and not the content of style images. 
 In the end, we have four groups of \textit{SCP} dataset with two different 
 combinations of $\alpha$ with the two style-sets \textit{RB} and \textit{CA}.

 \textbf{Open Source Images (OSI)} We will publicly release the \textbf{CP} and \textbf{SCP} datasets + annotations. 71 sample OSI links from the \textbf{CA} dataset are in the supplementary material, the rest can be downloaded from Beazley Archive Pottery Database\footnote{\url{https://www.beazley.ox.ac.uk/pottery/default.htm}} for research purposes. We'll release the permanent links to those images, along with the pose and person annotations.

\section{Proposed Method}\label{sec:propmethod}
 In this section, we first present our proposed style-based transfer learning approach to enhance 
 pose estimation. We then briefly explain our models that were trained and evaluated for all datasets mentioned in \cref{sec:datasets}. Lastly, we propose a perceptual loss as a regularizer to improve the estimation of perceptually similar poses with different styles.
 
\subsection{Pose Estimation Approach}\label{subsec:approach}
\begin{figure}[t]
 		\centering
 		\includegraphics[scale=0.5]{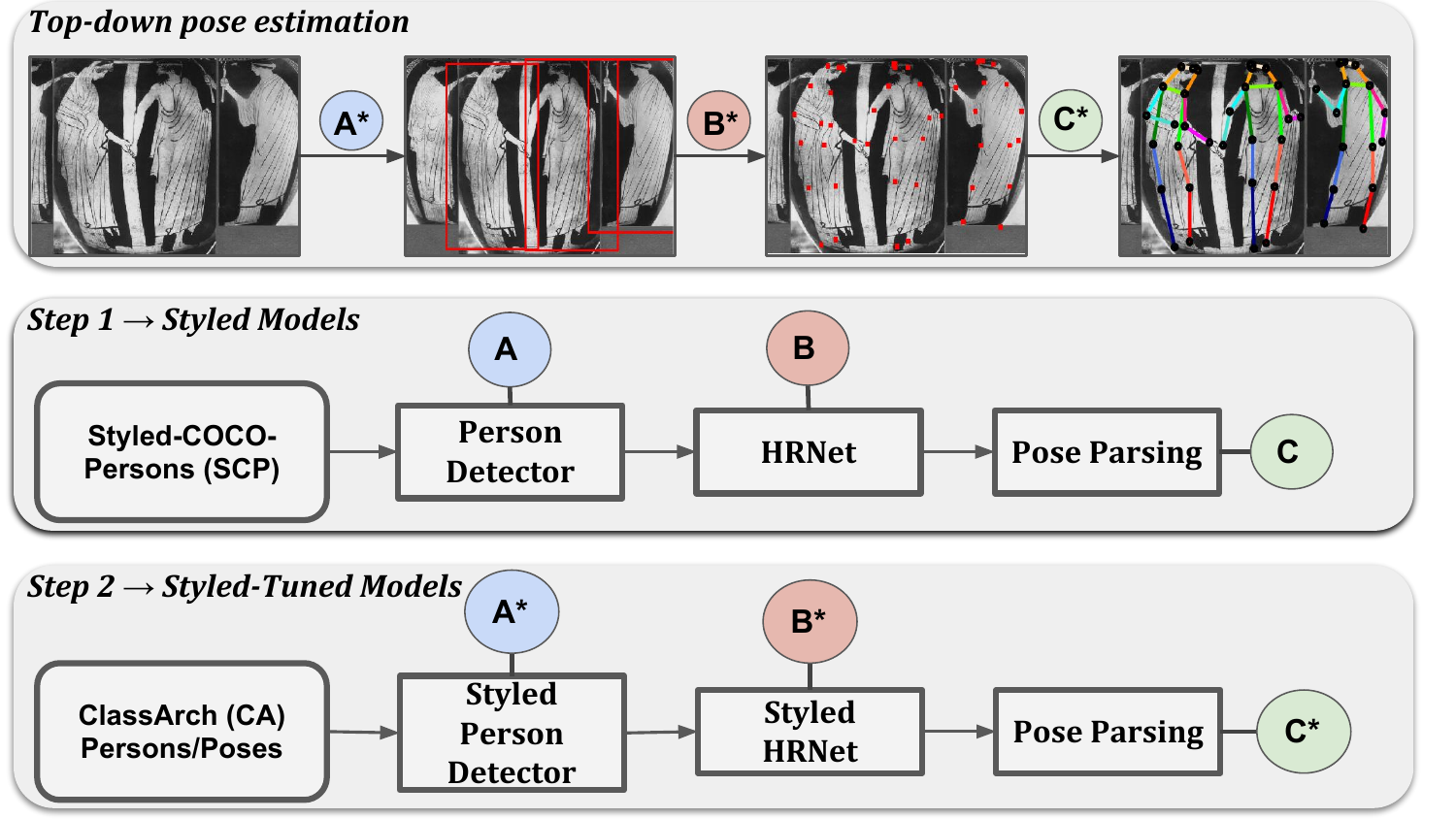}
 	\caption{(\textit{first row, Top-down pose estimation}) - 
 	(\textbf{A*}) styled person detector detects all instances, (\textbf{B*}) 
 	for which the body joint locations are predicted  using a person keypoint 
 	detector, (\textbf{C*}) The pose skeletons are assembled by connecting the 
 	detected keypoints for each person. \textbf{2 Step Training Approach:} 
 	\textbf{Step 1}~(\textit{second row, Styled Models}) Person Detector 
 	trained on \textit{SCP} persons data, and HRNet on \textit{SCP} poses data; 
 	\textbf{Step 2}~(\textit{third row, Styled-Tuned Models}) Styled Person 
 	Detector from \textit{second row} is fine-tuned on \textit{CA} persons 
 	data, and Styled HRNet is fine-tuned on \textit{CA} pose data.}
 	\label{fig:approach}
 \end{figure}

 We take a top-down approach to pose-estimation, which is divided into two 
 stages. Fig.~\ref{fig:approach} details the two stages of the top-down 
 approach. 
 The first stage (\textit{A/A*}) detects all the persons in an image and then 
 estimates the keypoints (\textit{B/B*}) for each person instance, then 
 creating the poses for each instance by pose-parsing (\textit{C/C*}). The models without * are trained on styled datasets while the ones with * are fine-tuned on the \textit{CA} dataset.
 We use 
 Faster-RCNN \citep{Girshick_FasterRCNN_2015} as our person detector that was 
 trained on the COCO \citep{Lin_COCODataset_2014} dataset. Top-down pose 
 estimation approaches dominate the COCO keypoint detection challenge in the 
 past few years, and several use the 
 HRNet \citep{Sun_HighResolutionPoseEstimationHRNet_2019} as their  backbones. 
 Hence, we chose HRNet-W32 (henceforth denoted as \textit{HRNet}) as our pose 
 estimation model.

 \paragraph{\textbf{2-Step Training Approach}}\label{subsubsec:2st}
 We adopt a 2 step approach to enhancing pose estimation, as shown 
 in Fig.~\ref{fig:approach}. In the first step (Fig.~\ref{fig:approach}, 
 \textit{second 
 row}), we train our detector and pose estimation model on styled data (different groups of \textit{SCP}) 
 to generate \textit{styled} models (A and B in \cref{fig:approach}). In this step, the \textit{styled} models 
 at the end of their training are expected to learn the styles of the target 
 data, while trying to maintain the performance on the original task. 

 In the second step (Fig.~\ref{fig:approach}, \textit{third row}), we fine-tune 
 these \textit{styled} models on our \textit{CA} data. During this step, the 
 models that have learned the styles in the first step, now focus on improving 
 their performance for the target dataset of \textit{CA}. 
 The final detector (A*) and pose estimator (B*) models are initialized with the A and B models from the first step and then fine-tuned on \textit{SCP} persons and poses data.

 We report all our experiments using four kinds of models for both tasks, 
 person detection and pose estimation. 

 \textbf{1.~Baseline models} are SOTA models. In case of the person detector, 
 we drop the heads for all the classes except one, and fine-tune it on the 
 `person' class, further denoted as our \textit{baseline} model. 

 \textbf{2.~Tuned models} are SOTA models fine-tuned on the \textit{CA} 
 dataset. For the detector, we drop all the heads except one (similar as for 
 baseline model) in Faster-RCNN and fine-tune it on persons data of the 
 \textit{CA} dataset. Likewise, for pose estimation, we take the SOTA HRNet and 
 fine-tune it on pose data of \textit{CA}.  

 \textbf{3.~Styled models} are SOTA models trained on a particular group of the 
 \textit{SCP} dataset. As explained in Sec.~\ref{subsubsec:ss}, there are 4 
 different groups of \textit{SCP} dataset. Depending on the values of $\alpha$ 
 and \textit{style-set} (RB or CA), the \textit{Styled} models are trained on 
 that particular group, for the detector as well as the pose estimator.  
 Accordingly, there are 4 different \textit{Styled} models for each of the 
 detector and the pose estimator. 

 \textbf{4.~Styled-Tuned (Sd$\rightarrow$Td) models} are \textit{Styled} models 
 ((3) above) fine-tuned on \textit{CA} dataset. Accordingly, for the detector, 
 \textit{Sd$\rightarrow$Td} model is a \textit{Styled} Faster-RCNN model 
 fine-tuned on \textit{CA} persons data. Similarly, for the pose estimator, 
 \textit{Sd$\rightarrow$Td} model is a \textit{Styled} HRNet model fine-tuned 
 on \textit{CA} poses data. Hence, depending on the group of the 
 \textit{Styled} model, there is an equivalent \textit{Sd$\rightarrow$Td} model.

 \subsection{Enforcing Perceptual Similarity} While training the \emph{styled}
 models, the network is fed with styled data. The advantage of doing this is to 
 allow the model to expand its capacity to recognise perceptually similar 
 persons/poses with different styles. In order to achieve content consistency 
 in the perceptual space, we enforce a pre-computed perceptual 
 loss \citep{johnsonPerceptualLossesRealTime2016} while training, in addition to 
 the regular loss. The model is penalised if it is not able to maintain 
 perceptual consistency. 

 Let's denote the task loss by $L_T$, where $L_T=L_{det}$ is for detector models, and $L_T=L_{pose}$ for pose models. 
 We adopt two flavours of the combined loss, each for the detector as well as 
 pose. In the first flavour~($L_{comb1}$), we adaptively weigh the perceptual 
 loss~($L_{percept}$) with the corresponding detector or the 
 pose loss, as shown in Eq.~\ref{eq:perceptual1}.
 While, in the second flavour~($L_{comb2}$), we weigh each loss term with optimal values of 
 $\lambda_{1}$ and $\lambda_{2}$ chosen using hyperparameter optimisation, as 
 shown in Eq.~\ref{eq:perceptual2}. 
 \begin{subequations}
    \begin{align}
 	L_{comb1} & = L_{T} + L_{T}\ast L_{percept} \label{eq:perceptual1} \\
    L_{comb2} &= \lambda_{1}\ast L_{T} + \lambda_{2}\ast L_{percept} 
            \label{eq:perceptual2}
    \end{align}
\end{subequations}

 \section{Experiments and Analysis}\label{sec:experiments}
 The exact number of images, person bounding boxes and pose annotations, along 
 with the corresponding train/val splits used for our experiments are mentioned 
 in Tab.~\ref{tab:datasets}. In this section, we describe the evaluation 
 protocol 
 to train our detector and pose estimator. We also present the experimental 
 results and discuss our findings.

 \subsection{Training Setup}
 In general, we use the standard parameters of the SOTA models and make 
 adjustments to suit our experimental needs. For person detection 
 (Faster-RCNN), we use an initial learning rate (\textit{lr$_{init}$}) of 
 0.0001 with a scheduler \textit{lr$_{scheduler}$} on plateau (3 epochs) which 
 reduces the \textit{lr} by a factor of 0.33. We use Adam \citep{adam} with its 
 default parameters and a batch-size~(\textit{bs}) of 8. Standard multi-task 
 loss metric, a combination of log loss and regression loss 
 ($L_{det}=L_{CLS}+L_{Reg}$), is used in our experiments for the detector. We 
 train for 25 epochs on the \textit{CP} dataset and 30 each on the \textit{SCP} 
 and \textit{CA} datasets. However, we found that our models usually converge 
 between 8--12 epochs.\\  
 Similar to the detector, we use Adam with its default values for pose models 
 (HRNet) and \textit{bs}$=$64. With an \textit{lr$_{init}$} of 0.01 and a 
 \textit{lr$_{scheduler}$} on plateau (3 epochs) which reduces the \textit{lr} 
 by a factor of 0.1. We train all pose models for 100 epochs. Akin to the 
 original HRNet \citep{Sun_HighResolutionPoseEstimationHRNet_2019}, we also use 
 the same configs (augmentations, image size) for fair comparison. Like HRNet, 
 Object Keypoint Similarity (OKS) is used as an evaluation metric in our 
 experiments as a simple Euclidean distance ($L_{pose}=L_{MSE}$) for pose 
 estimation. 

 In both cases, person detection and pose estimation, we report the mean 
 Average Precision (\textit{mAP}) as well as the corresponding mean Average 
 Recall (\textit{mAR}). 

 \subsection{Experiments}
 We compare different models in separate tables to give a clear understanding 
 of our methods. 
 As described in \cref{sec:datasets}, \textbf{Style-set}~(SS) represents two datasets \emph{CA} and \emph{RB} and \textbf{alpha}~($\alpha$) represents the amount style transferred in the \emph{SCP} dataset from no style ($\alpha$ = 0) to full style ($\alpha$ = 1).
 Tab.~\ref{tab:resultA}~(Results A) compares the 
 \textit{baseline} 
 models with \textit{styled} models for detector as well as pose. Similarly, 
 Tab.~\ref{tab:resultB}~(Results B) compares the \textit{tuned} with 
 \textit{styled-tuned}. Tab.~\ref{tab:resultC}~(Results C) shows the influence 
 of 
 using different data quantities to fine-tune our models, where as 
 Tab.~\ref{tab:perceptual} shows the advantage of using perceptual loss. 

 \paragraph{\textbf{Baseline \textit{vs}\ Styled models}} (Results A, 
 Tab.~\ref{tab:resultA}) It is important to understand the impact of styles on 
 the 
 main task for detection and pose estimation. We study the impact of 
 style-transfer by comparing \textit{baseline} and \textit{styled} models. As 
 shown in Tab.~\ref{tab:resultA}~(\emph{SCP} column), we observe that the 
 styled 
 models perform consistently much better than their baseline counterpart for 
 detection and pose estimation. When tested on the \emph{CA} dataset, counter-intuitively, these models underperform in detection. 
 One potential reason is that the network has never seen the complex vase dataset during training.
 Conversely for pose estimation, styled 
 models unambiguously are better for both \emph{SCP} and \emph{CA} datasets. 
 Specifically, styled models, which were not trained on \emph{CA}, give a 
 considerable jump in performance: 7.62 (mAP)~\&~8.06 (mAR) when tested on 
 \emph{CA}.

\begin{table}[t]
    \centering
    \caption{\textbf{Results A}: Comparing \textit{baseline} model and \textit{styled} models with different combinations of $\alpha$ (stylization factor) and \textit{SS} (\textbf{RB} or \textbf{CA}), \subref{tab:resultApose} for pose estimator and \subref{tab:resultAdetect} person detector. $\alpha=0.5$ or $\alpha=U$ (randomly sampled from a uniform distribution: $U=uniform(0,1)$). All values in terms of \textit{mAP}, except 
 	\emph{CA}{\tiny\textit{mAR}} (\textit{mAR}). The (+/-) is in reference to the corresponding baselines.}
 	\label{tab:resultA}
 	\centering
 	\footnotesize
 	\begin{subtable}{0.5\linewidth}
    \centering
        \begin{tabular}{lllllcc}\toprule
    	\textbf{Model} &\textbf{CP} &\textbf{SCP} &\textbf{CA} & 
 		\textbf{CA}\tiny{\textit{mAR}} &\textbf{$\alpha$} &\textbf{SS} 
 		\\\midrule
 		\textbf{Baseline} &\textbf{76.5} &46.2 &24.7 & 30.9 & - & - 
 		\\\cmidrule{2-7}
 		\multirow{4}{*}{\textbf{Styled}} &73.4 (-3.1) &\textbf{54.4 (+8.2)} &29.7 (+5.0) & 36.0 
 		&0.5 &RB\\
 		&74.0 (-2.5) &53.7 (+7.5) & 30.9 (+6.2) & 37.7 & $U$ &RB\\
 		&74.0 (-2.5) &53.8 (+7.6) & 30.6 (+5.9)  & 36.9 &0.5 &CA \\
 		& 74.3 (-2.2) & 53.5 (+7.3) & \textbf{32.3 (+7.6)} & \textbf{39.0} &$U$ &CA \\
 		\bottomrule
 		\end{tabular}
    \caption{Pose Estimation}
    \label{tab:resultApose}
    \end{subtable}
    \begin{subtable}{0.5\linewidth}   	
    \centering

    	\begin{tabular}{lllllcc}\toprule		
     		\textbf{Model} &\textbf{CP} &\textbf{SCP} &\textbf{CA} & 
     		\textbf{CA}\tiny{\textit{mAR}} &\textbf{$\alpha$} &\textbf{SS} 
     		\\\midrule
     		\textbf{Baseline} &\textbf{39.4} &24.2 
     		&\textbf{10.4} & 9.8 & - & - \\\cmidrule{2-7}
     		 &37.5 (-1.9) & \textbf{33.4 (+9.2)} & 7.6 (-2.2) & 9.8 &0.5 &RB\\
     		 &36.9 (-2.5)& 32.1 (+7.9) & 6.5 (-3.3) & 8.5 & $U$ &RB\\
     		 &37.7 (-1.7)& 33.2 (+9.0) & 8.2 (-1.6) & \textbf{10.0} &0.5 &CA \\
     		\multirow{-4}{*}{\textbf{Styled}}& 37.0 (-2.4) & 32.6 (+8.4) 
     		& 6.5 (-3.3) & 8.6 &$U$ &CA \\
 		\bottomrule
 	    \end{tabular}
    \caption{Person Detection}
    \label{tab:resultAdetect}
    \end{subtable}
    
\end{table}

 \paragraph{\textbf{Tuned \textit{vs}.\ Styled-Tuned models}} (Results B, 
 Tab.~\ref{tab:resultB}) With the goal of enhancing pose estimation on our 
 \emph{CA} dataset, a naive approach is to fine-tune on this data, we call 
 these models as \textit{Tuned models}. Then, we take the styled 
 models (Tab.~\ref{tab:resultA}), which have already learned the styles of 
 \emph{CA} data, and fine-tune them on our \emph{CA} data 
 (\textit{Styled-Tuned} or \textit{Sd$\rightarrow$Td}). As seen in 
 Tab.~\ref{tab:resultB}, the \textit{Sd$\rightarrow$Td} models give a better 
 performance as compared to their \textit{Tuned} counterparts for pose 
 estimation. Irrespective of the combination of $\alpha$ and \textit{SS}, the 
 pose models tend to perform better. We argue that this is partly because the 
 models gradually learn the styles (\textit{SCP}), while optimising for the 
 main task. During training, the \textit{Styled} 
 models~(Tab.~\ref{tab:resultA}) 
 are able to see the different spectrum of style intensities. They adapt the 
 styles while maintaining a consistent performance over the main task. 
 However, for person detection the performance of \textit{Sd$\rightarrow$Td} model is detrimental in comparison to the \textit{Tuned} counterpart.
One reason for this can be attributed to the overlapping objects: animals and persons -- which makes the person detection more difficult for the \textit{Sd$\rightarrow$Td} models as compared to their \textit{Tuned} counter parts. 
Another reason for lower precision is the lack of ground truth annotations for side characters of the scene.
For poses however, the overlap of keypoints when compared to the bounding boxes is very small and hence the model generalizes better from styled models in comparison to directly tuned models.\\

 \begin{table}[t]

 	\caption{\textbf{Results B}: Comparing \textit{tuned} model with the 
 	\textit{styled-tuned~(\emph{Sd}$\rightarrow$\emph{Td})} model, with 
 	different combinations of $\alpha$ (stylization factor) and \textit{SS} (\textbf{RB} or \textbf{CA}), for \subref{tab:resultsBpose}
 	pose estimator \subref{tab:resultsBdetect} and detector. $\alpha=0.5$ or 
 	$\alpha=U$ (randomly sampled from a uniform distribution: 
 	$U=uniform(0,1)$). All values in \textit{mAP}, except 
 	\textit{CA}{\tiny\textit{mAR}} (\textit{mAR}). The (+/-) is in reference to the corresponding baselines.}
 	\centering
 	\footnotesize
 	\begin{subtable}{0.50\linewidth}
    \centering
 	\begin{tabular}{lllllcc}\toprule
 		\textbf{Model} &\textbf{CP} &\textbf{SCP} &\textbf{CA} 
 		&\textbf{CA}\tiny{\textit{mAR}} &$\alpha$ &\textbf{SS} \\\midrule
 		\textbf{Tuned} & 14.0 & 9.3 & 65.6 & 72.3 & - & - \\\cmidrule{2-7}
 		& 11.8 (-2.2) & 10.5 (+1.2) & 66.8 (+1.2) & 73.3 &0.5 &RB \\
 		& 20.3 (+6.3) & 14.5 (+5.2) & \textbf{67.2 (+1.6)} & 73.3 & $U$ &RB \\
 		& \textbf{34.9 (+20.9)} & \textbf{22.4 (+13.1)} & 66.6 (+1.0) & 73.1 &0.5 &CA \\
 		\multirow{-4}{*}{\textbf{Sd$\rightarrow$Td}}& 28.0 (+14.0) & 18.5 (+9.2) & 67.1 (+1.5) & 
 		\textbf{73.6} & $U$ &CA\\
 		\bottomrule
 	\end{tabular}
 	\caption{Pose Estimation}
 	\label{tab:resultsBpose}
 	\end{subtable}
 	\begin{subtable}{0.50\linewidth}
    \centering
 	\begin{tabular}{lllllcc}\toprule
 		\textbf{Model} &\textbf{CP} &\textbf{SCP} &\textbf{CA} 
 		&\textbf{CA}\tiny{\textit{mAR}} &$\alpha$ &\textbf{SS} \\\midrule
 		\textbf{Tuned} & - & - & \textbf{49.4} & 
 		\textbf{37.0} & - & - \\\cmidrule{2-7}
 		 & - & - &44.3 (-5.1) & 32.9 (-4.1) & 0.5 &RB \\
 		& - & - & 43.0 (-6.4) & 32.6 (-4.4) & $U$ &RB \\
 		& - & - & 43.7 (-5.7) & 33.4 (-3.6) & 0.5 &CA \\
 		\multirow{-4}{*}{\textbf{Sd$\rightarrow$Td}}& 
 		- & - & 43.9 (-5.5) & 32.9 (-4.1) & $U$ &CA\\
 		\bottomrule
 	\end{tabular}
 	\caption{Person Detection}
 	\label{tab:resultsBdetect}
 	\end{subtable}
 \label{tab:resultB}
 \end{table}

 With Tab.~\ref{tab:resultA} and Tab.~\ref{tab:resultB}, we were able to 
 enhance the 
 performance of pose models, with \textit{styled} as well as 
 \textit{styled-tuned} models. \textit{Styled} models can help to improve the 
 performance with a 7.7\,pp (\textit{mAP}) jump in performance, without any 
 labels. While the \textit{styled-tuned} models show that fine-tuning with 
 \textit{styled} models is generally beneficial for the performance, the 
 \textit{Sd$\rightarrow$Td} model for pose gives a significant 1.6\,pp 
 (\textit{mAP}) performance improvement for \textit{CA} dataset over its 
 \textit{Tuned} counterpart. 

 \begin{table}[t]
 \caption{\textbf{Results C}: 
 Comparing \textit{tuned} model with the 
 \textit{styled-tuned~(\emph{Sd}$\rightarrow$\emph{Td})} model, by training on 
 different quantities (25\,\%, 50\,\%, 75\,\%, 100\,\%) of the \emph{CA} data 
 for pose estimation. It clearly shows that the styled model learns quicker. 
 All values in \textit{mAP}, $\alpha=0.5$ or $\alpha=U$ (randomly sampled from a uniform distribution: $U=uniform(0,1)$), 
 and \emph{SS} = \textbf{RB} or \textbf{CA}}
 \label{tab:resultC}
 	\centering
 	\footnotesize
 	\begin{tabular}{lcccccc}\toprule
 		\textbf{Model} &\textbf{25\,\%} &\textbf{50\,\%} &\textbf{75\,\%} 
 		&\textbf{100\,\%} & $\alpha$ & \textbf{SS}\\\midrule
 		\textbf{Tuned} & 61.3 &65.0 &65.1 & 65.6 & - & - \\\cmidrule{2-7}
 		\multirow{4}{*}{\textbf{Sd$\rightarrow$\textbf{Td}}}& 60.6 & 
 		\textbf{65.4} & 65.1 & 66.8 & 0.5 & RB\\
 		&\textbf{62.1}& 64.7& \textbf{66.5}& \textbf{67.2}& $U$& RB\\  
 		&60.6 &65.2 & 65.4 & 66.6 & 0.5 & CA\\
 		&61.4 & 64.8 & 65.7 & 67.1 & $U$& CA\\ 
 		\bottomrule
 	\end{tabular}
 \end{table}

 \paragraph{\textbf{Influence of data quantity}} (Results C, Tab.~\ref{tab:resultC}): 
 Tab.~\ref{tab:resultC} shows that \textit{Styled-Tuned} models learn faster 
 than 
 their \textit{Tuned} counterpart, for each of the corresponding splits of 
 \textit{CA} data (model with $\alpha=U$, \textit{SS=RB} is more consistent). 
 Specifically, the model with $\alpha=0.5$, \textit{SS=RB} and 50\% of 
 \textit{CA} data gives equivalent performance to the \textit{Tuned} model 
 trained with whole \textit{CA} data. 
 We see that the deep learning based models converge faster when they have a suitable initialization of weights~\cite{glorot2010understanding}. 
 We argue that training on styled data helps the model to get a better initialization with respect to the dataset distribution. 
 Consequently, the convergence is faster.

 \paragraph{\textbf{Perceptual Loss Comparison:}} Tab.~\ref{tab:perceptual} shows the 
 influence of perceptual loss for the model performance. $L_{comb1}$ is the 
 experiment with adaptively weighing the detector as well as pose losses 
 (Eq.~\ref{eq:perceptual1}). For $L_{comb2}$ (Eq.~\ref{eq:perceptual2}), we 
 determined the values of $\lambda$s through a parameter search \citep{optuna}: 
 $\lambda_{1}=0.43$ and $\lambda_{2}=0.92$ for person detection and 
 $\lambda_{1}=0.47, \lambda_{2}=0.018$ for pose estimation.
 We present results for the combination \textit{SS=CA}~\&~$\alpha=U$ for the 
 detector, and we chose \textit{SS=CA}~\&~$\alpha=0.5$ for pose estimator. 

 Tab.~\ref{tab:perceptual} shows that the perceptual loss (adaptive:$L_{comb1}$ 
 or 
 parameterised one:$L_{comb2}$) indeed helps the \textit{styled} as well as 
 \textit{Sd$\rightarrow$Td} models to improve their performance, in general.
 From \cref{tab:tab:perceptualA}, we can see that the results for $L_{comb1}$ and $L_{comb2}$ are equal, however we have to note that this is an empirical observation for different values of lambda.
 Additionally, we note that perceptual loss does not harm the \textit{styled} model for pose estimation,  but actually helps since the \textit{styled} model is fine-tuned on 
 \textit{CA}. 

 \begin{table}[t]
 	\caption{\textbf{Perceptual Loss Comparison}: Comparing different 
 	loss combinations using the \textit{styled} or 
 	\textit{styled-tuned~(\emph{Sd}$\rightarrow$\emph{Td})} model on the 
 	\emph{CA} dataset for \subref{tab:tab:perceptualA} pose estimation
 	and \subref{tab:tab:perceptualB} detection, \ie just the detector or pose loss $L_{det}$/$L_{pose}$ or in 
 	combination with perceptual loss $L_{comb}$ in two different 
 	variants (\cref{eq:perceptual1,eq:perceptual2}. For \subref{tab:tab:perceptualA}: $\alpha=0.5$ 
 	and for \subref{tab:tab:perceptualB}: 
 	$\alpha=U$. \textit{Style-Set} (\emph{SS}) is 
 	\textit{CA} for both.}
 	\label{tab:perceptual}
 	\centering
 	\footnotesize
 	\begin{subtable}{0.49\linewidth}
 	\centering
 	    \begin{tabular}{lccc}\toprule
 		\textbf{Model} & \textit{mAP} & \textit{mAR} & $L_{comb}$ \\\midrule
 		& \textbf{30.6} & \textbf{36.9} & $L_{pose}$ \\
 		& 30.5 & 36.5 & $L_{comb1}$ \\
 		\multirow{-3}{*}{\textbf{Styled}}& 30.5 & 36.5 & $L_{comb2}$ 
 		\\\cmidrule{2-4}
 		& 66.6 & 73.1 & $L_{pose}$ \\
 		& \textbf{67.2} & \textbf{73.6} & $L_{comb1}$ \\
 		\multirow{-3}{*}{\textbf{Sd$\rightarrow$Td}}& \textbf{67.2} & 
 		\textbf{73.6} & $L_{comb2}$ \\
 		\bottomrule
 	\end{tabular}
 	\caption{Pose Estimation}
 	\label{tab:tab:perceptualA}
 	\end{subtable}
 	\begin{subtable}{0.49\linewidth}
 	\centering
 	    \begin{tabular}{lccc}\toprule
 		\textbf{Model} & \textit{mAP} & \textit{mAR} & $L_{comb}$ \\\midrule
 		& 6.5 & 8.6 & $L_{det}$\\
 		& \textbf{11.2} & 11.0 & $L_{comb1}$ \\
 		\multirow{-3}{*}{\textbf{Styled}}& 9.8 & 
 		\textbf{11.2} & $L_{comb2}$ \\\cmidrule{2-4}
 		& 43.9 & 32.9 & $L_{det}$ \\
 		& 45.4 & \textbf{34.7} & $L_{comb1}$ \\
 		\multirow{-3}{*}{\textbf{Sd$\rightarrow$Td}}& 
 		\textbf{45.7} & 34.0 & $L_{comb2}$ \\
 		\bottomrule
 	\end{tabular}
 	\caption{Person Detection}
 	\label{tab:tab:perceptualB}
 	\end{subtable}
 	
 \end{table}

 \subsection{Qualitative Pose Estimation Results}
\begin{figure*}[t]
 	\begin{subfigure}{0.19\linewidth}
 		\centering
 		\includegraphics[width=1.015\linewidth,height=\linewidth]{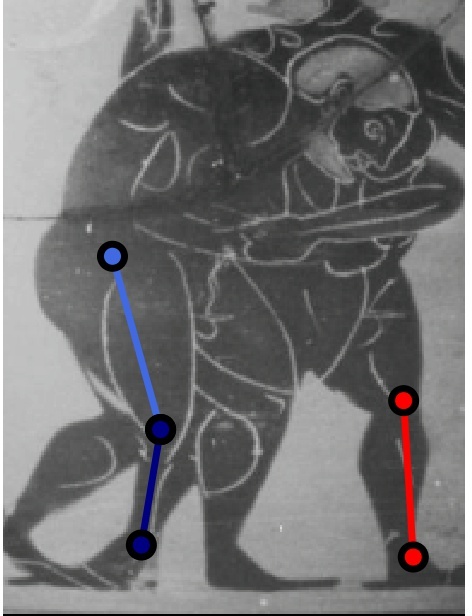}
 	\end{subfigure}
 	\begin{subfigure}{0.19\linewidth}
 		\centering
 		\includegraphics[width=1.015\linewidth,height=\linewidth]{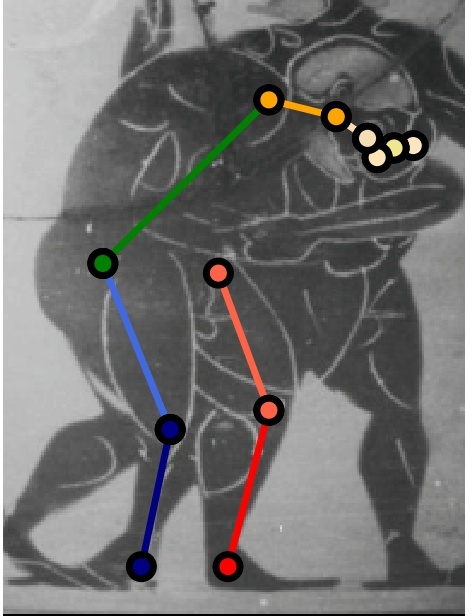}
 	\end{subfigure}
 	\begin{subfigure}{0.19\linewidth}
 		\centering
 		\includegraphics[width=1.025\linewidth,height=\linewidth]{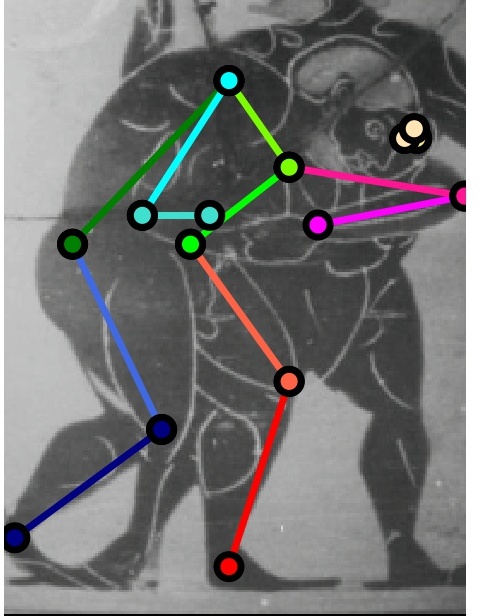}
 	\end{subfigure}
 	\begin{subfigure}{0.19\linewidth}
 		\centering
 		\includegraphics[width=1.03\linewidth,height=\linewidth]{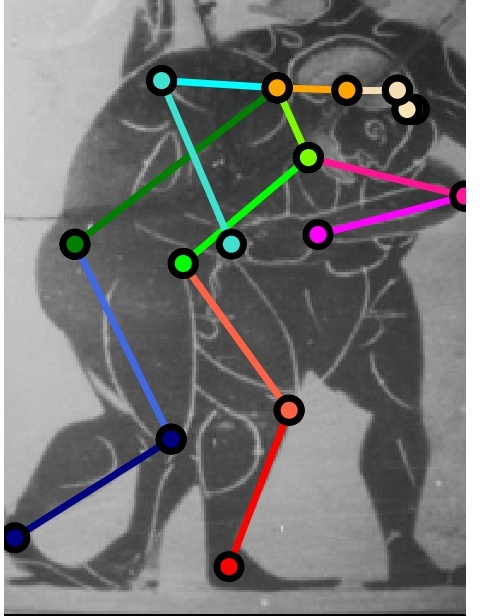}
 	\end{subfigure}
 	\begin{subfigure}{0.19\linewidth}
 		\centering
 		\includegraphics[width=\linewidth,height=\linewidth]{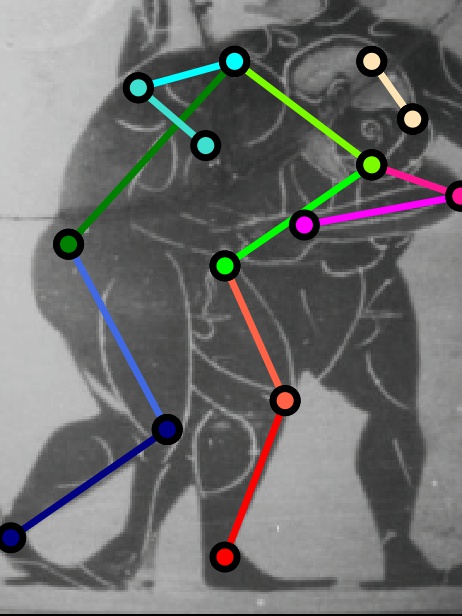}
 	\end{subfigure}

 	\begin{subfigure}{0.19\linewidth}
 		\centering
 		\includegraphics[width=1.015\linewidth,height=\linewidth]{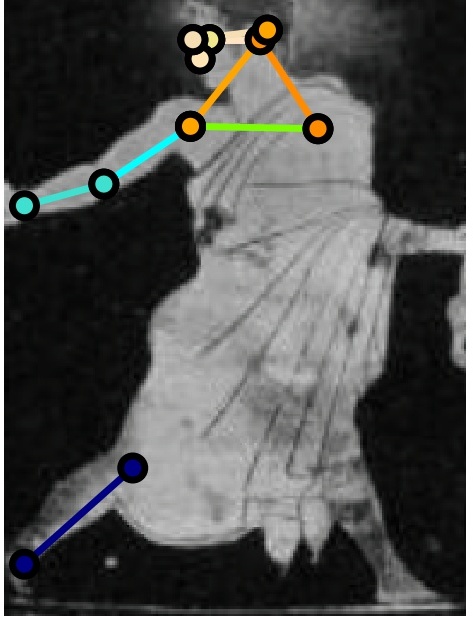}
 	\end{subfigure}
 	\begin{subfigure}{0.19\linewidth}
 		\centering
 		\includegraphics[width=1.027\linewidth,height=\linewidth]{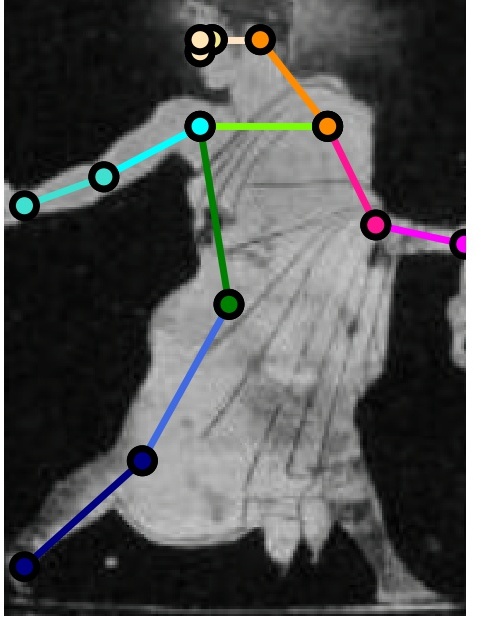}
 	\end{subfigure}
 	\begin{subfigure}{0.19\linewidth}
 		\centering
 		\includegraphics[width=1.02\linewidth,height=\linewidth]{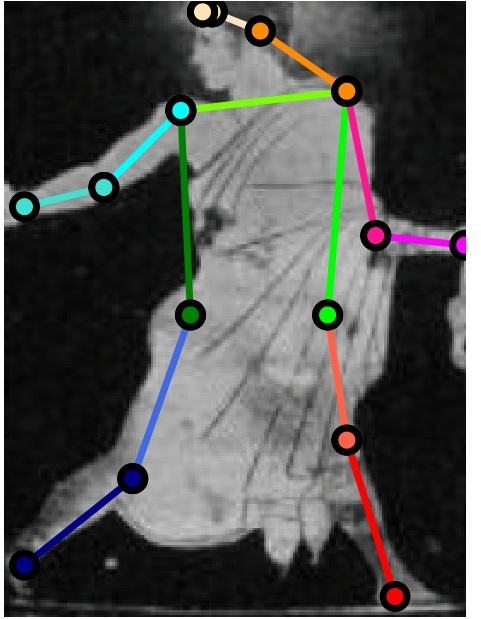}
 	\end{subfigure}
 	\begin{subfigure}{0.19\linewidth}
 		\centering
 		\includegraphics[width=1.03\linewidth,height=\linewidth]{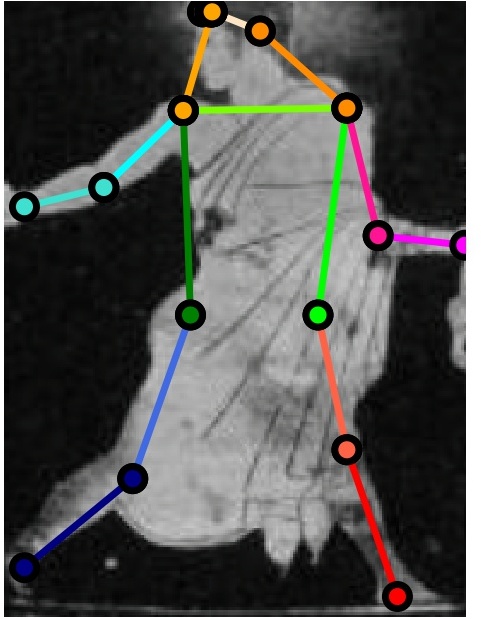}
 	\end{subfigure}
 	\begin{subfigure}{0.19\linewidth}
 		\centering
 		\includegraphics[width=\linewidth,height=\linewidth]{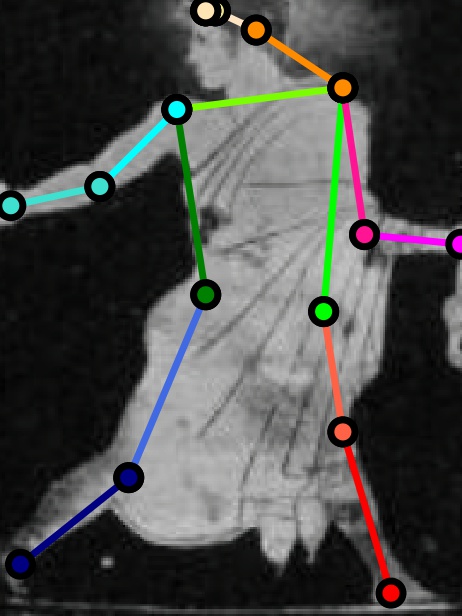}
 	\end{subfigure}

 	\begin{subfigure}{0.19\linewidth}
 		\centering
 		\includegraphics[width=\linewidth,height=\linewidth]{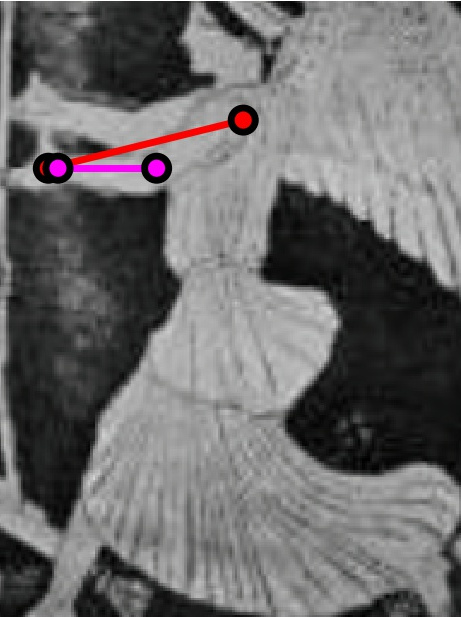}
 	\end{subfigure}
 	\begin{subfigure}{0.19\linewidth}
 		\centering
 		\includegraphics[width=0.98\linewidth,height=\linewidth]{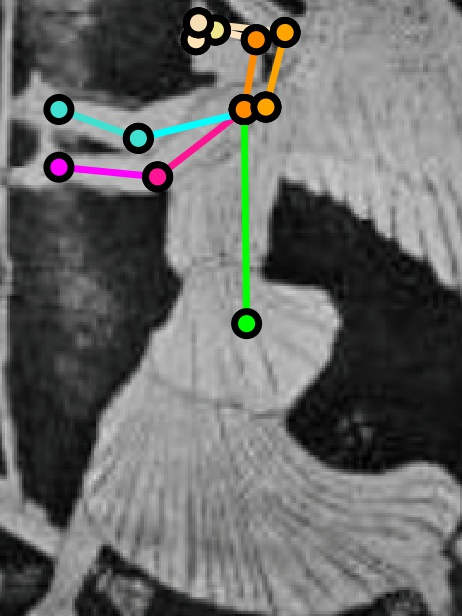}
 	\end{subfigure}
 	\begin{subfigure}{0.19\linewidth}
 		\centering
 		\includegraphics[width=0.97\linewidth,height=\linewidth]{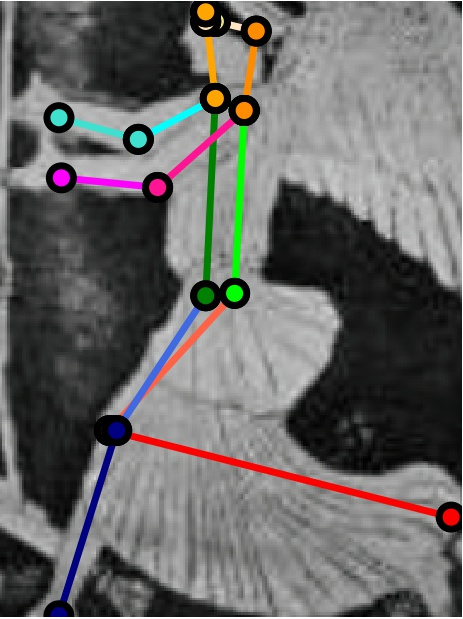}
 	\end{subfigure}
 	\begin{subfigure}{0.19\linewidth}
 		\centering
 		\includegraphics[width=0.99\linewidth,height=\linewidth]{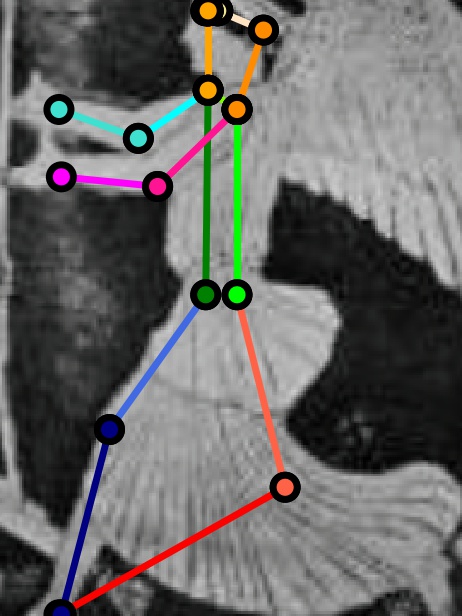}
 	\end{subfigure}
     \begin{subfigure}{0.19\linewidth}
 		\centering
 		\includegraphics[width=\linewidth,height=\linewidth]{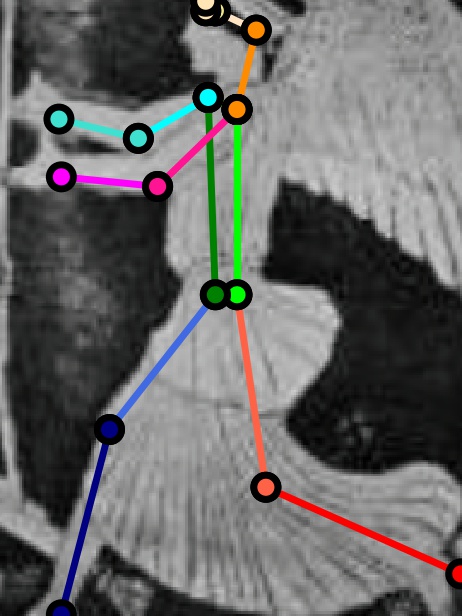}
 	\end{subfigure}

 	\begin{subfigure}{0.19\linewidth}
 		\centering
 		\includegraphics[width=\linewidth,height=\linewidth]{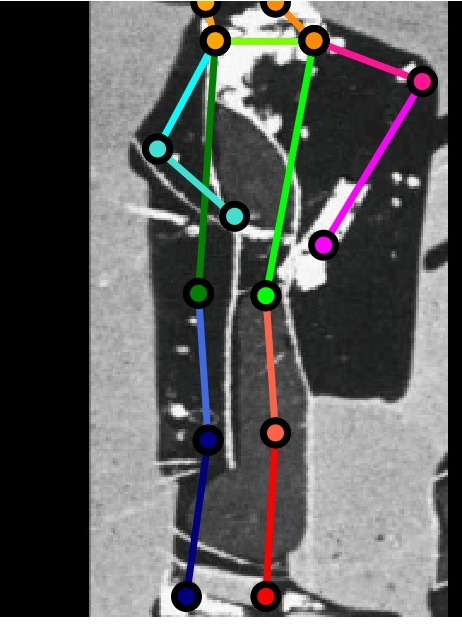}
 		\caption{\emph{Baseline}}
 		\label{fig:baseline}	
 	\end{subfigure}
 	\begin{subfigure}{0.19\linewidth}
 		\centering
 		\includegraphics[width=0.98\linewidth,height=\linewidth]{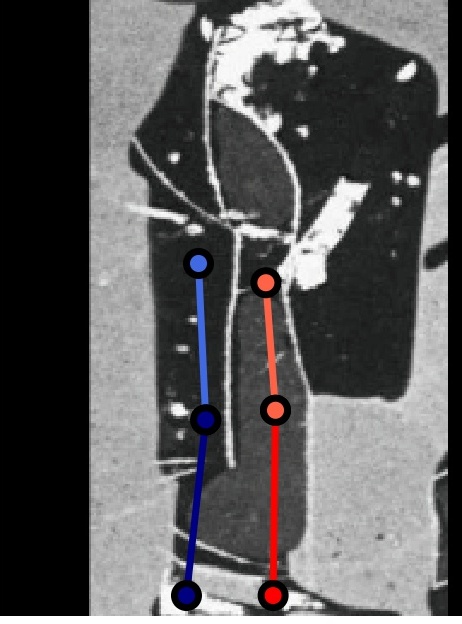}
 		\caption{\emph{Styled}}
 		\label{fig:styled}
 	\end{subfigure}
 	\begin{subfigure}{0.19\linewidth}
 		\centering
 		\includegraphics[width=\linewidth,height=\linewidth]{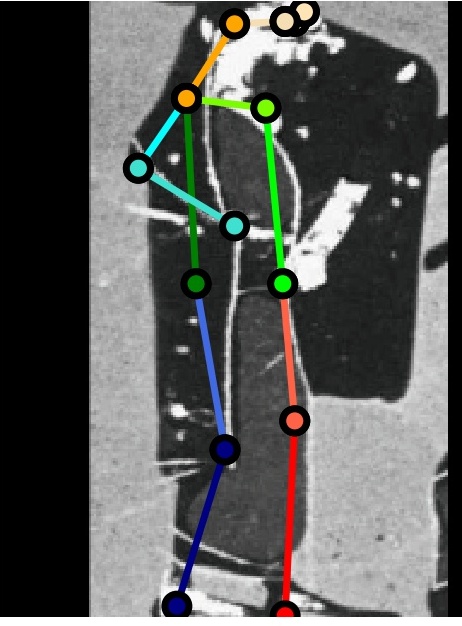}
 		\caption{\emph{Tuned}}
 		\label{fig:tuned}			
 	\end{subfigure}
 	\begin{subfigure}{0.19\linewidth}
 		\centering
 		\includegraphics[width=\linewidth,height=\linewidth]{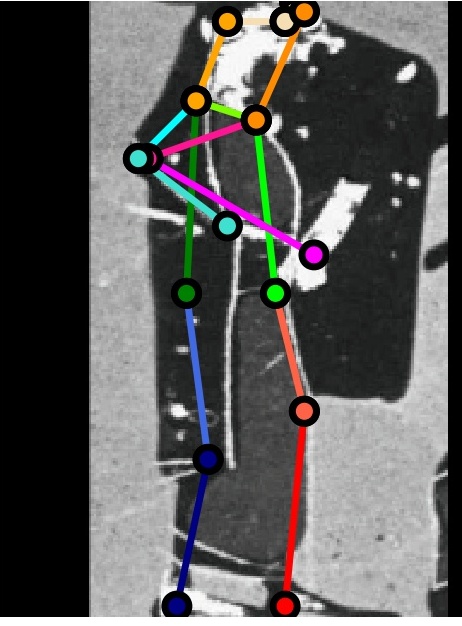}
 		\caption{\footnotesize\emph{Sd$\rightarrow$Td}}
 		\label{fig:styledtuned}
 	\end{subfigure}	
 	\begin{subfigure}{0.19\linewidth}
 		\centering
 		\includegraphics[width=\linewidth,height=\linewidth]{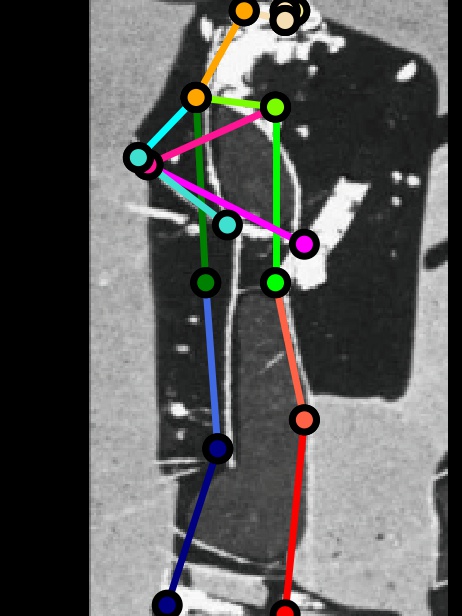}
 		\caption{\footnotesize\emph{Sd$\rightarrow$Td}$_{p}$}
 		\label{fig:lambdas}			
 	\end{subfigure}
 	\caption{\textbf{Pose models comparison}: Pose Predictions on 4 
 	examples each from \subref{fig:baseline} \textit{baseline}, 
 	\subref{fig:styled} \textit{styled}, \subref{fig:tuned} \textit{tuned}, 
 	\subref{fig:styledtuned} \textit{styled-tuned~(St$\rightarrow$Td)} and 
 	\subref{fig:lambdas} \textit{styled-tuned} with perceptual 
 	loss~(\emph{Sd$\rightarrow$Td}$_{p}$) models. The results clearly show the 
 	superiority of predicted poses with the \emph{St$\rightarrow$Td} and 
 	\emph{Sd$\rightarrow$Td}$_{p}$ models. The characters starting from the top 
 	are called \textit{wrestler}, \textit{fleeing}, \textit{persecutor} and 
 	\textit{bride}}.
 	\label{fig:results_sup}
 \end{figure*}

 Tabs.~\ref{tab:resultB}~\&~\ref{tab:perceptual} show that 
 \textit{styled-tuned} models 
 are 
 consistently giving a better performance than any other method. We visualise 
 the predictions of these models for comparison of their performances. 
 Fig.~\ref{fig:results_sup} shows four characters (\textit{wrestler}, 
 \textit{fleeing}, 
 \textit{persecutor}, \textit{bride}) and their pose predictions from each of 
 our 5 proposed models. 

 As shown in Fig.~\ref{fig:baseline}, the \textit{baseline} model is the 
 poorest in 
 pose predictions. It is not able to detect majority of keypoints, confuses 
 between the limbs if multiple characters are present and incorrectly predicts 
 the keypoint locations.

 \textit{Styled} model is generally much better (Fig.~\ref{fig:styled}) than 
 the 
 baseline model (also Tab.~\ref{tab:resultA}). It is able to predict more 
 keypoints 
 and does not get confused if multiple characters are present. However, it is 
 not able to predict all the visible keypoints and sometimes 
 (Fig.~\ref{fig:styled}, \textit{last row}) gives worse performance than even a 
 \textit{baseline} model. 

 \textit{Tuned}, \textit{styled-tuned~(Sd$\rightarrow$Td)} and 
 \textit{styled-tuned} with perceptual loss (\textit{Sd$\rightarrow$Td$_{p}$}) 
 models are overall quite superior to \textit{baseline} and \textit{styled} 
 models. They are able to predict almost all of the visible keypoints, do not 
 confuse between multiple characters and are quite precise with the keypoint 
 locations. However, there are subtle differences that make 
 \textit{Sd$\rightarrow$Td} models better. They are able to predict all the 
 visible keypoints as shown in Fig.~\ref{fig:styledtuned} and 
 Fig.~\ref{fig:lambdas}, 
 whereas tuned models miss some (\eg Fig.~\ref{fig:tuned} where the shoulder 
 joints 
 are missing). The keypoint location precision is also improved using 
 \textit{Sd$\rightarrow$Td} models. Visually it is difficult to generalise if 
 models with perceptual loss are better or not, however, they are more precise 
 (Fig.~\ref{fig:lambdas}, third row ankle is corrected, but a shoulder is 
 missed).

\section{Pose-Based Retrieval}\label{sec:results}
 Our experiments (Sec.~\ref{sec:experiments}) showed that \textit{Styled}
 models and \textit{Styled-Tuned} models achieve better keypoint detection results, quantitatively and qualitatively, than their corresponding counterparts. 
 In this section, we show that our two-step training pipeline is also beneficial for discovering similar images based on character poses. 
 We call the process of retrieving images based on poses as \emph{pose-based retrieval}.

 \subsection{Experimental Setup}
 The database for image retrieval and discovery is built from the \emph{CA} validation dataset. 
 The database consists of 303 images and their respective detected poses for best of  \textit{baseline}, \textit{styled}, \textit{tuned}, \textit{Sd$\rightarrow$Td} and \textit{Sd$\rightarrow$Td$_{p}$~(with perceptual loss)}. 
 We perform two retrieval experiments based on the class label for each image, which is either a \textit{character} or \textit{scene}.
 There are 15 unique \emph{characters} (\textbf{C}) and 5 \emph{Scenes} (\textbf{S}). 
 Given a query image, we rank the retrieved images based on the OKS metric \citep{Lin_COCODataset_2014}. 
 In order to evaluate the retrieval method, we use the precision as: $P^*=\frac{TP^*}{TP^*+FP^*}$, where $^*=@k$, consequently P@k := Precision at k; TP := true positives, FP := false positives and FN := false negatives.
 We report P@k and mAP, for k=1 and k=5. 
 In all our experiments, we exclude the self-retrieval (query itself) from the evaluation. 
 For this task, we compare all the presented models to highlight the quality of our proposed models from an application perspective. The focus of this work is on enhancing poses for Greek vase paintings and not presenting a novel image retrieval method and hence we do not compare with SOTA image retrieval methods.

\begin{table}[t]
 	\caption{\textbf{Retrieval Results}: The \textbf{(C)*} models show 
 	the retrieval values based on \textit{characters}, where as the 
 	\textbf{(S)*} models show for the \textit{scenes}. \textbf{P} is 
 	\textit{Precision}, and \textbf{mAP} is \textit{mean-Average Precision}. 
 	$\alpha=U$; \textit{SS: Style-Set}. \textit{Sd$\rightarrow$Td*} are 
 	\textit{style-tuned} models, where 
 	\textit{$p1=L_{comb1}$}~(Eq.~\ref{eq:perceptual1}) and 
 	\textit{$p2=L_{comb2}$}~(Eq.~\ref{eq:perceptual2})}
 	\label{tab:retrieval}
 	\centering
 	\footnotesize
 	\begin{tabular}{lccccc}\toprule
 		\textbf{Model} &\textbf{P@1} &\textbf{P@5} &\textbf{mAP} & $\alpha$ & 
 		\textbf{SS}\\\midrule
 		\textbf{(C)~Baseline}& 31.7 & 25.5 & 21.5 & - & -\\
 		\textbf{(C)~Styled}& 37.3 & 30.2 & 23.1 & $U$ & CA  \\  
 		\textbf{(C)~Tuned}& 43.0 & 39.8 & 27.5 & - & - \\  
 		\textbf{(C)~Sd$\rightarrow$Td}& 47.7 & \textbf{42.2} & 28.3 & $U$ & CA 
 		\\
 		\textbf{(C)~Sd$\rightarrow$Td$_{p1}$}& 45.7 & 41.4 & 28.0 & 0.5 & CA 
 		\\  
 		\textbf{(C)~Sd$\rightarrow$Td$_{p2}$}& \textbf{48.3} & 41.1 & 
 		\textbf{28.4} & 0.5 & CA \\  
 		\cmidrule{2-6}  
 		\textbf{(S)~Baseline}& 43.6 & 43.2 & 35.9 & - & - \\  
 		\textbf{(S)~Styled}& 46.9 & 43.6 & 37.3 & $U$ & CA \\  
 		\textbf{(S)~Tuned}& 56.4 & 52.7 & 41.5 & - & - \\  
 		\textbf{(S)~Sd$\rightarrow$\textbf{Td}}& 58.8 & \textbf{55.2} & 
 		\textbf{42.1} & $U$ & CA \\  
 		\textbf{(S)~Sd$\rightarrow$Td$_{p1}$} & 57.8 & 53.5 & 41.5 & 0.5 & CA 
 		\\  
 		\textbf{(S)~Sd$\rightarrow$Td$_{p2}$} & \textbf{58.8} & 53.4 & 41.8 & 
 		0.5 & CA \\  
 		\bottomrule
 	\end{tabular}
 \end{table}
 
 \begin{figure}[t]
  	\fboxsep=0pt
 	\fboxrule=1pt
 	\begin{subfigure}{0.12\linewidth}
 		\centering
 		\rotatebox[origin=c]{0}{$baseline$}
 	\end{subfigure}
 	\,
 	\hfill
 	\begin{subfigure}{0.12\linewidth}
 		\centering
 		\includegraphics[width=\linewidth,height=\linewidth]{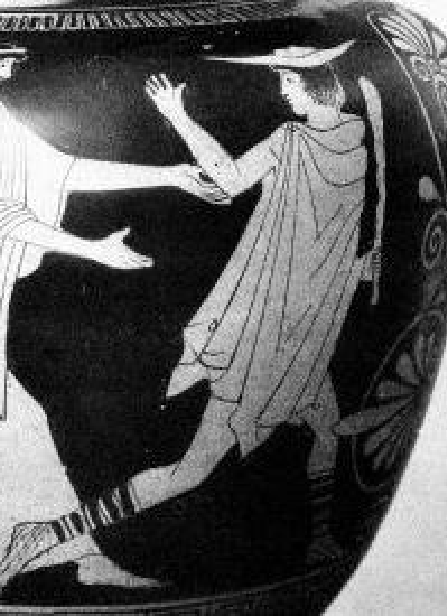}
 	\end{subfigure}
 	\,
 	\hfill
 	\begin{subfigure}{0.12\linewidth}
 		\centering
 		\includegraphics[width=\linewidth,height=\linewidth]{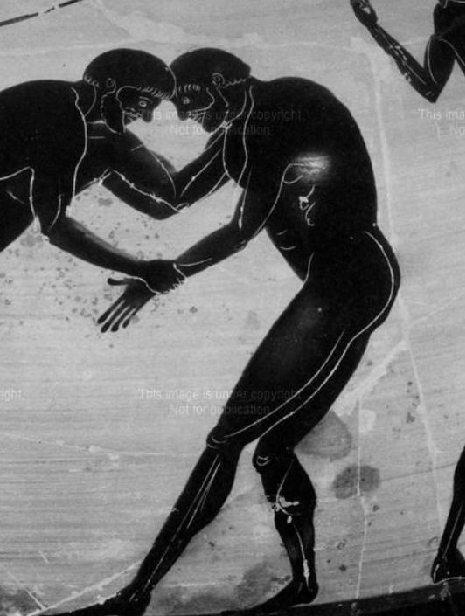}
 	\end{subfigure}
 	\,
 	\hfill
 	\begin{subfigure}{0.12\linewidth}
 		\centering
 		\includegraphics[width=\linewidth,height=\linewidth]{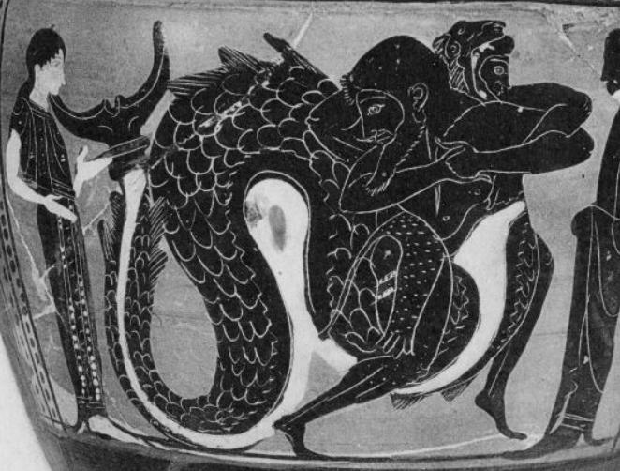}
 	\end{subfigure}
 	\,
 	\hfill
 	\begin{subfigure}{0.12\linewidth}
 		\centering
 		\includegraphics[width=\linewidth,height=\linewidth]{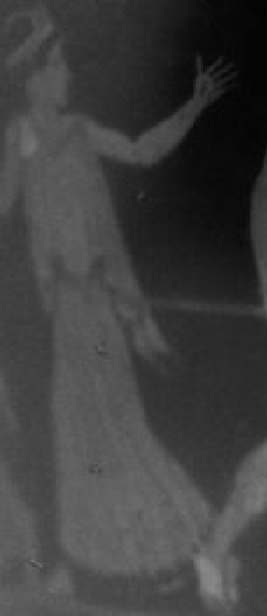}
 	\end{subfigure}
 	\,
 	\hfill
 	\begin{subfigure}{0.12\linewidth}
 		\centering
 		\includegraphics[width=\linewidth,height=\linewidth]{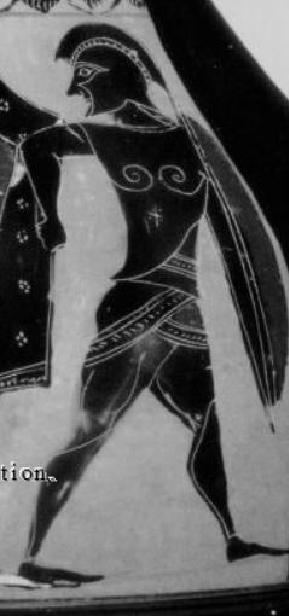}
 	\end{subfigure}
 	\,
 	\hfill
 	\begin{subfigure}{0.12\linewidth}
 		\centering
 		\includegraphics[width=\linewidth,height=\linewidth]{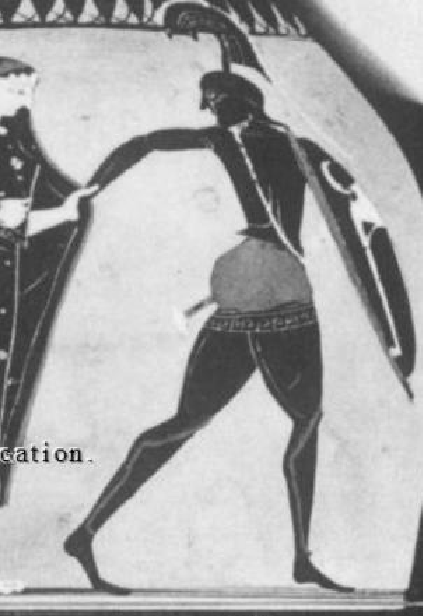}
 	\end{subfigure}

	\begin{subfigure}{0.12\linewidth}
 		\centering
 		\rotatebox[origin=c]{0}{$styled$}
 	\end{subfigure}
 	\,
 	\hfill
     \begin{subfigure}{0.12\linewidth}
 		\centering
 		\includegraphics[width=\linewidth,height=\linewidth]{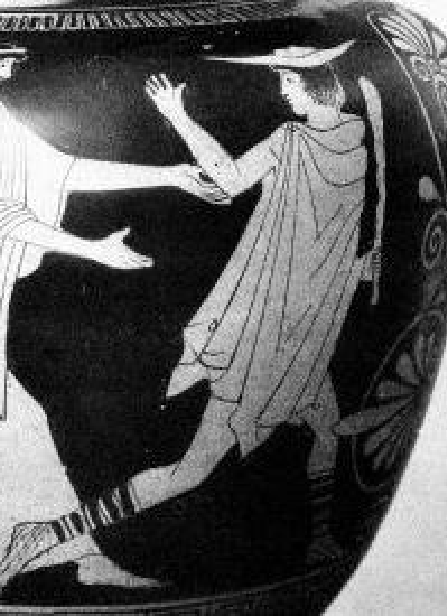}
 	\end{subfigure}
 	\,
 	\hfill
 	\begin{subfigure}{0.12\linewidth}
 		\centering
 		\includegraphics[width=\linewidth,height=\linewidth]{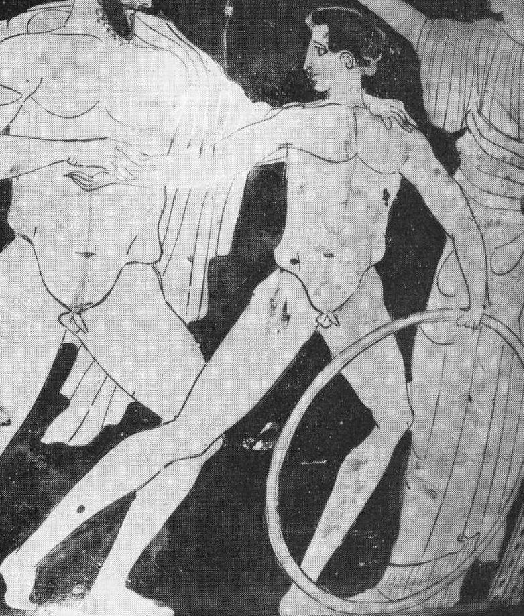}
 	\end{subfigure}
 	\,
 	\hfill
 	\begin{subfigure}{0.12\linewidth}
 		\centering
 		\includegraphics[width=0.93\linewidth,height=0.93\linewidth,cfbox=green 1pt 1pt]{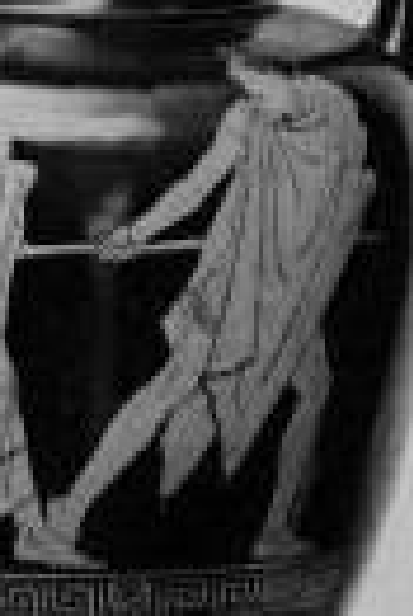}
 	\end{subfigure}
 	\,
 	\hfill
 	\begin{subfigure}{0.12\linewidth}
 		\centering
 		\includegraphics[width=\linewidth,height=\linewidth]{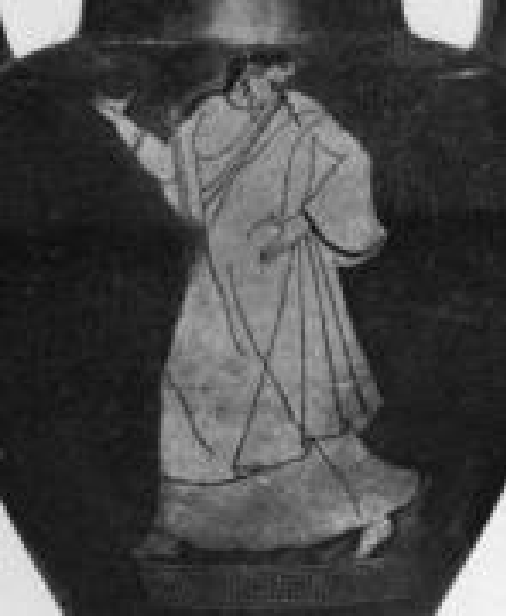}
 	\end{subfigure}
 	\,
 	\hfill
 	\begin{subfigure}{0.12\linewidth}
 		\centering
 		\includegraphics[width=\linewidth,height=\linewidth]{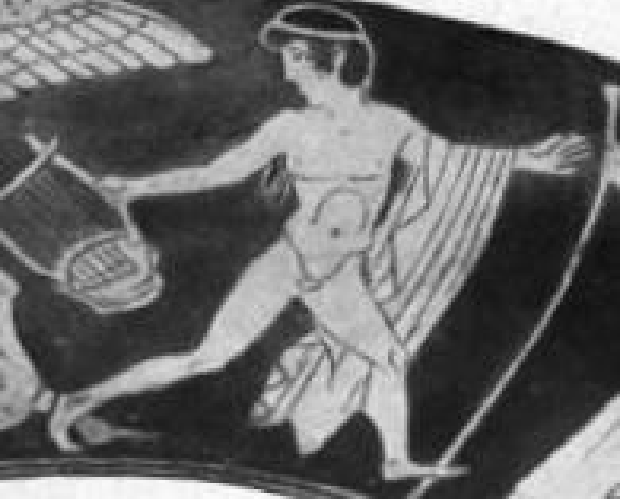}
 	\end{subfigure}
 	\,
 	\hfill
 	\begin{subfigure}{0.12\linewidth}
 	    \centering
 		\includegraphics[width=\linewidth,height=\linewidth]{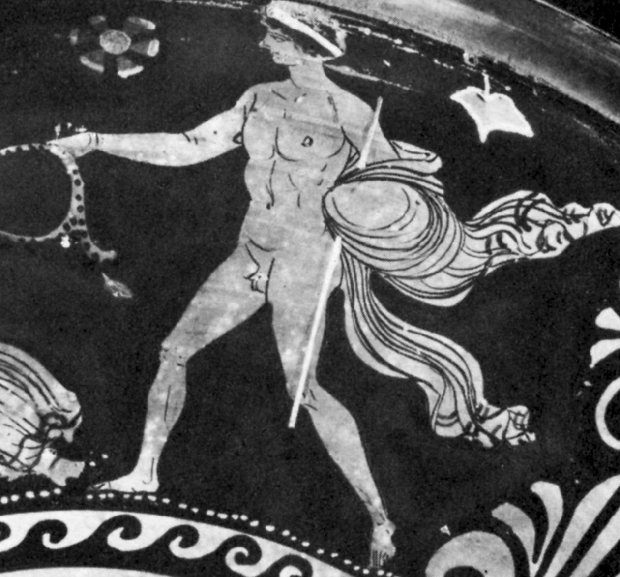}
 	\end{subfigure}

	\begin{subfigure}{0.12\linewidth}
 		\centering
 		\rotatebox[origin=c]{0}{$tuned$}
 	\end{subfigure}
 	\,
 	\hfill
    \begin{subfigure}{0.12\linewidth}
 		\centering
 		\includegraphics[width=\linewidth,height=\linewidth]{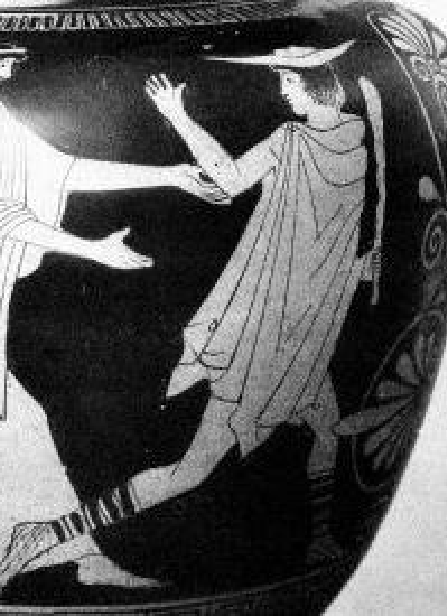}
 	\end{subfigure}
 	\,
 	\hfill
 	\begin{subfigure}{0.12\linewidth}
 		\centering
 		\includegraphics[width=0.93\linewidth,height=0.93\linewidth,cfbox=green 1pt 1pt]{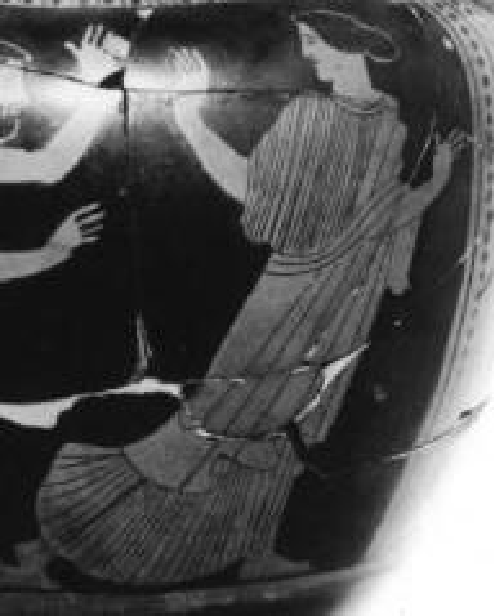}
 	\end{subfigure}
 	\,
 	\hfill
 	\begin{subfigure}{0.12\linewidth}
 		\centering
 		\includegraphics[width=0.93\linewidth,height=0.93\linewidth,cfbox=green 1pt 1pt]{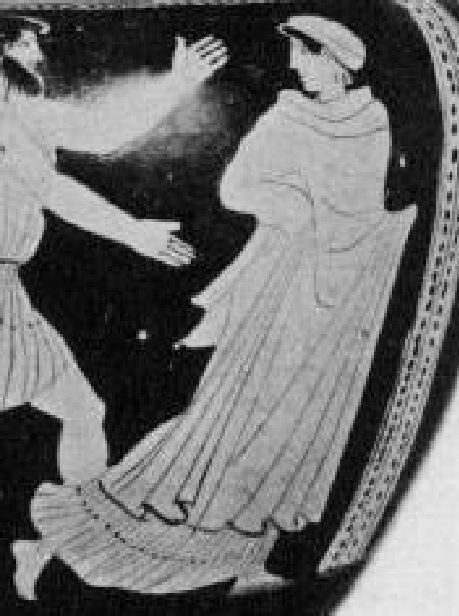}
 	\end{subfigure}
 	\,
 	\hfill
 	\begin{subfigure}{0.12\linewidth}
 		\centering
 		\includegraphics[width=\linewidth,height=\linewidth]{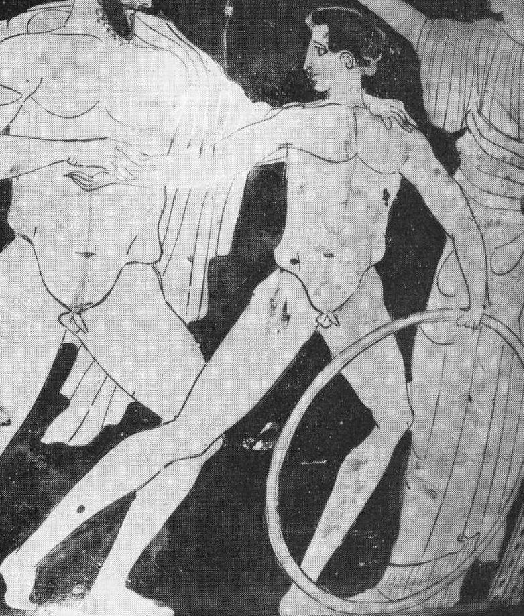}
 	\end{subfigure}
 	\,
 	\hfill
 	\begin{subfigure}{0.12\linewidth}
 		\centering
 		\includegraphics[width=0.93\linewidth,height=0.93\linewidth,cfbox=green 1pt 1pt]{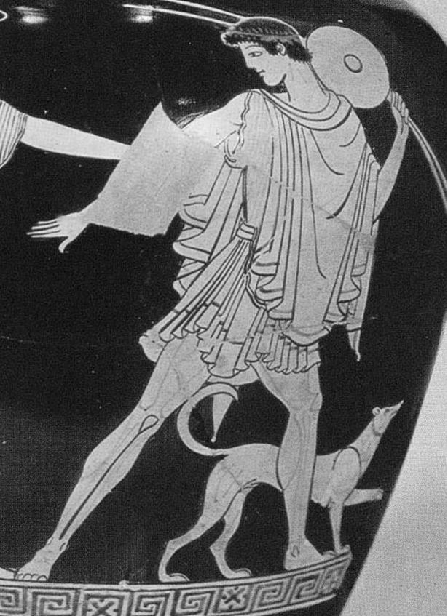}
 	\end{subfigure}
 	\,
 	\hfill
 	\begin{subfigure}{0.12\linewidth}
 		\centering
 		\includegraphics[width=0.93\linewidth,height=0.93\linewidth,cfbox=green 1pt 1pt]{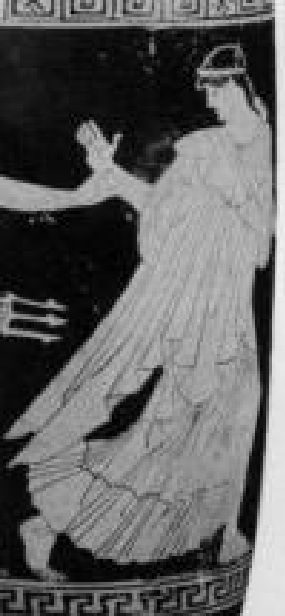}
 	\end{subfigure}

    \begin{subfigure}{0.12\linewidth}
 		\centering
 		\rotatebox[origin=c]{0}{\textit{Sd$\rightarrow$Td}}
 	\end{subfigure}
 	\,
 	\hfill
    \begin{subfigure}{0.12\linewidth}
 		\centering
 		\includegraphics[width=\linewidth,height=\linewidth]{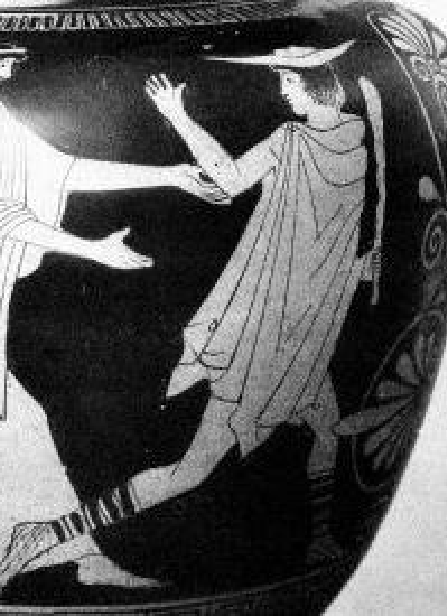}
 	\end{subfigure}
 	\,
 	\hfill
 	\begin{subfigure}{0.12\linewidth}
 		\centering
 		\includegraphics[width=0.93\linewidth,height=0.93\linewidth,cfbox=green 1pt 1pt]{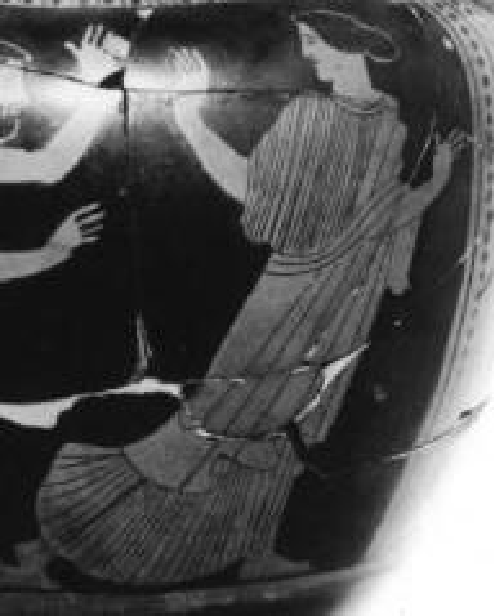}
 	\end{subfigure}
 	\,
 	\hfill
 	\begin{subfigure}{0.12\linewidth}
 		\centering
 		\includegraphics[width=0.93\linewidth,height=0.93\linewidth,cfbox=green 1pt 1pt]{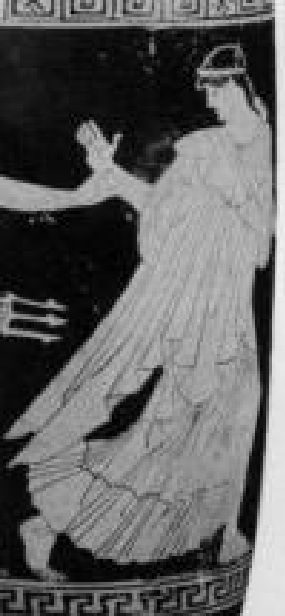}
 	\end{subfigure}
 	\,
 	\hfill
 	\begin{subfigure}{0.12\linewidth}
 		\centering
 		\includegraphics[width=0.93\linewidth,height=0.93\linewidth,cfbox=green 1pt 1pt]{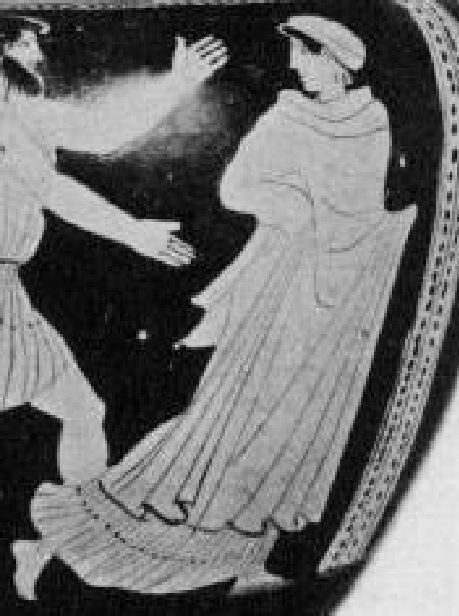}
 	\end{subfigure}
 	\,
 	\hfill
 	\begin{subfigure}{0.12\linewidth}
 		\centering
 		\includegraphics[width=0.93\linewidth,height=0.93\linewidth,cfbox=green 1pt 1pt]{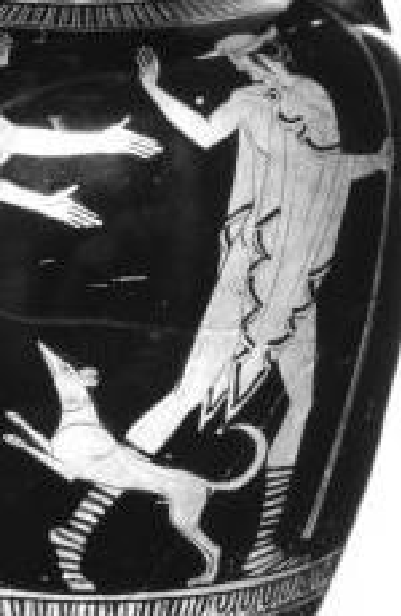}
 	\end{subfigure}
 	\,
 	\hfill
 	\begin{subfigure}{0.12\linewidth}
 		\centering
 		\includegraphics[width=0.93\linewidth,height=0.93\linewidth,cfbox=green 1pt 1pt]{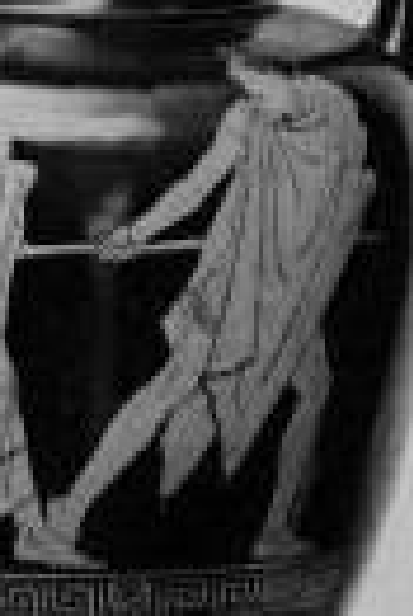}
 	\end{subfigure}

    \begin{subfigure}{0.12\linewidth}
 		\centering
 		\rotatebox[origin=c]{0}{\textit{Sd$\rightarrow$Td$_{p1}$}}
 	\end{subfigure}
 	\,\hfill
    \begin{subfigure}{0.12\linewidth}
 		\centering
 		\includegraphics[width=\linewidth,height=\linewidth]{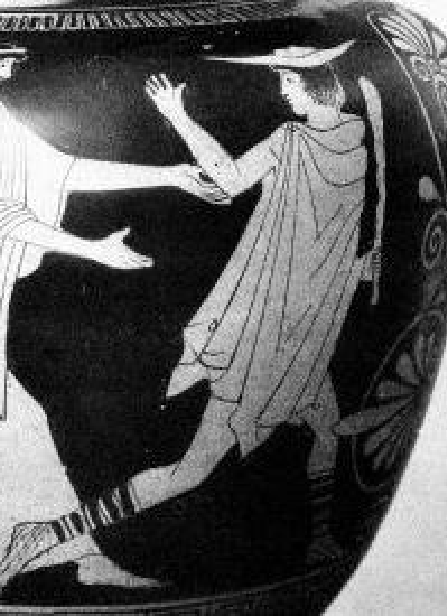}
 	\end{subfigure}
 	\,\hfill
 	\begin{subfigure}{0.12\linewidth}
 		\centering
 		\includegraphics[width=0.93\linewidth,height=0.93\linewidth,cfbox=green 1pt 1pt]{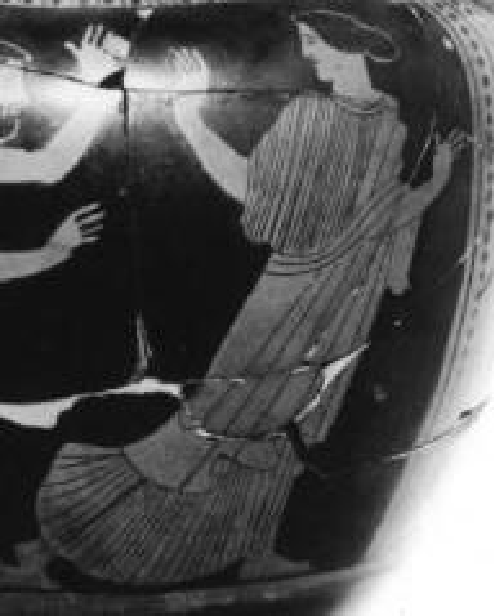}
 	\end{subfigure}
 	\,\hfill
 	\begin{subfigure}{0.12\linewidth}
 		\centering
 		\includegraphics[width=0.93\linewidth,height=0.93\linewidth,cfbox=green 1pt 1pt]{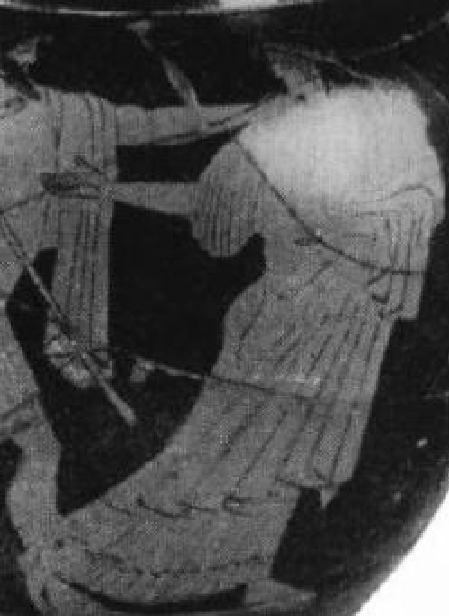}
 	\end{subfigure}
 	\,\hfill
 	\begin{subfigure}{0.12\linewidth}
 		\centering
 		\includegraphics[width=\linewidth,height=\linewidth]{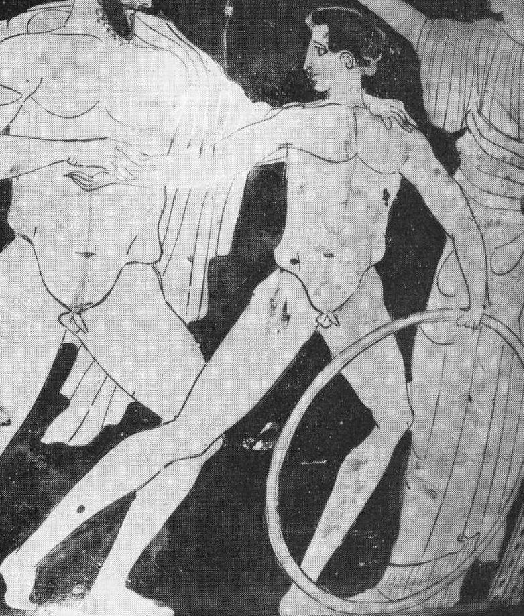}
 	\end{subfigure}
 	\,\hfill
 	\begin{subfigure}{0.12\linewidth}
 		\centering
 		\includegraphics[width=0.93\linewidth,height=0.93\linewidth,cfbox=green 1pt 1pt]{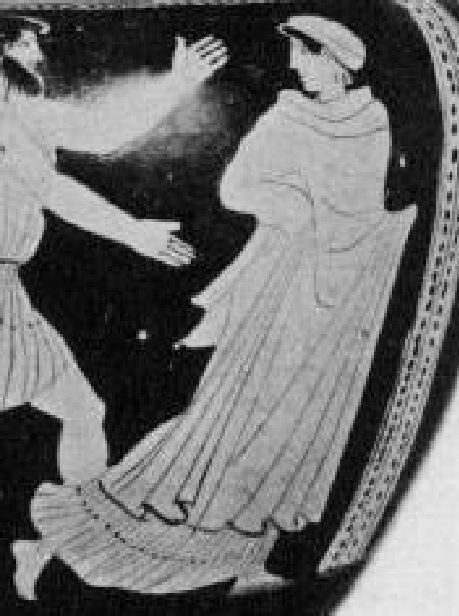}
 	\end{subfigure}
 	\,\hfill
 	\begin{subfigure}{0.12\linewidth}
 		\centering
 		\includegraphics[width=0.93\linewidth,height=0.93\linewidth,cfbox=green 1pt 1pt]{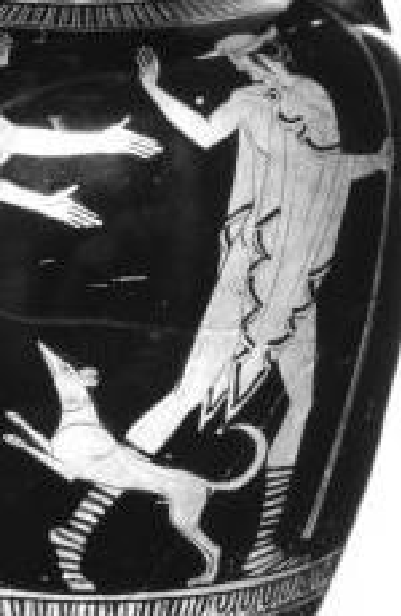}
 	\end{subfigure}

 	\begin{subfigure}{0.12\linewidth}
 		\centering
 		\rotatebox[origin=c]{0}{\textit{Sd$\rightarrow$Td$_{p2}$}}
 	\end{subfigure}
 	\,
 	\hfill
    \begin{subfigure}{0.12\linewidth}
 		\centering
 		\includegraphics[width=\linewidth,height=\linewidth]{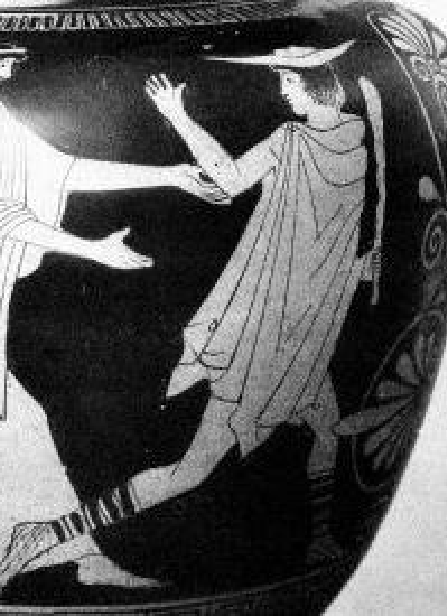}
         \caption{$q$}
         \label{fig:results-retrieval1}
 	\end{subfigure}
 	\,
 	\hfill
 	\begin{subfigure}{0.12\linewidth}
 		\centering
 		\includegraphics[width=0.93\linewidth,height=0.93\linewidth,cfbox=green 1pt 1pt]{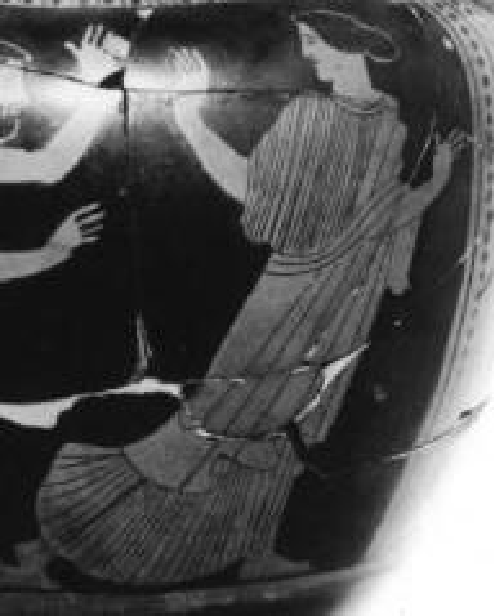}
         \caption{$m_1(q)$}
 		\label{fig:results-retrieval2}
 	\end{subfigure}
 	\,
 	\hfill
 	\begin{subfigure}{0.12\linewidth}
 		\centering
 		\includegraphics[width=0.93\linewidth,height=0.93\linewidth,cfbox=green 1pt 1pt]{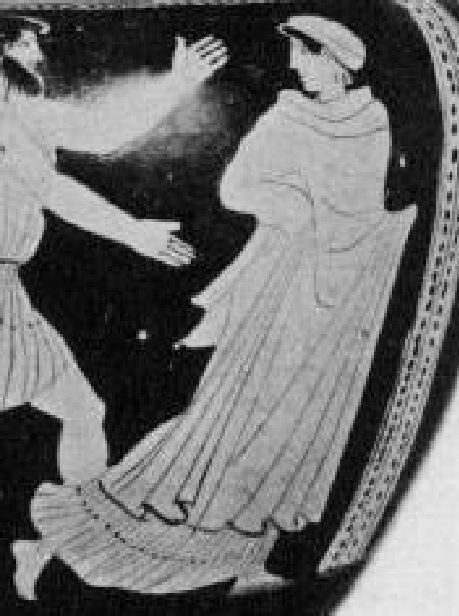}
         \caption{$m_2(q)$}
 		\label{fig:results-retrieval3}
 	\end{subfigure}
 	\,
 	\hfill
 	\begin{subfigure}{0.12\linewidth}
 		\centering
 		\includegraphics[width=0.93\linewidth,height=0.93\linewidth,cfbox=green 1pt 1pt]{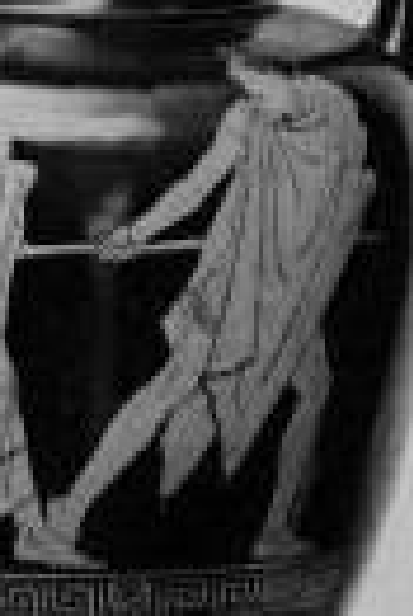}
         \caption{$m_3(q)$}
 		\label{fig:results-retrieval4}
 	\end{subfigure}
 	\,
 	\hfill
 	\begin{subfigure}{0.12\linewidth}
 		\centering
 		\includegraphics[width=0.93\linewidth,height=0.93\linewidth,cfbox=green 1pt 1pt]{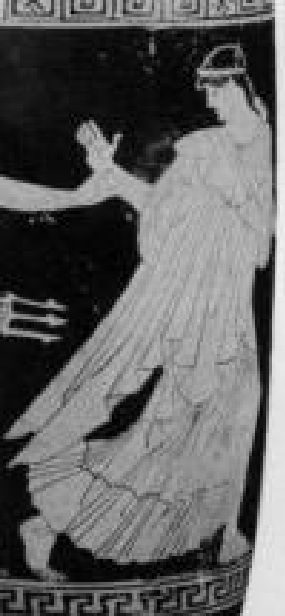}
         \caption{$m_4(q)$}
 		\label{fig:results-retrieval5}
 	\end{subfigure}
 	\,
 	\hfill
 	\begin{subfigure}{0.12\linewidth}
 		\centering
 		\includegraphics[width=0.93\linewidth,height=0.93\linewidth,cfbox=green 1pt 1pt]{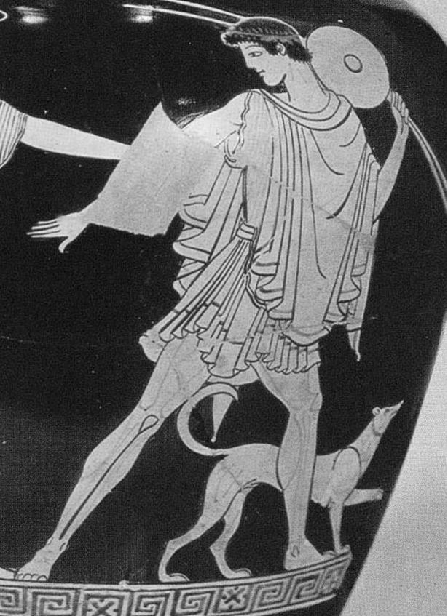}
 		\caption{$m_5(q)$}
 		\label{fig:results-retrieval6}
 	\end{subfigure}

 	\caption{\textbf{Discovery and Retrieval comparison}: 
 	\subref{fig:results-retrieval1} are query images (\emph{fleeing} character, 
 	\emph{pursuit} scene), and the remaining 5 columns ($m_{i}(q)$) are the five most similar images across four models; \nth{1} row -- 
 	\textit{baseline}, \nth{2} row -- \textit{styled}, \nth{3} row -- 
 	\textit{Tuned}, \nth{4} row -- \textit{Sd$\rightarrow$Td}, \nth{5} and 
 	\nth{6} row -- \textit{Sd$\rightarrow$Td$_{p1}$} and 
 	\textit{Sd$\rightarrow$Td$_{p2}$} respectively. The results clearly show 
 	that the styled-tuned models retrieve the most precise results based on 
 	poses.}
 	\label{fig:results-retrieval}
 \end{figure}

 \subsection{Retrieval Results and Discovery}
 Tab.~\ref{tab:retrieval} presents our pose-based retrieval results.
 We observe that \emph{styled-tuned} models are consistently better for \textbf{C} and \textbf{S}, and the \textit{styled} are better than \textit{baselines} counterparts.
 Fig.~\ref{fig:results-retrieval} displays a query image \subref{fig:results-retrieval1} along with the top-5 ranked retrievals for the six different evaluated models. 
 It can be observed that the tuned and the styled-tuned models outperform the baseline and the styled models.
 Fig.~\ref{fig:results-retrieval} row 1 shows poor retrieval results for the baseline model. 
 For a pursuit scene, the first two retrieved samples belong to a wrestling scene while the last 2 belong to leading of the bride scene. 
 The styled model (row 2) is already better wherein all the five retrievals belong to the pursuit scene, however at a character level, it retrieves a persecutor at the \nth{5} retrieval. 
 Tuned (row 3) and \emph{styled-tuned} (row 4-6) models perform similarly well. 
 On a closer look, we see that the \nth{2} and \nth{4} retrieved samples of the \emph{styled-tuned} model are closer to the query sample compared to the tuned model. 

 \section{Conclusion}
 We presented a two-stage training approach for using style transfer and transfer learning in combination with perceptual consistency to improve pose estimation in ancient Greek vase paintings. 
 We show that the use of styled transfer learning as a domain adaptation technique for such data significantly improves the performance of state-of-the-art pose estimation models on unlabelled data by~6\,\% mean average precision (mAP) as well as mean average recall (mAR).
 We also analysed the impact of styles as progressive learning in a comprehensive manner showing that models learn generic domain styles.
 We experimentally showed that our proposed method outperforms their corresponding counterparts for human pose estimation. 
 In general, our method can be applied to diverse unlabelled datasets without explicit supervised learning. 
 Our method also provides a way for exploring diverse cross-domain datasets with low or no labels using human poses as a tool. 
 Finally, we also show that our method can be used for pose-based image retrieval and discovery of similar, relevant poses and corresponding scenes in collections such as ancient Greek vase paintings. 
For future work, we plan to a) introduce the geometric structures of vases in COCO and Styled-COCO datasets as an augmentation technique during training, b) use shape information of the persons into our framework or using segmentation as a prior for vase paintings.

\begin{acks}
This paper is partially funded by the FAU Emerging Fields Initiative (\href{https://www.fau.de/research/verbundforschung/emerging-fields-initiative/}{EFI}) project ``Iconographics. Computational Understanding of Iconography and Narration in Visual Cultural Heritage'' as well as partially funded by the EU H2020 project ``Odeuropa'' under grant agreement No.\ 101004469. 
The authors would also like to thank NVIDIA for their hardware donation.
\end{acks}

\bibliographystyle{ACM-Reference-Format}
\bibliography{sample-base}


\begin{thebibliography}{56}


\ifx \showCODEN    \undefined \def \showCODEN     #1{\unskip}     \fi
\ifx \showDOI      \undefined \def \showDOI       #1{#1}\fi
\ifx \showISBNx    \undefined \def \showISBNx     #1{\unskip}     \fi
\ifx \showISBNxiii \undefined \def \showISBNxiii  #1{\unskip}     \fi
\ifx \showISSN     \undefined \def \showISSN      #1{\unskip}     \fi
\ifx \showLCCN     \undefined \def \showLCCN      #1{\unskip}     \fi
\ifx \shownote     \undefined \def \shownote      #1{#1}          \fi
\ifx \showarticletitle \undefined \def \showarticletitle #1{#1}   \fi
\ifx \showURL      \undefined \def \showURL       {\relax}        \fi
\providecommand\bibfield[2]{#2}
\providecommand\bibinfo[2]{#2}
\providecommand\natexlab[1]{#1}
\providecommand\showeprint[2][]{arXiv:#2}

\bibitem[\protect\citeauthoryear{Akiba, Sano, Yanase, Ohta, and Koyama}{Akiba
  et~al\mbox{.}}{2019}]%
        {optuna}
\bibfield{author}{\bibinfo{person}{Takuya Akiba}, \bibinfo{person}{Shotaro
  Sano}, \bibinfo{person}{Toshihiko Yanase}, \bibinfo{person}{Takeru Ohta},
  {and} \bibinfo{person}{Masanori Koyama}.} \bibinfo{year}{2019}\natexlab{}.
\newblock \showarticletitle{Optuna: A next-generation hyperparameter
  optimization framework}. In \bibinfo{booktitle}{\emph{Proceedings of the 25th
  ACM SIGKDD International Conference on Knowledge Discovery \& Data Mining}}.
  \bibinfo{publisher}{ACM}, \bibinfo{address}{United States},
  \bibinfo{pages}{2623--2631}.
\newblock


\bibitem[\protect\citeauthoryear{Becker}{Becker}{2013}]%
        {pathos}
\bibfield{author}{\bibinfo{person}{Colleen~M. Becker}.}
  \bibinfo{year}{2013}\natexlab{}.
\newblock \showarticletitle{Aby Warburg's Pathosformel as Methodological
  Paradigm}.
\newblock \bibinfo{journal}{\emph{The Journal of Art Historiography}}
  \bibinfo{volume}{9} (\bibinfo{year}{2013}), \bibinfo{pages}{9--CB1}.
\newblock


\bibitem[\protect\citeauthoryear{Bell and Impett}{Bell and Impett}{2019}]%
        {bell2019ikonographie}
\bibfield{author}{\bibinfo{person}{Peter Bell} {and} \bibinfo{person}{Leonardo
  Impett}.} \bibinfo{year}{2019}\natexlab{}.
\newblock \showarticletitle{Ikonographie und Interaktion. Computergest{\"u}tzte
  Analyse von Posen in Bildern der Heilsgeschichte}.
\newblock \bibinfo{journal}{\emph{Das Mittelalter}} \bibinfo{volume}{24},
  \bibinfo{number}{1} (\bibinfo{year}{2019}), \bibinfo{pages}{31--53}.
\newblock


\bibitem[\protect\citeauthoryear{Bell, Schlecht, and Ommer}{Bell
  et~al\mbox{.}}{2013}]%
        {bell2013nonverbal}
\bibfield{author}{\bibinfo{person}{Peter Bell}, \bibinfo{person}{Joseph
  Schlecht}, {and} \bibinfo{person}{Bj{\"o}rn Ommer}.}
  \bibinfo{year}{2013}\natexlab{}.
\newblock \showarticletitle{Nonverbal communication in medieval illustrations
  revisited by computer vision and art history}.
\newblock \bibinfo{journal}{\emph{Visual Resources}} \bibinfo{volume}{29},
  \bibinfo{number}{1-2} (\bibinfo{year}{2013}), \bibinfo{pages}{26--37}.
\newblock


\bibitem[\protect\citeauthoryear{Cao, Simon, Wei, and Sheikh}{Cao
  et~al\mbox{.}}{2017}]%
        {Cao_RealTimePoseEstimationAffinityFields_2017}
\bibfield{author}{\bibinfo{person}{Zhe Cao}, \bibinfo{person}{Tomas Simon},
  \bibinfo{person}{Shih-En Wei}, {and} \bibinfo{person}{Yaser Sheikh}.}
  \bibinfo{year}{2017}\natexlab{}.
\newblock \showarticletitle{Realtime multi-person 2d pose estimation using part
  affinity fields}. In \bibinfo{booktitle}{\emph{Proceedings of the IEEE
  conference on computer vision and pattern recognition}}.
  \bibinfo{publisher}{IEEE Computer Society}, \bibinfo{address}{United States},
  \bibinfo{pages}{7291--7299}.
\newblock


\bibitem[\protect\citeauthoryear{Carneiro, da~Silva, Del~Bue, and
  Costeira}{Carneiro et~al\mbox{.}}{2012}]%
        {Carneiro_ArtisticImageClassification_2012}
\bibfield{author}{\bibinfo{person}{Gustavo Carneiro},
  \bibinfo{person}{Nuno~Pinho da Silva}, \bibinfo{person}{Alessio Del~Bue},
  {and} \bibinfo{person}{Jo{\~a}o~Paulo Costeira}.}
  \bibinfo{year}{2012}\natexlab{}.
\newblock \showarticletitle{Artistic Image Classification: An Analysis on the
  PRINTART Database}. In \bibinfo{booktitle}{\emph{Computer Vision -- ECCV
  2012}}, \bibfield{editor}{\bibinfo{person}{Andrew Fitzgibbon},
  \bibinfo{person}{Svetlana Lazebnik}, \bibinfo{person}{Pietro Perona},
  \bibinfo{person}{Yoichi Sato}, {and} \bibinfo{person}{Cordelia Schmid}}
  (Eds.). \bibinfo{publisher}{Springer Berlin Heidelberg},
  \bibinfo{address}{Berlin, Heidelberg}, \bibinfo{pages}{143--157}.
\newblock
\showISBNx{978-3-642-33765-9}


\bibitem[\protect\citeauthoryear{Carreira, Agrawal, Fragkiadaki, and
  Malik}{Carreira et~al\mbox{.}}{2016}]%
        {Carreira_HumanPoseEstimationWithIterativeErrorFeedback_2016}
\bibfield{author}{\bibinfo{person}{João Carreira}, \bibinfo{person}{Pulkit
  Agrawal}, \bibinfo{person}{Katerina Fragkiadaki}, {and}
  \bibinfo{person}{Jitendra Malik}.} \bibinfo{year}{2016}\natexlab{}.
\newblock \showarticletitle{Human Pose Estimation with Iterative Error
  Feedback}. In \bibinfo{booktitle}{\emph{2016 IEEE Conference on Computer
  Vision and Pattern Recognition (CVPR)}}. \bibinfo{publisher}{IEEE},
  \bibinfo{address}{United States}, \bibinfo{pages}{4733--4742}.
\newblock
\urldef\tempurl%
\url{https://doi.org/10.1109/CVPR.2016.512}
\showDOI{\tempurl}


\bibitem[\protect\citeauthoryear{Cetinic, Lipic, and Grgic}{Cetinic
  et~al\mbox{.}}{2018}]%
        {cetinic2018fine}
\bibfield{author}{\bibinfo{person}{Eva Cetinic}, \bibinfo{person}{Tomislav
  Lipic}, {and} \bibinfo{person}{Sonja Grgic}.}
  \bibinfo{year}{2018}\natexlab{}.
\newblock \showarticletitle{Fine-tuning convolutional neural networks for fine
  art classification}.
\newblock \bibinfo{journal}{\emph{Expert Systems with Applications}}
  \bibinfo{volume}{114} (\bibinfo{year}{2018}), \bibinfo{pages}{107--118}.
\newblock


\bibitem[\protect\citeauthoryear{Chen, Wang, Peng, Zhang, Yu, and Sun}{Chen
  et~al\mbox{.}}{2018}]%
        {Chen_CascadedPyramidNetworksPoseEstimation_2018}
\bibfield{author}{\bibinfo{person}{Yilun Chen}, \bibinfo{person}{Zhicheng
  Wang}, \bibinfo{person}{Yuxiang Peng}, \bibinfo{person}{Zhiqiang Zhang},
  \bibinfo{person}{Gang Yu}, {and} \bibinfo{person}{Jian Sun}.}
  \bibinfo{year}{2018}\natexlab{}.
\newblock \showarticletitle{Cascaded Pyramid Network for Multi-person Pose
  Estimation}. In \bibinfo{booktitle}{\emph{2018 IEEE/CVF Conference on
  Computer Vision and Pattern Recognition}}. \bibinfo{publisher}{IEEE},
  \bibinfo{address}{United States}, \bibinfo{pages}{7103--7112}.
\newblock
\urldef\tempurl%
\url{https://doi.org/10.1109/CVPR.2018.00742}
\showDOI{\tempurl}


\bibitem[\protect\citeauthoryear{Crowley and Zisserman}{Crowley and
  Zisserman}{2014}]%
        {Crowley_ObjectRetrievalInPainiting_2014}
\bibfield{author}{\bibinfo{person}{Elliot Crowley} {and}
  \bibinfo{person}{Andrew Zisserman}.} \bibinfo{year}{2014}\natexlab{}.
\newblock \showarticletitle{The State of the Art: Object Retrieval in Paintings
  using Discriminative Regions}. In \bibinfo{booktitle}{\emph{British Machine
  Vision Conference, {BMVC} 2014, Nottingham, UK, September 1-5, 2014}}.
  \bibinfo{publisher}{{BMVA} Press}, \bibinfo{address}{UK}.
\newblock


\bibitem[\protect\citeauthoryear{Crowley and Zisserman}{Crowley and
  Zisserman}{2013}]%
        {crowley2013gods}
\bibfield{author}{\bibinfo{person}{Elliot~J Crowley} {and}
  \bibinfo{person}{Andrew Zisserman}.} \bibinfo{year}{2013}\natexlab{}.
\newblock \showarticletitle{Of gods and goats: Weakly supervised learning of
  figurative art}.
\newblock \bibinfo{journal}{\emph{learning}}  \bibinfo{volume}{8}
  (\bibinfo{year}{2013}), \bibinfo{pages}{14}.
\newblock


\bibitem[\protect\citeauthoryear{Deng, Zheng, Ye, Kang, Yang, and Jiao}{Deng
  et~al\mbox{.}}{2018}]%
        {deng2018image}
\bibfield{author}{\bibinfo{person}{Weijian Deng}, \bibinfo{person}{Liang
  Zheng}, \bibinfo{person}{Qixiang Ye}, \bibinfo{person}{Guoliang Kang},
  \bibinfo{person}{Yi Yang}, {and} \bibinfo{person}{Jianbin Jiao}.}
  \bibinfo{year}{2018}\natexlab{}.
\newblock \showarticletitle{Image-Image Domain Adaptation with Preserved
  Self-Similarity and Domain-Dissimilarity for Person Re-identification}. In
  \bibinfo{booktitle}{\emph{2018 IEEE/CVF Conference on Computer Vision and
  Pattern Recognition}}. \bibinfo{publisher}{IEEE}, \bibinfo{address}{US},
  \bibinfo{pages}{994--1003}.
\newblock
\urldef\tempurl%
\url{https://doi.org/10.1109/CVPR.2018.00110}
\showDOI{\tempurl}


\bibitem[\protect\citeauthoryear{Fang, Xie, Tai, and Lu}{Fang
  et~al\mbox{.}}{2017}]%
        {Fang_AlphaPose_2017}
\bibfield{author}{\bibinfo{person}{Hao-Shu Fang}, \bibinfo{person}{Shuqin Xie},
  \bibinfo{person}{Yu-Wing Tai}, {and} \bibinfo{person}{Cewu Lu}.}
  \bibinfo{year}{2017}\natexlab{}.
\newblock \showarticletitle{RMPE: Regional Multi-person Pose Estimation}. In
  \bibinfo{booktitle}{\emph{2017 IEEE International Conference on Computer
  Vision (ICCV)}}. \bibinfo{publisher}{IEEE}, \bibinfo{address}{US},
  \bibinfo{pages}{2353--2362}.
\newblock
\urldef\tempurl%
\url{https://doi.org/10.1109/ICCV.2017.256}
\showDOI{\tempurl}


\bibitem[\protect\citeauthoryear{Gatys, Ecker, and Bethge}{Gatys
  et~al\mbox{.}}{2016}]%
        {Gatys_ImageStyleTransferUsingCNNs_2016}
\bibfield{author}{\bibinfo{person}{Leon~A. Gatys},
  \bibinfo{person}{Alexander~S. Ecker}, {and} \bibinfo{person}{Matthias
  Bethge}.} \bibinfo{year}{2016}\natexlab{}.
\newblock \showarticletitle{Image Style Transfer Using Convolutional Neural
  Networks}. In \bibinfo{booktitle}{\emph{2016 IEEE Conference on Computer
  Vision and Pattern Recognition (CVPR)}}. \bibinfo{publisher}{IEEE},
  \bibinfo{address}{US}, \bibinfo{pages}{2414--2423}.
\newblock
\urldef\tempurl%
\url{https://doi.org/10.1109/CVPR.2016.265}
\showDOI{\tempurl}


\bibitem[\protect\citeauthoryear{Girshick}{Girshick}{2015}]%
        {Girshick_FastRCNN_2015}
\bibfield{author}{\bibinfo{person}{Ross Girshick}.}
  \bibinfo{year}{2015}\natexlab{}.
\newblock \showarticletitle{Fast R-CNN}. In \bibinfo{booktitle}{\emph{2015 IEEE
  International Conference on Computer Vision (ICCV)}}.
  \bibinfo{publisher}{IEEE}, \bibinfo{address}{US},
  \bibinfo{pages}{1440--1448}.
\newblock
\urldef\tempurl%
\url{https://doi.org/10.1109/ICCV.2015.169}
\showDOI{\tempurl}


\bibitem[\protect\citeauthoryear{Girshick, Donahue, Darrell, and
  Malik}{Girshick et~al\mbox{.}}{2014}]%
        {Girshick_RCNN_2014}
\bibfield{author}{\bibinfo{person}{Ross Girshick}, \bibinfo{person}{Jeff
  Donahue}, \bibinfo{person}{Trevor Darrell}, {and} \bibinfo{person}{Jitendra
  Malik}.} \bibinfo{year}{2014}\natexlab{}.
\newblock \showarticletitle{Rich Feature Hierarchies for Accurate Object
  Detection and Semantic Segmentation}. In \bibinfo{booktitle}{\emph{2014 IEEE
  Conference on Computer Vision and Pattern Recognition}}.
  \bibinfo{publisher}{IEEE}, \bibinfo{address}{US}, \bibinfo{pages}{580--587}.
\newblock
\urldef\tempurl%
\url{https://doi.org/10.1109/CVPR.2014.81}
\showDOI{\tempurl}


\bibitem[\protect\citeauthoryear{Giuliani}{Giuliani}{2003}]%
        {Giuliani_BildUndMithosGesichteGriechischenKunst_2003}
\bibfield{author}{\bibinfo{person}{Luca Giuliani}.}
  \bibinfo{year}{2003}\natexlab{}.
\newblock \bibinfo{booktitle}{\emph{Bild und Mythos: Geschichte der
  Bilderz{\"a}hlung in der griechischen Kunst}}.
\newblock \bibinfo{publisher}{CH Beck}, \bibinfo{address}{München}.
\newblock


\bibitem[\protect\citeauthoryear{Glorot and Bengio}{Glorot and Bengio}{2010}]%
        {glorot2010understanding}
\bibfield{author}{\bibinfo{person}{Xavier Glorot} {and} \bibinfo{person}{Yoshua
  Bengio}.} \bibinfo{year}{2010}\natexlab{}.
\newblock \showarticletitle{Understanding the difficulty of training deep
  feedforward neural networks}. In \bibinfo{booktitle}{\emph{Proceedings of the
  Thirteenth International Conference on Artificial Intelligence and
  Statistics}} \emph{(\bibinfo{series}{Proceedings of Machine Learning
  Research}, Vol.~\bibinfo{volume}{9})},
  \bibfield{editor}{\bibinfo{person}{Yee~Whye Teh} {and} \bibinfo{person}{Mike
  Titterington}} (Eds.). \bibinfo{publisher}{PMLR}, \bibinfo{address}{Chia
  Laguna Resort, Sardinia, Italy}, \bibinfo{pages}{249--256}.
\newblock


\bibitem[\protect\citeauthoryear{Hall, Cai, Wu, and Corradi}{Hall
  et~al\mbox{.}}{2015}]%
        {Cai_RecognisingObjectsInArt_2015}
\bibfield{author}{\bibinfo{person}{Peter Hall}, \bibinfo{person}{Hongping Cai},
  \bibinfo{person}{Qi Wu}, {and} \bibinfo{person}{Tadeo Corradi}.}
  \bibinfo{year}{2015}\natexlab{}.
\newblock \showarticletitle{Cross-depiction problem: Recognition and synthesis
  of photographs and artwork}.
\newblock \bibinfo{journal}{\emph{Computational Visual Media}}
  \bibinfo{volume}{1}, \bibinfo{number}{2} (\bibinfo{date}{01 Jun}
  \bibinfo{year}{2015}), \bibinfo{pages}{91--103}.
\newblock
\showISSN{2096-0662}
\urldef\tempurl%
\url{https://doi.org/10.1007/s41095-015-0017-1}
\showDOI{\tempurl}


\bibitem[\protect\citeauthoryear{He, Zhang, Ren, and Sun}{He
  et~al\mbox{.}}{2016}]%
        {He_DeepResidualLearningResNet_2016}
\bibfield{author}{\bibinfo{person}{Kaiming He}, \bibinfo{person}{Xiangyu
  Zhang}, \bibinfo{person}{Shaoqing Ren}, {and} \bibinfo{person}{Jian Sun}.}
  \bibinfo{year}{2016}\natexlab{}.
\newblock \showarticletitle{Deep residual learning for image recognition}. In
  \bibinfo{booktitle}{\emph{Proceedings of the IEEE conference on computer
  vision and pattern recognition}}. \bibinfo{publisher}{IEEE},
  \bibinfo{address}{US}, \bibinfo{pages}{770--778}.
\newblock


\bibitem[\protect\citeauthoryear{Hoffman, Tzeng, Park, Zhu, Isola, Saenko,
  Efros, and Darrell}{Hoffman et~al\mbox{.}}{2018}]%
        {hoffman2018cycada}
\bibfield{author}{\bibinfo{person}{Judy Hoffman}, \bibinfo{person}{Eric Tzeng},
  \bibinfo{person}{Taesung Park}, \bibinfo{person}{Jun-Yan Zhu},
  \bibinfo{person}{Phillip Isola}, \bibinfo{person}{Kate Saenko},
  \bibinfo{person}{Alexei Efros}, {and} \bibinfo{person}{Trevor Darrell}.}
  \bibinfo{year}{2018}\natexlab{}.
\newblock \showarticletitle{Cycada: Cycle-consistent adversarial domain
  adaptation}. In \bibinfo{booktitle}{\emph{International conference on machine
  learning}}. PMLR, \bibinfo{publisher}{PMLR}, \bibinfo{address}{Sweden},
  \bibinfo{pages}{1989--1998}.
\newblock


\bibitem[\protect\citeauthoryear{Huang, Liu, Van Der~Maaten, and
  Weinberger}{Huang et~al\mbox{.}}{2017}]%
        {Huang_DenselyConnectedConvolutionalNetworksDenseNet_2017}
\bibfield{author}{\bibinfo{person}{Gao Huang}, \bibinfo{person}{Zhuang Liu},
  \bibinfo{person}{Laurens Van Der~Maaten}, {and} \bibinfo{person}{Kilian~Q
  Weinberger}.} \bibinfo{year}{2017}\natexlab{}.
\newblock \showarticletitle{Densely connected convolutional networks}. In
  \bibinfo{booktitle}{\emph{Proceedings of the IEEE conference on computer
  vision and pattern recognition}}. \bibinfo{publisher}{IEEE},
  \bibinfo{address}{United States}, \bibinfo{pages}{4700--4708}.
\newblock


\bibitem[\protect\citeauthoryear{Huang and Belongie}{Huang and
  Belongie}{2017}]%
        {Huang_ArbitraryStyleTransferAdaptiveInstanceNormalization_2017}
\bibfield{author}{\bibinfo{person}{Xun Huang} {and} \bibinfo{person}{Serge
  Belongie}.} \bibinfo{year}{2017}\natexlab{}.
\newblock \showarticletitle{Arbitrary style transfer in real-time with adaptive
  instance normalization}. In \bibinfo{booktitle}{\emph{Proceedings of the IEEE
  International Conference on Computer Vision}}. \bibinfo{publisher}{IEEE},
  \bibinfo{address}{United States}, \bibinfo{pages}{1501--1510}.
\newblock


\bibitem[\protect\citeauthoryear{Impett and Moretti}{Impett and
  Moretti}{2017}]%
        {impett2017totentanz}
\bibfield{author}{\bibinfo{person}{Leonardo Impett} {and}
  \bibinfo{person}{Franco Moretti}.} \bibinfo{year}{2017}\natexlab{}.
\newblock \bibinfo{booktitle}{\emph{Totentanz. Operationalizing Aby Warburg’s
  Pathosformeln}}.
\newblock \bibinfo{type}{{T}echnical {R}eport}. \bibinfo{institution}{Stanford
  Literary Lab}, \bibinfo{address}{Stanford}.
\newblock


\bibitem[\protect\citeauthoryear{Inoue, Furuta, Yamasaki, and Aizawa}{Inoue
  et~al\mbox{.}}{2018}]%
        {inoue2018cross}
\bibfield{author}{\bibinfo{person}{Naoto Inoue}, \bibinfo{person}{Ryosuke
  Furuta}, \bibinfo{person}{Toshihiko Yamasaki}, {and}
  \bibinfo{person}{Kiyoharu Aizawa}.} \bibinfo{year}{2018}\natexlab{}.
\newblock \showarticletitle{Cross-domain weakly-supervised object detection
  through progressive domain adaptation}. In
  \bibinfo{booktitle}{\emph{Proceedings of the IEEE conference on computer
  vision and pattern recognition}}. \bibinfo{publisher}{IEEE},
  \bibinfo{address}{United States}, \bibinfo{pages}{5001--5009}.
\newblock


\bibitem[\protect\citeauthoryear{Insafutdinov, Pishchulin, Andres, Andriluka,
  and Schiele}{Insafutdinov et~al\mbox{.}}{2016}]%
        {Insafutdinov_DeeperCut_2016}
\bibfield{author}{\bibinfo{person}{Eldar Insafutdinov}, \bibinfo{person}{Leonid
  Pishchulin}, \bibinfo{person}{Bjoern Andres}, \bibinfo{person}{Mykhaylo
  Andriluka}, {and} \bibinfo{person}{Bernt Schiele}.}
  \bibinfo{year}{2016}\natexlab{}.
\newblock \showarticletitle{Deepercut: A deeper, stronger, and faster
  multi-person pose estimation model}. In \bibinfo{booktitle}{\emph{European
  Conference on Computer Vision}}. \bibinfo{publisher}{Springer},
  \bibinfo{address}{Netherlands}, \bibinfo{pages}{34--50}.
\newblock


\bibitem[\protect\citeauthoryear{Jenicek and Chum}{Jenicek and Chum}{2019}]%
        {Jenicek_LinkingArtThroughHumanPoses_2019}
\bibfield{author}{\bibinfo{person}{Tomas Jenicek} {and}
  \bibinfo{person}{Ond{\v{r}}ej Chum}.} \bibinfo{year}{2019}\natexlab{}.
\newblock \showarticletitle{Linking Art through Human Poses}. In
  \bibinfo{booktitle}{\emph{2019 International Conference on Document Analysis
  and Recognition (ICDAR)}}. \bibinfo{publisher}{IEEE},
  \bibinfo{address}{Australia}, \bibinfo{pages}{1338--1345}.
\newblock


\bibitem[\protect\citeauthoryear{Johnson, Hendriks, Berezhnoy, Brevdo, Hughes,
  Daubechies, Li, Postma, and Wang}{Johnson et~al\mbox{.}}{2008}]%
        {Johnson_ImageProcessingArtistIdentification_2008}
\bibfield{author}{\bibinfo{person}{C~Richard Johnson}, \bibinfo{person}{Ella
  Hendriks}, \bibinfo{person}{Igor~J Berezhnoy}, \bibinfo{person}{Eugene
  Brevdo}, \bibinfo{person}{Shannon~M Hughes}, \bibinfo{person}{Ingrid
  Daubechies}, \bibinfo{person}{Jia Li}, \bibinfo{person}{Eric Postma}, {and}
  \bibinfo{person}{James~Z Wang}.} \bibinfo{year}{2008}\natexlab{}.
\newblock \showarticletitle{Image processing for artist identification}.
\newblock \bibinfo{journal}{\emph{IEEE Signal Processing Magazine}}
  \bibinfo{volume}{25}, \bibinfo{number}{4} (\bibinfo{year}{2008}),
  \bibinfo{pages}{37--48}.
\newblock


\bibitem[\protect\citeauthoryear{Johnson, Alahi, and {Fei-Fei}}{Johnson
  et~al\mbox{.}}{2016}]%
        {johnsonPerceptualLossesRealTime2016}
\bibfield{author}{\bibinfo{person}{Justin Johnson}, \bibinfo{person}{Alexandre
  Alahi}, {and} \bibinfo{person}{Li {Fei-Fei}}.}
  \bibinfo{year}{2016}\natexlab{}.
\newblock \showarticletitle{Perceptual {{Losses}} for {{Real}}-{{Time Style
  Transfer}} and {{Super}}-{{Resolution}}}.
\newblock In \bibinfo{booktitle}{\emph{Computer {{Vision}} \textendash{}
  {{ECCV}} 2016}}, \bibfield{editor}{\bibinfo{person}{Bastian Leibe},
  \bibinfo{person}{Jiri Matas}, \bibinfo{person}{Nicu Sebe}, {and}
  \bibinfo{person}{Max Welling}} (Eds.). Vol.~\bibinfo{volume}{9906}.
  \bibinfo{publisher}{{Springer International Publishing}},
  \bibinfo{address}{{Cham}}, \bibinfo{pages}{694--711}.
\newblock
\showISBNx{978-3-319-46474-9 978-3-319-46475-6}
\urldef\tempurl%
\url{https://doi.org/10.1007/978-3-319-46475-6_43}
\showDOI{\tempurl}


\bibitem[\protect\citeauthoryear{Kingma and Ba}{Kingma and Ba}{2015}]%
        {adam}
\bibfield{author}{\bibinfo{person}{Diederik~P Kingma} {and}
  \bibinfo{person}{Jimmy~Lei Ba}.} \bibinfo{year}{2015}\natexlab{}.
\newblock \showarticletitle{Adam: A method for stochastic gradient descent}. In
  \bibinfo{booktitle}{\emph{ICLR: International Conference on Learning
  Representations}}. \bibinfo{publisher}{ICLR}, \bibinfo{address}{US},
  \bibinfo{pages}{--}.
\newblock


\bibitem[\protect\citeauthoryear{Li, Fang, Yang, Wang, Lu, and Yang}{Li
  et~al\mbox{.}}{2017}]%
        {li2017universal}
\bibfield{author}{\bibinfo{person}{Yijun Li}, \bibinfo{person}{Chen Fang},
  \bibinfo{person}{Jimei Yang}, \bibinfo{person}{Zhaowen Wang},
  \bibinfo{person}{Xin Lu}, {and} \bibinfo{person}{Ming-Hsuan Yang}.}
  \bibinfo{year}{2017}\natexlab{}.
\newblock \showarticletitle{Universal style transfer via feature transforms}.
  In \bibinfo{booktitle}{\emph{Advances in neural information processing
  systems}}. \bibinfo{publisher}{NIPS}, \bibinfo{address}{USA},
  \bibinfo{pages}{386--396}.
\newblock


\bibitem[\protect\citeauthoryear{Lin, Maire, Belongie, Hays, Perona, Ramanan,
  Doll{\'a}r, and Zitnick}{Lin et~al\mbox{.}}{2014}]%
        {Lin_COCODataset_2014}
\bibfield{author}{\bibinfo{person}{Tsung-Yi Lin}, \bibinfo{person}{Michael
  Maire}, \bibinfo{person}{Serge Belongie}, \bibinfo{person}{James Hays},
  \bibinfo{person}{Pietro Perona}, \bibinfo{person}{Deva Ramanan},
  \bibinfo{person}{Piotr Doll{\'a}r}, {and} \bibinfo{person}{C~Lawrence
  Zitnick}.} \bibinfo{year}{2014}\natexlab{}.
\newblock \showarticletitle{Microsoft coco: Common objects in context}. In
  \bibinfo{booktitle}{\emph{European conference on computer vision}}.
  \bibinfo{publisher}{Springer}, \bibinfo{address}{Switzerland},
  \bibinfo{pages}{740--755}.
\newblock


\bibitem[\protect\citeauthoryear{Long, Cao, Wang, and Jordan}{Long
  et~al\mbox{.}}{2015}]%
        {long2015learning}
\bibfield{author}{\bibinfo{person}{Mingsheng Long}, \bibinfo{person}{Yue Cao},
  \bibinfo{person}{Jianmin Wang}, {and} \bibinfo{person}{Michael Jordan}.}
  \bibinfo{year}{2015}\natexlab{}.
\newblock \showarticletitle{Learning transferable features with deep adaptation
  networks}. In \bibinfo{booktitle}{\emph{International conference on machine
  learning}}. \bibinfo{publisher}{PMLR}, \bibinfo{address}{France},
  \bibinfo{pages}{97--105}.
\newblock


\bibitem[\protect\citeauthoryear{Madhu, Kosti, M{\"u}hrenberg, Bell, Maier, and
  Christlein}{Madhu et~al\mbox{.}}{2019}]%
        {Madhu_RecognizingCharactersInArt_2019}
\bibfield{author}{\bibinfo{person}{Prathmesh Madhu}, \bibinfo{person}{Ronak
  Kosti}, \bibinfo{person}{Lara M{\"u}hrenberg}, \bibinfo{person}{Peter Bell},
  \bibinfo{person}{Andreas Maier}, {and} \bibinfo{person}{Vincent Christlein}.}
  \bibinfo{year}{2019}\natexlab{}.
\newblock \showarticletitle{Recognizing Characters in Art History Using Deep
  Learning}. In \bibinfo{booktitle}{\emph{Proceedings of the 1st Workshop on
  Structuring and Understanding of Multimedia heritAge Contents}}.
  \bibinfo{publisher}{ACM}, \bibinfo{address}{France}, \bibinfo{pages}{15--22}.
\newblock


\bibitem[\protect\citeauthoryear{Maria~Carlucci, Porzi, Caputo, Ricci, and
  Rota~Bulo}{Maria~Carlucci et~al\mbox{.}}{2017}]%
        {maria2017autodial}
\bibfield{author}{\bibinfo{person}{Fabio Maria~Carlucci},
  \bibinfo{person}{Lorenzo Porzi}, \bibinfo{person}{Barbara Caputo},
  \bibinfo{person}{Elisa Ricci}, {and} \bibinfo{person}{Samuel Rota~Bulo}.}
  \bibinfo{year}{2017}\natexlab{}.
\newblock \showarticletitle{Autodial: Automatic domain alignment layers}. In
  \bibinfo{booktitle}{\emph{Proceedings of the IEEE International Conference on
  Computer Vision}}. \bibinfo{publisher}{IEEE}, \bibinfo{address}{Italy},
  \bibinfo{pages}{5067--5075}.
\newblock


\bibitem[\protect\citeauthoryear{McNiven}{McNiven}{1983}]%
        {Mcniven_GesturesInAtticVasePaintings_1983}
\bibfield{author}{\bibinfo{person}{Timothy~John McNiven}.}
  \bibinfo{year}{1983}\natexlab{}.
\newblock \bibinfo{booktitle}{\emph{Gestures in Attic Vase Painting: use and
  meaning, 550-450 BC}}.
\newblock \bibinfo{publisher}{University of Michigan},
  \bibinfo{address}{Michigan}.
\newblock


\bibitem[\protect\citeauthoryear{Mensink and Van~Gemert}{Mensink and
  Van~Gemert}{2014}]%
        {Mensink_RijksmuseumChallenge_2014}
\bibfield{author}{\bibinfo{person}{Thomas Mensink} {and} \bibinfo{person}{Jan
  Van~Gemert}.} \bibinfo{year}{2014}\natexlab{}.
\newblock \showarticletitle{The rijksmuseum challenge: Museum-centered visual
  recognition}. In \bibinfo{booktitle}{\emph{Proceedings of International
  Conference on Multimedia Retrieval}}. \bibinfo{publisher}{ICML},
  \bibinfo{address}{China}, \bibinfo{pages}{451--454}.
\newblock


\bibitem[\protect\citeauthoryear{Moon, Chang, and Lee}{Moon
  et~al\mbox{.}}{2019}]%
        {Moon_2019_CVPR}
\bibfield{author}{\bibinfo{person}{Gyeongsik Moon}, \bibinfo{person}{Ju~Yong
  Chang}, {and} \bibinfo{person}{Kyoung~Mu Lee}.}
  \bibinfo{year}{2019}\natexlab{}.
\newblock \showarticletitle{PoseFix: Model-Agnostic General Human Pose
  Refinement Network}. In \bibinfo{booktitle}{\emph{Proceedings of the IEEE/CVF
  Conference on Computer Vision and Pattern Recognition (CVPR)}}.
  \bibinfo{publisher}{IEEE}, \bibinfo{address}{US}, \bibinfo{pages}{--}.
\newblock


\bibitem[\protect\citeauthoryear{Nevatia and Binford}{Nevatia and
  Binford}{1977}]%
        {Nevatia_DescriptionAndRecognitionOfCurvedObjects_1977}
\bibfield{author}{\bibinfo{person}{Ramakant Nevatia} {and}
  \bibinfo{person}{Thomas~O Binford}.} \bibinfo{year}{1977}\natexlab{}.
\newblock \showarticletitle{Description and recognition of curved objects}.
\newblock \bibinfo{journal}{\emph{Artificial intelligence}}
  \bibinfo{volume}{8}, \bibinfo{number}{1} (\bibinfo{year}{1977}),
  \bibinfo{pages}{77--98}.
\newblock


\bibitem[\protect\citeauthoryear{Newell, Yang, and Deng}{Newell
  et~al\mbox{.}}{2016}]%
        {Newell_StackedHourglassNetworksPoseEstimation_2016}
\bibfield{author}{\bibinfo{person}{Alejandro Newell}, \bibinfo{person}{Kaiyu
  Yang}, {and} \bibinfo{person}{Jia Deng}.} \bibinfo{year}{2016}\natexlab{}.
\newblock \showarticletitle{Stacked hourglass networks for human pose
  estimation}. In \bibinfo{booktitle}{\emph{European conference on computer
  vision}}. Springer, \bibinfo{publisher}{Springer},
  \bibinfo{address}{Netherlands}, \bibinfo{pages}{483--499}.
\newblock


\bibitem[\protect\citeauthoryear{Papandreou, Zhu, Kanazawa, Toshev, Tompson,
  Bregler, and Murphy}{Papandreou et~al\mbox{.}}{2017}]%
        {Papandreou_MultiPersonPoseEstimationInTheWild_2017}
\bibfield{author}{\bibinfo{person}{George Papandreou}, \bibinfo{person}{Tyler
  Zhu}, \bibinfo{person}{Nori Kanazawa}, \bibinfo{person}{Alexander Toshev},
  \bibinfo{person}{Jonathan Tompson}, \bibinfo{person}{Chris Bregler}, {and}
  \bibinfo{person}{Kevin Murphy}.} \bibinfo{year}{2017}\natexlab{}.
\newblock \showarticletitle{Towards accurate multi-person pose estimation in
  the wild}. In \bibinfo{booktitle}{\emph{Proceedings of the IEEE Conference on
  Computer Vision and Pattern Recognition}}. \bibinfo{publisher}{IEEE},
  \bibinfo{address}{United States}, \bibinfo{pages}{4903--4911}.
\newblock


\bibitem[\protect\citeauthoryear{Pishchulin, Insafutdinov, Tang, Andres,
  Andriluka, Gehler, and Schiele}{Pishchulin et~al\mbox{.}}{2016}]%
        {Pishchulin_DeepCut_2016}
\bibfield{author}{\bibinfo{person}{Leonid Pishchulin}, \bibinfo{person}{Eldar
  Insafutdinov}, \bibinfo{person}{Siyu Tang}, \bibinfo{person}{Bjoern Andres},
  \bibinfo{person}{Mykhaylo Andriluka}, \bibinfo{person}{Peter~V Gehler}, {and}
  \bibinfo{person}{Bernt Schiele}.} \bibinfo{year}{2016}\natexlab{}.
\newblock \showarticletitle{Deepcut: Joint subset partition and labeling for
  multi person pose estimation}. In \bibinfo{booktitle}{\emph{Proceedings of
  the IEEE conference on computer vision and pattern recognition}}.
  \bibinfo{publisher}{IEEE}, \bibinfo{address}{US},
  \bibinfo{pages}{4929--4937}.
\newblock


\bibitem[\protect\citeauthoryear{Ren, He, Girshick, and Sun}{Ren
  et~al\mbox{.}}{2017}]%
        {Girshick_FasterRCNN_2015}
\bibfield{author}{\bibinfo{person}{Shaoqing Ren}, \bibinfo{person}{Kaiming He},
  \bibinfo{person}{Ross Girshick}, {and} \bibinfo{person}{Jian Sun}.}
  \bibinfo{year}{2017}\natexlab{}.
\newblock \showarticletitle{Faster R-CNN: Towards Real-Time Object Detection
  with Region Proposal Networks}.
\newblock \bibinfo{journal}{\emph{IEEE Transactions on Pattern Analysis and
  Machine Intelligence}} \bibinfo{volume}{39}, \bibinfo{number}{6}
  (\bibinfo{year}{2017}), \bibinfo{pages}{1137--1149}.
\newblock
\urldef\tempurl%
\url{https://doi.org/10.1109/TPAMI.2016.2577031}
\showDOI{\tempurl}


\bibitem[\protect\citeauthoryear{Rodriguez and Mikolajczyk}{Rodriguez and
  Mikolajczyk}{2019}]%
        {dabmvc}
\bibfield{author}{\bibinfo{person}{AL Rodriguez} {and} \bibinfo{person}{K
  Mikolajczyk}.} \bibinfo{year}{2019}\natexlab{}.
\newblock \showarticletitle{Domain adaptation for object detection via style
  consistency}. In \bibinfo{booktitle}{\emph{BMVC}}. \bibinfo{publisher}{BMVA
  Press}, \bibinfo{address}{UK}, \bibinfo{pages}{--}.
\newblock


\bibitem[\protect\citeauthoryear{Roy, Siarohin, Sangineto, Bulo, Sebe, and
  Ricci}{Roy et~al\mbox{.}}{2019}]%
        {roy2019unsupervised}
\bibfield{author}{\bibinfo{person}{Subhankar Roy}, \bibinfo{person}{Aliaksandr
  Siarohin}, \bibinfo{person}{Enver Sangineto}, \bibinfo{person}{Samuel~Rota
  Bulo}, \bibinfo{person}{Nicu Sebe}, {and} \bibinfo{person}{Elisa Ricci}.}
  \bibinfo{year}{2019}\natexlab{}.
\newblock \showarticletitle{Unsupervised domain adaptation using
  feature-whitening and consensus loss}. In
  \bibinfo{booktitle}{\emph{Proceedings of the IEEE/CVF Conference on Computer
  Vision and Pattern Recognition}}. \bibinfo{publisher}{IEEE},
  \bibinfo{address}{US}, \bibinfo{pages}{9471--9480}.
\newblock


\bibitem[\protect\citeauthoryear{Russo, Carlucci, Tommasi, and Caputo}{Russo
  et~al\mbox{.}}{2018}]%
        {russo2018source}
\bibfield{author}{\bibinfo{person}{Paolo Russo}, \bibinfo{person}{Fabio~M
  Carlucci}, \bibinfo{person}{Tatiana Tommasi}, {and} \bibinfo{person}{Barbara
  Caputo}.} \bibinfo{year}{2018}\natexlab{}.
\newblock \showarticletitle{From source to target and back: symmetric
  bi-directional adaptive gan}. In \bibinfo{booktitle}{\emph{Proceedings of the
  IEEE Conference on Computer Vision and Pattern Recognition}}.
  \bibinfo{publisher}{IEEE}, \bibinfo{address}{US},
  \bibinfo{pages}{8099--8108}.
\newblock


\bibitem[\protect\citeauthoryear{Saleh and Elgammal}{Saleh and
  Elgammal}{2016}]%
        {Saleh_Elgammal_2016}
\bibfield{author}{\bibinfo{person}{Babak Saleh} {and} \bibinfo{person}{Ahmed
  Elgammal}.} \bibinfo{year}{2016}\natexlab{}.
\newblock \showarticletitle{Large-scale Classification of Fine-Art Paintings:
  Learning The Right Metric on The Right Feature}.
\newblock \bibinfo{journal}{\emph{International Journal for Digital Art
  History}} \bibinfo{volume}{2}, \bibinfo{number}{2} (\bibinfo{date}{Oct.}
  \bibinfo{year}{2016}), \bibinfo{pages}{--}.
\newblock
\urldef\tempurl%
\url{https://doi.org/10.11588/dah.2016.2.23376}
\showDOI{\tempurl}


\bibitem[\protect\citeauthoryear{Seguin, Costiner, di~Lenardo, and
  Kaplan}{Seguin et~al\mbox{.}}{2018}]%
        {Seguin_ArtDigitizationTechniques_2018}
\bibfield{author}{\bibinfo{person}{Benoit Seguin}, \bibinfo{person}{Lisandra
  Costiner}, \bibinfo{person}{Isabella di Lenardo}, {and}
  \bibinfo{person}{Fr{\'e}d{\'e}ric Kaplan}.} \bibinfo{year}{2018}\natexlab{}.
\newblock \showarticletitle{New techniques for the digitization of art
  historical photographic archives-the case of the cini foundation in venice}.
  In \bibinfo{booktitle}{\emph{Archiving Conference}}
  \emph{(\bibinfo{series}{2018}, \bibinfo{number}{1})}. Society for Imaging
  Science and Technology, \bibinfo{publisher}{Society for Imaging Science and
  Technology}, \bibinfo{address}{US}, \bibinfo{pages}{1--5}.
\newblock


\bibitem[\protect\citeauthoryear{Stansbury-O'Donnell}{Stansbury-O'Donnell}{2009}]%
        {Stansbury_StructuralDifferenciationPursuitScenes_2009}
\bibfield{author}{\bibinfo{person}{Mark Stansbury-O'Donnell}.}
  \bibinfo{year}{2009}\natexlab{}.
\newblock \showarticletitle{Structural differentiation of pursuit scenes}.
\newblock \bibinfo{journal}{\emph{$\sigma$ $\tau$o: Yatromanolakis}}
  \bibinfo{volume}{-}, \bibinfo{number}{-} (\bibinfo{year}{2009}),
  \bibinfo{pages}{341--372}.
\newblock


\bibitem[\protect\citeauthoryear{Sun and Saenko}{Sun and Saenko}{2016}]%
        {sun2016deep}
\bibfield{author}{\bibinfo{person}{Baochen Sun} {and} \bibinfo{person}{Kate
  Saenko}.} \bibinfo{year}{2016}\natexlab{}.
\newblock \showarticletitle{Deep coral: Correlation alignment for deep domain
  adaptation}. In \bibinfo{booktitle}{\emph{European conference on computer
  vision}}. Springer, \bibinfo{publisher}{Springer},
  \bibinfo{address}{Netherlands}, \bibinfo{pages}{443--450}.
\newblock


\bibitem[\protect\citeauthoryear{Sun, Xiao, Liu, and Wang}{Sun
  et~al\mbox{.}}{2019}]%
        {Sun_HighResolutionPoseEstimationHRNet_2019}
\bibfield{author}{\bibinfo{person}{Ke Sun}, \bibinfo{person}{Bin Xiao},
  \bibinfo{person}{Dong Liu}, {and} \bibinfo{person}{Jingdong Wang}.}
  \bibinfo{year}{2019}\natexlab{}.
\newblock \showarticletitle{Deep high-resolution representation learning for
  human pose estimation}. In \bibinfo{booktitle}{\emph{Proceedings of the IEEE
  conference on computer vision and pattern recognition}}.
  \bibinfo{publisher}{IEEE}, \bibinfo{address}{US},
  \bibinfo{pages}{5693--5703}.
\newblock


\bibitem[\protect\citeauthoryear{Tompson, Jain, LeCun, and Bregler}{Tompson
  et~al\mbox{.}}{2014}]%
        {Tompson_JointModelGraphicalPoseEstimation_2014}
\bibfield{author}{\bibinfo{person}{Jonathan~J Tompson}, \bibinfo{person}{Arjun
  Jain}, \bibinfo{person}{Yann LeCun}, {and} \bibinfo{person}{Christoph
  Bregler}.} \bibinfo{year}{2014}\natexlab{}.
\newblock \showarticletitle{Joint training of a convolutional network and a
  graphical model for human pose estimation}. In
  \bibinfo{booktitle}{\emph{Advances in neural information processing
  systems}}. \bibinfo{publisher}{NIPS}, \bibinfo{address}{Canada},
  \bibinfo{pages}{1799--1807}.
\newblock


\bibitem[\protect\citeauthoryear{Toshev and Szegedy}{Toshev and
  Szegedy}{2014}]%
        {Toshev_DeepPose_2014}
\bibfield{author}{\bibinfo{person}{Alexander Toshev} {and}
  \bibinfo{person}{Christian Szegedy}.} \bibinfo{year}{2014}\natexlab{}.
\newblock \showarticletitle{Deeppose: Human pose estimation via deep neural
  networks}. In \bibinfo{booktitle}{\emph{Proceedings of the IEEE conference on
  computer vision and pattern recognition}}. \bibinfo{publisher}{IEEE},
  \bibinfo{address}{USA}, \bibinfo{pages}{1653--1660}.
\newblock


\bibitem[\protect\citeauthoryear{Wei, Ramakrishna, Kanade, and Sheikh}{Wei
  et~al\mbox{.}}{2016}]%
        {wei2016convolutional}
\bibfield{author}{\bibinfo{person}{Shih-En Wei}, \bibinfo{person}{Varun
  Ramakrishna}, \bibinfo{person}{Takeo Kanade}, {and} \bibinfo{person}{Yaser
  Sheikh}.} \bibinfo{year}{2016}\natexlab{}.
\newblock \showarticletitle{Convolutional pose machines}. In
  \bibinfo{booktitle}{\emph{Proceedings of the IEEE conference on Computer
  Vision and Pattern Recognition}}. \bibinfo{publisher}{IEEE},
  \bibinfo{address}{US}, \bibinfo{pages}{4724--4732}.
\newblock


\bibitem[\protect\citeauthoryear{Westlake, Cai, and Hall}{Westlake
  et~al\mbox{.}}{2016}]%
        {westlake2016detecting}
\bibfield{author}{\bibinfo{person}{Nicholas Westlake},
  \bibinfo{person}{Hongping Cai}, {and} \bibinfo{person}{Peter Hall}.}
  \bibinfo{year}{2016}\natexlab{}.
\newblock \showarticletitle{Detecting people in artwork with cnns}. In
  \bibinfo{booktitle}{\emph{European Conference on Computer Vision}}.
  \bibinfo{publisher}{Springer}, \bibinfo{address}{Netherlands},
  \bibinfo{pages}{825--841}.
\newblock


\bibitem[\protect\citeauthoryear{Xiao, Wu, and Wei}{Xiao et~al\mbox{.}}{2018}]%
        {Xiao_SimpleBaselinePoseEstimation_2018}
\bibfield{author}{\bibinfo{person}{Bin Xiao}, \bibinfo{person}{Haiping Wu},
  {and} \bibinfo{person}{Yichen Wei}.} \bibinfo{year}{2018}\natexlab{}.
\newblock \showarticletitle{Simple baselines for human pose estimation and
  tracking}. In \bibinfo{booktitle}{\emph{Proceedings of the European
  conference on computer vision (ECCV)}}. \bibinfo{publisher}{ECCV},
  \bibinfo{address}{Germany}, \bibinfo{pages}{466--481}.
\newblock


\end{thebibliography}

\end{document}